# ENCRYPTION AND ENCODING OF FACIAL IMAGES INTO QUICK-RESPONSE AND HIGH-CAPACITY-COLOR-2D CODE FOR BIOMETRIC PASSPORT SECURITY SYSTEM

**A THESIS**

*Submitted by*

**ZIAUL HAQUE CHOUDHURY**
RRN: 140873107002

Under the guidance of
**(Dr. M. MUNIR AHAMED RABBANI)**

*in partial fulfillment for the award of the degree of*

**DOCTOR OF PHILOSOPHY**

**Department of Information Technology**

**School of Computer, Information and Mathematical Sciences**

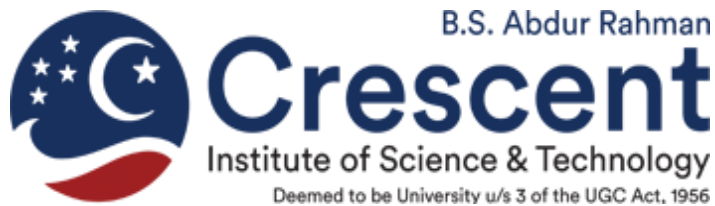

www.crescent.education

**MARCH 2021**

**B.S. Abdur Rahman**
# Crescent
**Institute of Science & Technology**
Deemed to be University u/s 3 of the UGC Act, 1956
www.crescent.education

Date: 05.01.2022

## CERTIFICATE

Correction carried out in Ph.D. thesis titled **"ENCRYPTION AND ENCODING OF FACIAL IMAGES INTO QUICK-RESPONSE AND HIGH-CAPACITY-COLOR-2D CODE FOR BIOMETRIC PASSPORT SECURITY SYSTEM"** submitted by the Research Scholar **Mr. ZIAUL HAQUE CHOUDHURY (RRN:140873107002).**

The thesis evaluation reports given by both Indian and Foreign examiners have been reviewed by the doctoral committee and the committee is completely satisfied with the reports of the examiners. The suggestions given by the examiner are incorporated and the corrected thesis is submitted.

Signature of the supervisor

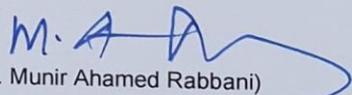

(Dr. M. Munir Ahamed Rabbani)

**B.S. Abdur Rahman**
**Crescent**
Institute of Science & Technology
Deemed to be University u/s 3 of the UGC Act, 1956
GST Road, Vandalur, Chennai 600 048

## OFFICE OF DEAN (Research)
(Empowering CRESCENT through Exemplary Research)

## PROCEEDINGS OF THE Ph.D. VIVA - VOCE EXAMINATION OF Mr. ZIAUL HAQUE CHOUDHURY

---

The Ph.D. VIVA - VOCE Examination of Mr. Ziaul Haque Choudhury (RRN:140873107002) on his Ph.D. Thesis entitled, "ENCRYPTION AND ENCODING OF FACIAL IMAGES INTO QUICK-RESPONSE AND HIGH-CAPACITY-COLOR-2D CODE FOR BIOMETRIC PASSPORT SECURITY SYSTEM" was conducted on 05.01.2022 at 11:00 A.M. through online mode.

The following Members of the Oral Examination Board were present:

| | | |
|---|---|---|
| 1. | Dr. Shashidhar G Koolagudi | Indian Examiner |
| 2. | Dr. H. Khanna Nehemia | Subject Expert |
| 3. | Dr. M. Munir Ahamed Rabbani | Supervisor & Convener |
| 4. | Dr. I. Sathik Ali | Ex-Officio |

The research scholar, Mr. Ziaul Haue Choudhury (RRN:140873107002) presented the salient features of his Ph.D. work. This was followed by questions from the Oral board members. The questions raised by the Foreign and Indian Examiners were also put to the scholar. The scholar answered the questions to the full satisfaction of the Oral board members.

The corrections suggested by the Indian and Foreign examiners have been carried out and incorporated in the Thesis before the Oral examination.

Based on the scholar's research work, his presentation and also the clarifications and answers by the scholar to the questions, the **Oral Examination Board recommends** that Mr. Ziaul Haque Choudhury be awarded Ph.D. degree in the Faculty / School of Computer, Information and Mathematical Sciences, Department of Information Technology.

1. Indian Examiner : Signature with date 05.01-2022

2. Subject Expert : Signature with date
   (H KHANNA NEHEMIAH)

3. Supervisor & Convener : Dr. M. A. RABBANI  Signature with date 5/1/2022

4. Ex-Officio : Signature with date 05/01/2022

## BONAFIDE CERTIFICATE

Certified that this thesis titled "ENCRYPTION AND ENCODING OF FACIAL IMAGES INTO QUICK-RESPONSE AND HIGH-CAPACITY-COLOR-2D CODE FOR BIOMETRIC PASSPORT SECURITY SYSTEM" is the bonafide work of Mr. ZIAUL HAQUE CHOUDHURY (RRN: 140873107002) who carried out the thesis work under our supervision. Certified further, that to the best of our knowledge, the work reported herein does not form part of any other thesis or dissertation on the basis of which a degree or award was conferred on an earlier occasion on this or any other candidate.

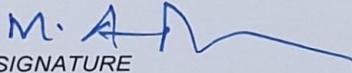
SIGNATURE
**Dr. M. MUNIR AHAMED RABBANI**
**RESEARCH SUPERVISOR**
Professor of IT & COE
B. S. Abdur Rahman Crescent
Institute of Science & Technology
Vandalur, Chennai - 600 048

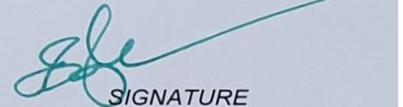
SIGNATURE
**Dr. I. SATHIK ALI**
**HEAD OF THE DEPARTMENT**
Department of Information Technology
B. S. Abdur Rahman Crescent
Institute of Science & Technology
Vandalur, Chennai - 600 048

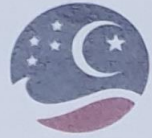

B.S. Abdur Rahman
Crescent
Institute of Science & Technology
Deemed to be University u/s 3 of the UGC Act, 1956
GST Road, Vandalur, Chennai 600 048

## DECLARATION BY THE RESEARCH SCHOLAR

This is to declare that the thesis entitled *"ENCRYPTION AND ENCODING OF FACIAL IMAGES INTO QUICK-RESPONSE AND HIGH-CAPACITY-COLOR-2D CODE FOR BIOMETRIC PASSPORT SECURITY SYSTEM"* submitted by me to the office of Dean (Research), B.S. Abdur Rahman Crescent Institute of Science & Technology, Chennai for the award of the degree of **Doctor of Philosophy** is a bonafide record of research work carried out by me under the supervision of **Prof. Dr. M. Munir Ahamed Rabbani,** Department of Information Technology. The contents of the thesis have not been submitted to any other Institute or University for the award of any other degree or diploma.

| | |
|---|---|
| Student Name | : Ziaul Haue Choudhury |
| RRN | : 140873107002 |
| Department | : Information Technology |
| Student's Signature | : 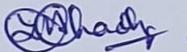 |
| Chennai | : 600048 |
| Date | : 28.1.2022. |

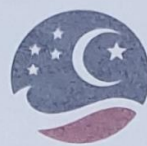

B.S. Abdur Rahman
**Crescent**
Institute of Science & Technology
Deemed to be University u/s 3 of the UGC Act, 1956
GST Road, Vandalur, Chennai 600 048

## Certificate for check against Plagiarism

This is to certify that the thesis titled *"Encryption and Encoding of Facial Images into Quick-Response and High-Capacity-Color-2D Code for Biometric Passport Security System"* submitted by **Mr. Ziaul Haque Choudhury (140873107002), Department of Information Technology** to the office of Dean (Research), B.S. Abdur Rahman Crescent Institute of Science & Technology, Chennai for award of **Ph.D.** is a bonafide record of the research work done by him under my supervision. The contents of the thesis have been verified for originality through plagiarism check software **"TURNITIN"** and no unacceptable similarity was found through the software check as per the UGC Norms.

**Name of the Research Supervisor : Prof. Dr. M. Munir Ahamed Rabbani**

**Designation of the Research Supervisor : Professor and COE**

**School of Computer, Information and Mathematical Sciences**

Signature: 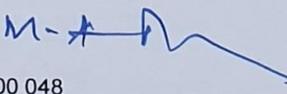

Chennai - 600 048

Office Seal

Date: 28-1-2022

Dr. M. MUNIR AHAMED RABBANI
CONTROLLER OF EXAMINATIONS
B.S. Abdur Rahman Crescent
Institute of Science and Technology
Vandalur, Chennai - 600 048.

# ACKNOWLEDGEMENT

First and foremost, my heart whelmed thanks to **The Almighty** for giving me an opportunity to do my doctoral degree in an esteemed institution.

I sincerely thank **Dr. A. PEER MOHAMED**, Vice-Chancellor and **Dr. A. AZAD**, Registrar, B.S. Abdur Rahman Crescent Institute of Science and Technology, for providing every essential facility for doing my Research Work.

I am extremely thankful to my supervisor **Dr. M. MUNIR AHAMED RABBANI**, COE and Professor, School of Computer, Information and Mathematical Sciences, for his constant encouragement, valuable suggestions and her keen interest in the completion of my research work.

I sincerely thank Professor **Dr. VENKATESAN SELVAM**, Dean (School of Computer, Information, and Mathematical Sciences), Professor **Dr. I. RAJA MOHAMED**, Dean (Academic Research), **Dr. MD KHURSHID ALAM KHAN**, Deputy Dean (Research), and **Dr. I. SATHIK ALI**, Associate Professor and Head, Department of Information Technology, for his professional guidance and continued assistance in the process.

I submit my sincere thanks to Professor Emeritus **Dr. P. ANANDHAKUMAR**, Department of Information Technology, Madras Institute of Technology, Anna University, Chennai, for their insightful comments, valuable advice, support and encouragement.

Last but not the least, I would like to thank all my professors and faculty members of IT department for their support. I am extremely indebted to my mother and family members for their adorable support throughout my research.

<div align="right">**ZIAUL HAQUE CHOUDHURY**</div>

# ABSTRACT

Nowadays, the use of forged e-passport is increasing which is threatening to national security. It is important to improve national security against international crime or terrorism. There is a weak verification process caused by a lack of identification processes such as a physical check, biometric check, and electronic check. The e-passport can prevent the passport clone, or forge from illegal immigration. Also, an e-passport contains the personal and biometric information of a person. It has some current and possible applications relating to national security and law enforcement such as border security, illegal immigration, terrorism, criminals, fake passport, and so on. In the past decade, several countries have implemented e-passport based on Machine Readable Travel Documents (MRTD) and International Civil Aviation Organization (ICAO) standardization. However, there is a threat scenario in the privacy infringement problem due to Radio Frequency Identification Device (RFID) such as data leakage threats, identity theft, tracking, and host listing.

In this thesis, a multimodal biometric, secure encrypted data and encrypted biometric encoded into the QR code-based biometric-passport authentication method is proposed for national security applications. Firstly, using the Extended Profile - Local Binary Patterns (EP-LBP), a Canny edge detector, and the Scale Invariant Feature Transform (SIFT) algorithm with Image File Information (IMFINFO) process, the facial mark size recognition is initially achieved. Secondly, by using the Active Shape Model (ASM) into Active Appearance Model (AAM) to follow the hand and infusion the hand geometry characteristics for verification and identification, hand geometry recognition is achieved. Thirdly, the encrypted biometric passport information that is publicly accessible is encoded into the QR code and inserted into the electronic passport to improve protection. Further, Personal information and biometric data are encrypted by applying the Advanced Encryption Standard (AES) and the Secure Hash Algorithm (SHA) 256 algorithm. It will enhance the biometric passport security system.

A face recognition based on facial blemishes detection algorithm for biometric passport authentication with encrypted High Capacity Color 2-Dimensional (HCC2D) code is proposed. This includes facial blemishes features detection to generate the template and encrypted by applying the Secure Force (SF) algorithm to secure biometric information. Facial blemishes are detected by applying the Active Shape Model (AAM) using Principle Component Analysis (PCA) and Canny edge detector with Speed Up Robust Feature Detection (SURF) algorithm. The location, size, and colors are detected to identify the person to improve accuracy. Finally, biometric data encrypted and encoded into the HCC2D code. This method will enhance biometric passport security to protect the biometric information from an intruder.

We have also demonstrated an algorithm and proposed for facial marks detection from cosmetic or makeup applied faces for a secure biometric passport in the field of personal identification for national security. This paper focuses on face recognition to improve the biometric authentication for a biometric-passport and also introduces facial permanent mark detection from the makeup or cosmetic applied faces, twins, and similar faces. An algorithm is proposed to detect the cosmetic applied facial permanent marks such as mole, freckle, birthmark, pockmark, and so on. Active Shape Model (ASM) into Active Appearance Model (AAM) using Principal Component Analysis (PCA) is applied to detect the facial landmarks. Facial permanent marks are detected by applying the Canny Edge Detector and Gradient Field – Histogram of Oriented Gradient (GF-HOG). The proposed method will enhance national security and it will improve the biometric authentication for the biometric-passport.

# TABLE OF CONTENTS

# LIST OF TABLES

# LIST OF FIGURES

# LIST OF ABBREVIATIONS

| | |
|---|---|
| AA | Active Authentication |
| AAM | Active Appearance Model |
| AES | Advanced Encryption Standard |
| ARR | Average Recognition Rate |
| ASM | Active Shape Model |
| BAC | Basic Access Control |
| CCA | Canonical Correlation Analysis |
| CDS | Document Signers Certificate |
| C_HOG | Canny with Histogram of Oriented Gradients |
| C_LoG | Canny with Laplacian of Gaussian |
| CNN | Convolutional Neural Network |
| C_S | Canny with SURF |
| CSCA | Country Signing Certification Authority |
| C_SIFT | Canny with Scale Invariant Feature Transform |
| C_SURF | Canny with Speeded Up Robust Features |
| DGs | Data Groups |
| DS | Document Signers |
| DSRC | Dedicated Short Range Communication |
| DST | Digital Signature Transponder |
| DVCA | Document Verifier Certification Authorities |
| EAC | Extended Access Control |
| EAC | East African Community |
| ECG | Electrocardiogram |
| EER | Equal Error Rate |
| eMRPs | Electronic Machine Readable Passports |
| EPC | Electronic Product Code |
| EP-LBP | Extended Profile-Local Binary Patterns |
| EU | European Union |
| FAR | False Acceptance Rate |
| FEI | Fundacao Educational Inaciana |
| FM | Fix Matrix |
| FRR | False Recognition Rate |
| FV | Fisher Vector |
| GF-HOG | Gradient Field-Histogram of Oriented Gradients |
| HCC2D | High Capacity Color Two Dimension |
| HOG | Histogram of Oriented Gradients |
| IC | Integrated Chip |
| ICAO | International Civil Aviation Organization |
| IITK | Indian Institute of Technology Kanpur |
| IMFINFO | Image File Information |
| IS | Inspection System |
| LDS | Logical Data Structure |
| LEAP | Localized Encryption and Authentication Protocol |
| LoG | Laplacian of Gaussian |
| LS | Left Shift |
| MIFS | Makeup Induced Face Spoofing |
| MRTD | Machine Readable Travel Documents |
| MRZ | Machine Readable Zone |
| NCC | Normalized Cross Correlation |
| OP | Operation Point |
| PC | Personal Computer |
| PCA | Principal Component Analysis |
| PKI | Public Key Infrastructure |
| PLS | Partial Least Squares |
| QR Code | Quick Response Code |
| rCCA | Regularized Canonical Correlation Analysis |
| RDSCNN | Residual Depth-wise Separable Convolution Neural Network |
| RF | Radio Frequency |
| RFID | Radio Frequency Identification |
| SF | Secure Force |
| SHA | Secure Hash Algorithm |
| SIFT | Scale Invariant Feature Transform |
| S_LoG | Sobel operator with LoG |
| SMT | Scars, Marks, and Tattoos |
| SOD | Security Object Descriptor |
| SPOC | Single Point of Contact |
| SURF | Speeded Up Robust Features |
| SVM | Support Vector Machine |
| TA | Terminal Authentication |
| VJ | Viola and Jones |
| 2D | Two Dimensional |

# LIST OF SYMBOLS

| | | |
|---|---|---|
| *I* | - | Each face picture |
| $S_i$ | - | Individual gallery face image |
| $S_\mu$ | - | Mean Shape |
| $M_s$ | - | User specific mask |
| $t_i$ | - | Threshold value |
| $C_c$ | - | Predetermined value |
| $I_1$ and $I_2$ | - | Facial image |
| $N_1$ and $N_2$ | - | Facial marks detection |
| $n_j \in N_2$ | - | Considered each facial mark |
| $x_j$ and $y_j$ | - | Central coordinate points |
| $n_i \in N_1$ | - | Each facial mark |
| $R_i$ | - | A rectangular area |
| $I_2$ | - | Central coordinates |
| $A$ | - | Annotated marks |
| $n_i$ | - | Recognized facial marks |
| $t_0$ | - | Generated threshold value |
| $W_{fmm}$ and $W_{fr}$ | - | The selection of the weight |
| $k_1$ | - | Hidden key |
| $SHk_1$. | - | Key altered from $K_1$ |
| $C_t$ | - | Ciphertext |
| $S_1$ | - | Countenance of characters of document for the key |
| $S_2$ | - | Character amount of the key document. |
| $P_i$ | - | The data for the pixel to occupy |
| $I_R$ | - | Each row of the image |
| $J_A$ | - | Pixel value |
| $S$ | - | First image |
| $k_2$ | - | Deciphered by $K_1$ |
| $K_3$ | - | Master key |
| $I$ and $T$ | - | Eigenvalues |
| $A_i$ and $A_t$ | - | Shape and texture |
| $I_\mu$ and $T_\mu$ ($b_X$ and $b_G$) | - | Means weight vectors of $I$ and $T$ |
| $I_{new}$ and $T_{new}$ | - | Facial image shape and texture |
| $F_i$ | - | Each face image |
| $F_\mu$ | - | Mean form |
| $\sigma$ | - | Variance of Gaussian filter |
| $R_k$ | - | Round key |
| $F$ | - | Function of the Secure Force |
| $T$ | - | Triangle |
| $T'$ | - | Corresponding triangle |
| $r'_1, r'_2,$ and $r'_3$ | - | Corresponding parallel vertices |
| $r_1, r_2,$ and $r_3$ | - | Parallel Vertices |
| $p$ | - | Point inside the triangle |
| $p'$ | - | Parallel point inside the triangle |
| $M_g$ | - | Generic mask |

# CHAPTER 1

# INTRODUCTION

## 1.1 OVERVIEW

Biometric techniques are the basis for secure identity authentication of a person in this advanced era. Numerous countries have commenced issuing biometric passports with a biometric information embedded chip. An electronic passport is known as an e-passport or biometric passport that contains the Radio Frequency Identification (RFID) embedded inside the front page or back page or center page. Since 1968, the International Civil Aviation Organization (ICAO) has initiated to operate based on the Machine Readable Travel Documents (MRTD) standard. The definition of a Machine Readable Zone (MRZ), its guidance and specification for biometric passport was published in 1980. It was implemented by the United States, Canada, and Australia. Eventually, in 1997 initiated to work on biometrics passport [10]. The attack in the United States of America on September 11, 2001, warned the government world-wide to deal with and study the border protection and the security consequences that were in exercise. One of the authentic technologies for preserving citizens' identity has been the application of biometrics internationally. ICAO started the research work on the biometric passports in the year 1997 and developed a set of international guidelines for the creation and specification of globally interoperable biometric standards. ICAO developed a new design for biometric passport embedding the RFID chip in 2004 with the development of biometric technology [11] [12] [13]. Since then the biometric passport technique has been a worldwide concern and a lot of work has been done to make it reliable and secure.

A digitized file that integrates security measures to authenticate the identity of the passport holder is an e-passport. Biometric passports are intended to strengthen border security, boost privacy protection against identity fraud and theft by ensuring authentication of the document bearer [1].

The system for issuing new biometric passports has been operationalized in several European countries. It is the responsibility of the ICAO to define the criteria and regulations that should be followed by biometric passports [2]. The criteria incorporate biometric features such as facial image, fingerprint, an iris image, and other biometrics with the RFID technologies to be applied and the Public Key Infrastructure (PKI) [3]. The demand for more reliable authentication methods must be deployed since the level of protection and authentication imposter has been raised. The new standardized biometric includes biometric features for instance facial, fingerprint, and iris recognition strengthen the security and safety mechanisms. While a biometric passport is a very advanced method of authentication, it can extend to many privacy issues and threats. RFID is a system that transfers identifying information from an electronic tag by applying radio waves. It is prone to remote attacks as the information stored in the RFID chip is transmitted wirelessly.

## 1.2 BACKGROUND OF THE STUDY

### 1.2.1 Biometric Passport

In 1988, Davida and Desmedt presented the definition of e-passport or biometric passport [14] [15]. An e-passport is an electronic identification certificate and a paper containing several features of biometric to certify a traveler's citizenship. It includes the front or back cover or the middle of the page with an antenna and a chip. The chip is used to store user information such as personal details and biometric information (facial photograph) that is printed on the biometric passport. A biometric passport also includes certain biometric identification according to the choices taken by different countries and technological advances. The data stored in the chip is based on the recommendation of the ICAO guideline doc 9303 protocol [16] [17] [18]. The ICAO presently considers facial recognition, fingerprint, and iris scan as standard biometric for identifiers. The ICAO-based biometric passport guideline clarifies the incorporation of chips with MRTDs besides using standards such as ISO/IEC 7816, ISO 14443, ISO11770, RSA, ISO 9796, DSA, and ECDSA. Figure 1.1 shows the biometric passport picture.

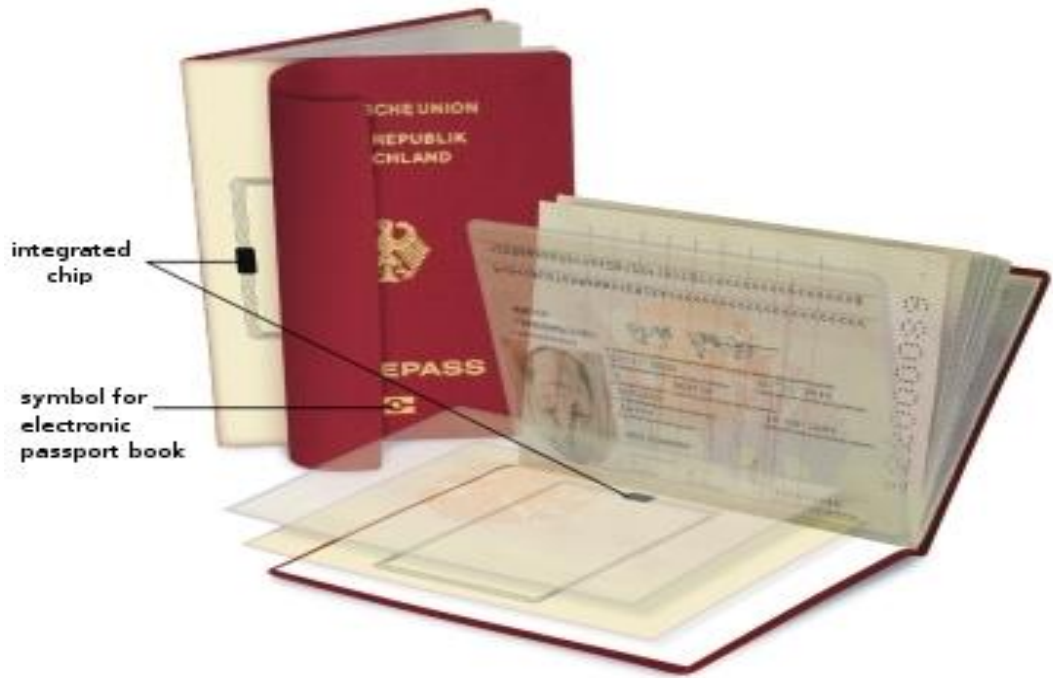

**Figure 1.1: Biometric Passport**

### 1.2.2 Biometric Passport Mechanism

A biometric passport is a basic document for the identification and enhancement of border control protection issues. In order to check the biometric passport portion of an MRZ to maintain the integrated data, the border protection officer will use the MRZ reader. By placing the biometric passport close to the biometric passport reader machine and the stored data is retrieved from the wireless chip. Eventually, information verification is carried out by means of Basic Access Control (BAC), Passive Authentication (PA) information encryption system, and integrity validation by applying either Active Authentication (AA) or passive authentication. PA is obligatory, while AA and BAC are non-mandatory.

## 1.2.3 Radio Frequency Identification (RFID)

Dedicated Short Range Communication (DSRC) is also known as RFID. RFID is a method that combines the electrostatic or electromagnetic coupling in the electromagnetic spectrum region of radio frequency (RF) to classify and monitor unique artifacts. RFID tags are employed all over, for instance, in inventory tracking services, smart cars, tracking animals, and particularly in biometric passports [19]. The RFID device diagram is given in figure 1.2. Three elements are included in the RFID system: a transceiver, a transponder, and an antenna. The transponder is triggered while obtains the signal from the antenna and transmit information back to the antenna after activation. In order to increase service efficiency in many applications, RFID techniques have been used.

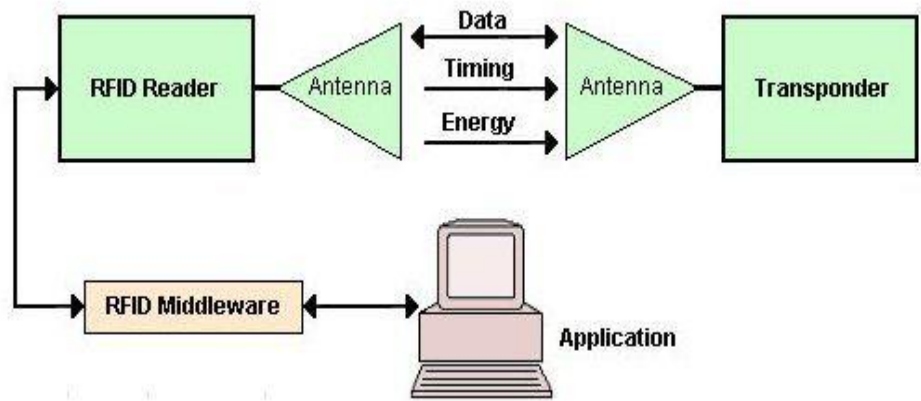

**Figure 1.2: RFID System**

Passive RFID tags are known as RFID tags that do not have onboard batteries. To activate the RFID, it employs power from the reader and begins to broadcast the signals endlessly inside a certain range up to a few meters. Alternatively, hundreds of feet can be transferred to the signal with active RFID tags that contain internal batteries. The RFID circuit diagram appears in figure 1.3.

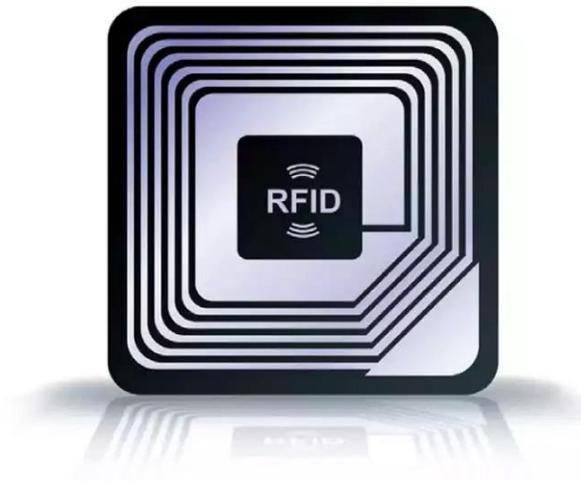

**Figure 1.3: RFID Circuit**

### 1.2.4  Symbol Inside the Chip

An MRTD has a wireless communication chip that is Integrated Circuit (IC) embedded inside the biometric passport that can be applied for biometric authentication based on ICAO resolution. All MRTDs are therefore labeled with the following symbol, the MRTD-based e-passport symbol is shown in figure 1.4. The symbol appears only on an MRTD that comprises a wireless communication facility with an integrated circuit on a biometric passport [18].

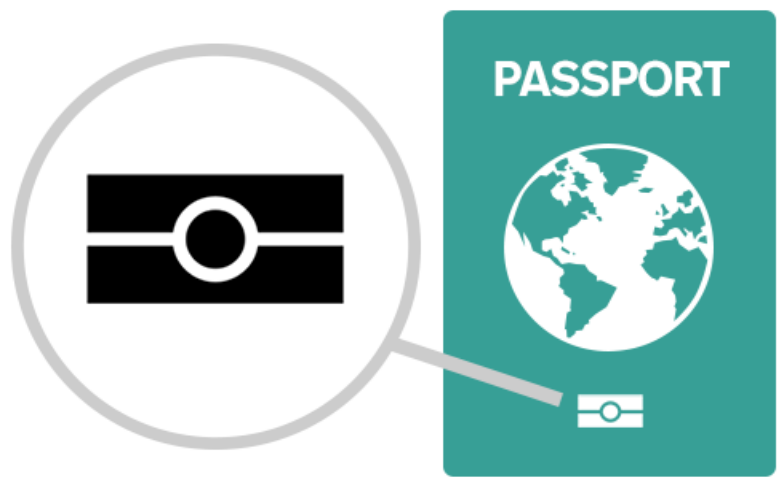

**Figure 1.4: The e-passport symbol based on MRTD**

## 1.2.5 Biometric Passport and Its Security

### 1.2.5.1 The ICAO Specification

Cryptographic algorithm standards and control techniques that are required to be enforced in biometric passports have been specified in the ICAO standards [18]. It has one compulsory cryptographic features and five advanced protection techniques that are optional.

### 1.2.5.2 Passive Authentication (PA)

For security and authentication, the ICAO requirements have both optional and mandatory features. Compulsory features offer only a minimal level of protection and do not preclude information stored on the contactless Integrated Chip (IC) from being copied. PA verifies and authenticates the unity of the information contained in the Logical Data Structure (LDS) and Security Object Descriptor (SOD). The PA process is employed to check the quality and validity of the data on the chip [20]. A dedicated PKI that is also known as the signing PKI, is introduced by the PA system. The chip carried out a Logical Data Structure (LDS) that coordinates information into Data Groups (DGs). The emerging body has independently measured and put these hashes in the Security Object Document to ensure the integrity of each of the DGs. A digital signature created over the immersion of the hash-values by the emerging agency (Document Signer, DS) ensures the validity of these hash-values. In all data types, the credibility and authenticity of all information are therefore assured. The IS computes the individual DG's hash values and reads the data to compare them with the hash values present in the SOD when a border control e-MRTD is provided.

The Inspection System (IS) affirms the signature to confirm that these hash-values are authentic and unchanged. A corresponding signature guarantees that the information is unchanged and genuine in the data classes. An IS requires a DS certificate that has generated a signature over the hashes of the data group in order to capable to execute PA. This certificate comprises the key required for the authenticity of this signature to be verified. This DS certificate is normally read from the e-MRTD. Else, it is important to have the DS certificate uncommitted from an extraneous origin. The recommended first-line delivery method for the DS certificate (CDS) is through the ICAO PKD based on the ICAO standard. ICAO suggests the inclusion of the CDS on the e-MRTD in the SOD. However, this is not a necessity, perhaps to avoid a huge storage space demand on the e-MRTD chip. Procuring the CDS from an external supplier can also be better.

**1.2.5.3 Active Authentication (AA)**

Active authentication helps the IS to differentiate among cloned and original e-MRTDs by inspecting the physical chip and the physical document for electronic information pertains. This is an optional mechanism and is therefore not applicable to all biometric passports [20]. In order to affirm that information is a part of the physical text, ICAO requests that the MRZ be associated with MRZ data from the data group (DG). A challenge reaction protocol is carried out between the IS and the chip to verify that the information is in the physical chip. A public key document stored in DG 15 and the corresponding private key in the protected portion of the chip is intended to be used. In the IS, there is a public key, but the private key is not available for reading. The initial e-MRTD knows only this private key.

A challenge is sent to the e-MRTD by the inspection system. With the private key, the e-MRTD signs this challenge and sends the answer for inspection to the IS. By checking the DG 15 signature with the public key, the IS affirms the response. The IS will check from the signature that the e-MRTD is the authentic private key and therefore genuine because of the key pair's similarity. To secure the chip as opposed to alteration or cloning, AA is a non-mandatory security feature that relies on public-key cryptography. To avoid replicating the SOD, it utilizes the challenge-response protocol and indicates that information has been read from the authentically selected chip. It also showed that there was no replacement for the chip.

**1.2.5.4 Basic Access Control (BAC)**

The BAC technique is designed to defend the chip information from unauthorized access and contactless communication between the reader and the chip from eavesdropping [20]. The IS needs access to the text's optically readable, individualized MRZ to execute BAC. To extract the symmetric cryptographic keys employed in BAC, the IS requires information from the MRZ such as the name, the holder's date of birth, document number, and the document expiry date. These keys provide accession to information on the chip and maintain the concealment (encryption) and integrity of messages between the Machine Readable Travel Document (MRTD) and IS in contactless communication. For European e-Machine Readable Travel Documents (e-MRTDs) and e-Machine Readable Passports (e-MRPs) are requires BAC for the European Union (EU). As a non-mandatory mechanism for e-MRTDs, the ICAO has defined BAC internationally. The result is that IS should access both MRTDs that endorse BAC and those that do not should be readable by IS. The paper contains all the details required to implement the optional BAC mechanism. To execute the BAC method, and an IS therefore does not require admittance to any external data.

**1.2.5.5 Extended Access Control (EAC)**

ICAO recommends that there should be more selective access to the more sensitive further extra biometric information and expresses that it can be achieved two types: information encryption or EAC. Although these choices are referred to by ICAO, at this time, ICAO does not recommend or define any criteria or practices in these fields [20]. Furthermore, the ICAO notes that the non-mandatory EAC mechanism is identical to the BAC mechanism so soon mentioned, but alternatively the document Basic Access keys, a document Extended Access key set is employed for EAC. The document description of an Extended Access key collection (chip-individual) is up to the compliance state. A symmetrical key can be found in the Extended Access Key Collection document. Extended Access Control includes processing potentialities on the e-MRTD chip.

The German Federal Office for Information Security (BSI) describes the EAC-mechanism in its technical guideline TR-03110, advanced protection measures for MRTD-Extended Access Control (EAC), Version 1.11 [23]. The European Union includes this EAC mechanism as an extra protection standard for the safety of the additional biometric data contained in the biometric passport (iris and fingerprint). EAC assures that only the IS approved by a biometric passport supplying authority may read the iris or fingerprint data of that biometric passport. To determine the validity of both e-MRTD and IS, EAC adds functionality. This allows only approved inspection systems to be able to provide access. In addition, EAC offers stronger cryptographic mechanisms than BAC for securing chip-reader contact. There are two components of EAC: Terminal Authentication and Chip Authentication.

**1.2.5.5.1 Chip Authentication**

It is a substitute for active authentication (chip cloning protection). The CA does not endure from so-called dispute semantics, as opposed to active authentication. In a specific situation, the semantics of the conflict will cause the transfer of the proprietor to be controlled. That's why Germany declined to admit an AA in its implementation of the e-Passport. A cryptographically strong mutual key for encoding the following communication is accessible after the DH process [24].

**1.2.5.5.2 Terminal Authentication**

Only licensed terminals have approved the admittance of biometric information for data classes. To access the information, the terminal must be fitted with a valid certificate from a specific country. A particular self-destruction period is set for each terminal. The length of this time is strictly dependent on the terms of employed of each terminal. Each terminal can be blocked and marked with a simple ID [25, 26].

### 1.2.5.6 Public Key Infrastructures (PKIs)

PKIs are required for the examination of biometric passports, it is categorized into three-class: Country signing PKI for passive authentication, Terminal Authentication, and Terminal Authentication.

### 1.2.5.6.1 Passive Authentication with Country Signing PKI

The PA technique is employed in the e-MRTD inspection operation to inspect the credibility and validity of the data on the e-MRTD chip [20]. An IS requires certificates to conduct PA from the country signing the emerging authority's PKI. It is known as a foreign authority. One or more Document Signers (DS) and a Country Signing Certification Authority (CSCA) constitute a country signing PKI hierarchy. Under the obligation of the emerging agency, DS and CSCA occupy. For a limited period and even also for a small number of biometric passports, DS keys are used. ICAO suggests that DS keys have a maximum use time of three months. After a certain total number of passports are signed with the key, some nations change the signing key, although the maximum time has not yet been reached. For the length of use of main pairs plus a maximum validity period of five or ten years, DS certificates are valid. This guarantees that the DS certificates are valid for a term of ten years and three months.

### 1.2.5.6.2 Terminal Authentication with Country Verifying PKI

In the e-MRTD inspection process, an Extended Access Control (EAC) is employed for the Terminal Authentication (TA) system to access more sensitive biometric data. A private-public key pair and a public key chain for TA to be carried out and monitored via the e-MRTD chip are included in an IS. One or more Document Verifier Certification Authorities (DVCA), a Country Verifying Certification Authority (CVCA), and Inspection Systems (IS) compose a country verifying the PKI hierarchy. The verifying authority is responsible for verifying the PKI. The IS requires keys for access to additional secure biometric information in domestic or international e-MRTDs and requires the national DVCA to request public-key certificates.

### 1.2.5.6.3 Communication Security with PKI

A Single Point of Contact (SPOC) is required to substitute certificates and certificate requests with other countries [26]. To secure communication between SPOCs, HTTPS (TLS) is hired. To use the HTTPS mechanism, a third PKI is required.

### 1.2.5.7 Machine Readable Passport

### 1.2.5.7.1 Machine Readable Travel Document (MRTD)

In 1968, MRTDs was started by the ICAO. A distinct MRZ should be optically scanned in 1980, the first specification states. It stores vital data such as the date of birth, name, document number, gender, date of validation, and citizenship of a person. Since 1997, ICAO has begun to extend its work to incorporate additional information, including biometrics. The fundamental requirements are that a facial picture, a digital signature of the issuing country, and a digital copy of the MRZ must be integrated into the MRTD. A PKI comes with a digital signature. The ICAO standard provides the PA data authentication protocol, optional BAC for chip security against unauthorized access, and AA for chip validation.

### 1.2.5.7.2 Logical Data Structure for MRTD

For global interoperability, an exchangeable Logical Data Structure (LDS) is required and the biometric passport guideline from the ICAO provides information on how a microchip inside the data should be stored. For the grouping of logical information together and collectively stored in an LDS, Data Group (DG) is employed. Data elements are divided into 16 data classes by the ICAO Guideline and the LDS is separated into three parts [18]. Table 1.1 includes the necessary and optional information elements for the LDS.

**Table 1.1   Logical Data Structure with Descriptions**

| | | | |
|---|---|---|---|
| Detail(s) Recoded in MRZ | | DG1 | Document type |
| | | | Issuing state or organization |
| | | | Name (of Holder) |
| | | | Document number |
| | | | Check digit-document number |
| | | | Nationality |
| | | | Date of birth |
| | | | Sex |
| | | | Date of expiry |
| | | | Check digit - DOB |
| | | | Optional data |
| | | | Composite check digit |
| | | | MRZ content specification |
| Encoded Identification Feature(s) | Global Interchange Feature | DG2 | Encoded face |
| | Additional Feature(s) | DG3 | Encoded finger(s) |
| | | DG4 | Encoded eye(s) |
| Displayed Identification Feature(s) | | DG5 | Displayed portrait |
| | | DG6 | Reserved for future use |
| | | DG7 | Displayed signature or usual mark |
| Encoded Security Feature(s) | | DG8 | Data feature(s) |
| | | DG9 | Structure feature(s) |
| | | DG10 | Substance feature(s) |
| | | DG11 | Additional personal detail(s) |
| | | DG12 | Additional document detail(s) |
| | | DG13 | Optional detail(s) |
| | | DG14 | Reserved for future use |
| | | DG15 | Active authentication public key information |
| | | DG16 | Person(s) to notify |

**1.2.5.7.2.1 Compulsory and Optional DG**

The information specified by the issuing state or agency is included in this section. It contains the information registered in the MRZ, DG1 contains basic personal data such as the name of the passport holder, passport number, nationality, date of birth, checksum, and expiry date which is the same data presented on the passport page. DG2 contains a digital photograph. DG3 stored the fingerprint, DG4 contains iris data and DG5 contains described on the page a photo of the individual. Other groups are carried out with additional data of a passport holder. The public key employed for EAC is kept in DG14 and DG15 carried out the public key for AA.

**1.2.5.8 Biometric Identification in Biometric Passport**

Biometric identification is a concept that defines an automated means of identifying a living person by measuring physiological or behavioral characteristics that are unique to each individual. A biometric passport employs three forms of physiological biometric identification systems are considered as facial data, finger data, and iris data [21]. According to ICAO, the mandatory biometric data is face recognition and optional biometrics are fingerprint recognition, iris recognition, and other related biometrics.

The specification for these forms of biometric identification is given in compliance with the international standard ISO / IEC 19794 [22] and ICAO, Doc 9303 [27] and complies with those standards for all issuing States. Digital photographs (JPEG2000 or JPEG format) and other biometric information are stored on the chip and electronic border control systems carry out a comparison outside of the passport.

The applications of biometrics methods based on ICAO vision includes:

- The stipulation of the primary exchangeable type of biometrics method for employ in border control (watch lists, verification) by carriers and issuers of documents, and specification of agreed additional biometrics methods.
- The stipulation of biometrics methods (verification, identification, and

watch lists) for employed by document issuers.
- 10-year data retrieval capability, the highest recommended validity for a travel paper.

To design a biometric passport based on ICAO and MRTDs, which involves the following procedures in a biometric passport:

(i) Capturing a raw biometric sample is the enrollment process. It is employed to take biometric image samples for storage for each new person (e-MRTD holder). The automated acquisition of biometrics is a capture device, such as a live-capture digital image camera, fingerprint scanner, photograph scanner, or live-capture iris zooming camera. For instance, for a facial recognition capture, normal pose facing the camera straight-on; whether fingerprints are captured rolled or flat; eyes fully open for iris capture; each capture system would require certain requirements and procedures specified for the capture process. For future confirmation of identity, the resulting image is compressed and then analyzed.

(ii) The template generation procedure retains the distinct and quotable biometric characteristics of the biometric image captured and typically extracts a template from the stored image using a proprietary software algorithm. This describes the image in a way that can then be contrasted with another sample image captured at the moment when confirmation of identity is needed and calculated by a comparative score. Quality control is inherent in this algorithm, in which, by some process, the sample is rated for consistency. The standards of quality must be as high as possible, as the quality of the originally captured image depends on all subsequent tests. If the result is not appropriate, the catch The procedure should be replicated.

(iii) In order to determine if an end-user has already registered with the system and if so, if the end-user has the same identity, the identification method takes the template derived from the new

(iii) sample and contrasts it with the templates of the registered end-users.

(iv) The verification process approves a new e-MRTD holder sample and compares it to a reference derived from that holder's stored image to decide if the holder has the same identity.

## 1.3 PROBLEM DEFINITIONS AND PURPOSE OF THE STUDY

Security measures carried out on RFID chips and knowledge about biometrics are unsafe. Since 1998, E-Passports have been in use, but the technology still remains insecure from attackers. In the privacy misdemeanor issue, there is a menacing scenario due to RFID tags, for instance, identity theft, threats of data leakage, host listing, and monitoring so on. The key vulnerability issues are: the RFID tag is disclosed to risks of information leakage as it can gather the data secretly by "skimming" or "eavesdropping" or other methods and RFID has a wide range of security problems that extend to problems with privacy violations [39, 40, 41]. As biometric passports have not been validated enough, the activity of terrorists and criminals in private and government property has increased considerably. The most common biometric attribute employed by individuals for specific validation is facial images. It is also estimated that this biometric approach should be extended for safety purposes. However, there are numerous possibilities of disappointment in face recognition because of poor matching photographs of new faces/unfamiliar faces from illicit migration, terrorism, fake passport holders, and criminals [101]. Subsequently, incorrect authentication is another concern due to cosmetic or makeup skin, similar and twins skin. Terrorists and illegal immigrants may clone and exploit the biometric passport for a wide variety of purposes.

It is important to concentrate on safe biometric authentication, especially face recognition, in order to avoid such activities. It is a non-intrusive approach and a publicly appropriate framework for applications for identification. To build a prototype that eliminates counterfeit, the replication of data, look-similar impostor, photo replacement, as any holder of a traditional passport booklet may do. To create a more reliable travel document that removes the fraud associated with an encrypted biometric passport information into QR and HCC2D code for a biometric passport with less human interference.  This system will permit to store the personal information and biometric data securely since it is not an active element. This study will strengthen the security system for biometric passports. Figure 1.5 shows the overall block diagram of the biometric passport.

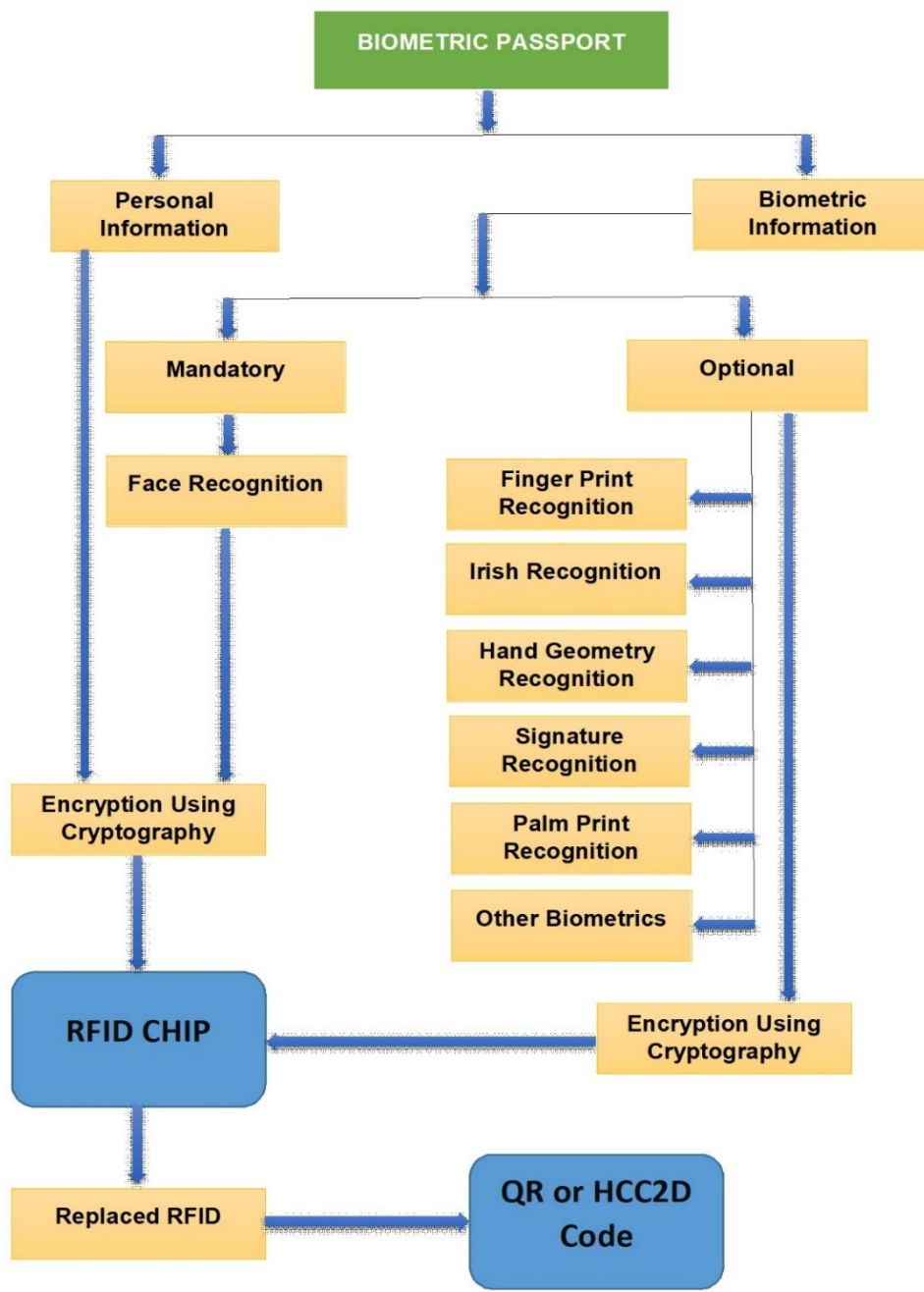

**Figure 1.5: Overall biometric passport diagram**

## 1.4 OBJECTIVE OF THE PROPOSED WORK

The E-passport includes extremely confidential personal details. The security of biographical and biometric information is very significant for an authentication system to be valuable and effective. Consequently, such information should also be protected from unauthorized access and the consistency of the system for information protection should be considered into priority.

The main objective of this research work is to adopt a technique of face recognition by overcoming the failure due to poor matching of images of unknown faces from illegal immigration, criminals, fake passport holders, and terrorism, and by encrypting the biometric passport information of the user to protect both personal and biometric information.

Thus, the main contribution of this research work are,

- A multi-biometric authentication and encryption using the Advanced Encryption Standard (AES) and Secure Hash Algorithm (SHA)-256 into Quick Response (QR) code is proposed in order to provide security.
- A facial blemishes detection and encryption by applying the Secure Force (SF) algorithm into the High Capacity Color Two Dimensional (HCC2D) code is introduced in order to improve security.
- Face recognition based on cosmetic applied is proposed to enhance biometric passport authentication.

## 1.5 ORGANIZATION OF THE THESIS

The organization of the thesis is directed toward secure biometric passport authentication by applying face recognition, multi-biometric authentication, and introducing biometric encryption and encoding into QR code and HCC2D code.

**Chapter 1** gives a general introduction of the biometric passport and background of the study on the biometric passport. It also narrates the purpose of the study and the objective of the research work.

**Chapter 2** discusses the related review thoroughly, including all significant particulars of previous research and various researcher's review of security and privacy emerges in the e-passport. The literature review demonstrates a systematic investigation to establish related research facts.

**Chapter 3** provides the overall framework of the proposed biometric passport authentication system. In this study, a framework to encrypt personal data and biometric information that are encoded into the QR code for biometric passport security is proposed. This method acquires facial mark size detection and hand geometry detection that is encrypted with AES and SHA-256 algorithm into the QR code to get over the menace challenges.

**Chapter 4** explains the face recognition method based on facial blemishes detection and encrypted by applying the SF algorithm into the HCC2D code. The facial blemishes detection technique is achieved by employing the Canny edge detector and Speeded Up Robust Features (SURF).

**Chapter 5** focuses on cosmetic based face recognition to ameliorate biometric verification for a biometric-passport and further presents facial permanent mark detection from the cosmetic or makeup utilized faces, similar faces, and twins. An algorithm is developed to find the permanent marks from the face, e.g. mole, birthmark, freckle, and pockmark from the cosmetic applied face. Active Shape Model (ASM) into Active Appearance Model (AAM) utilizing Principal Component Analysis (PCA) is used to find the facial landmarks. Facial permanent marks are detected by using the Canny Edge Detector and Gradient Field – Histogram of Oriented Gradient (GF-HOG).

**Chapter 6** presents the conclusions of the proposed research work and future enhancement.

# CHAPTER 2

# LITERATURE REVIEW

## 2.1 INTRODUCTION

To improve safety and protection at border checkpoints, several countries have adopted biometric passports in compliance with the ICAO electronic passport guidelines during the years 2004-2005 [29]. The biometric passport evolves on the basis of the ICAO standard has shown that it strengthens the biometric passports system with the consolidation of the MRTD biometric Wireless Microchip. An e-passport is required to be used for international travel by almost all individuals. It is created by binding a passport to a RFID system known as an electronic microchip. This chip acknowledges the information recognized by the biometric passport and enables it to be read by an appropriate e-passport readability system built at the border control point [29].

The biometric passport defines the option of facial recognition to be exchanged for machine-assisted validation internationally through biometric technology. Because of these factors, the technological and practical consideration of implementing the biometric technique in MRTDs is favored. It also considers the possibility that states can opt to carry supplements in order to identification and verification the iris, the fingerprint, the palm print, and face recognition. Due to global interconnectivity, the significance of national defense, border protection, and security has evolved rapidly from a national security perspective. To distinguish and safeguard against illegal immigrants, implementation includes the necessary authentication of the traveler. The importance of incorporating new e-passport components, such as biometric data to identify a person has been introduced [29].

In the government sector, biometric technologies are evolving more rapidly in terms of higher precision and citizens' security in verification and identification [30]. Bill Gates projected that the use of the biometric system for human characteristics, such as fingerprint, voice, and face recognition will be the most significant development in the successive decades of information technology during the PC Week online on October 8, 1997. The European Union submitted an electronic passport that was conceived to be the most important feature of information technology in the European countries in August 2006. Privacy, confidentiality, identity, and data are the methods suggested by Achilles Heel that lead to potential misapplication. A major biometric safety report, 2B or not 2B, was accomplished in the Netherlands [31, 135]. In their experiment trial passports were evolved with two forms of biometric information that contain fingerprints and facial images were conducted on a group of 15,000 individuals.

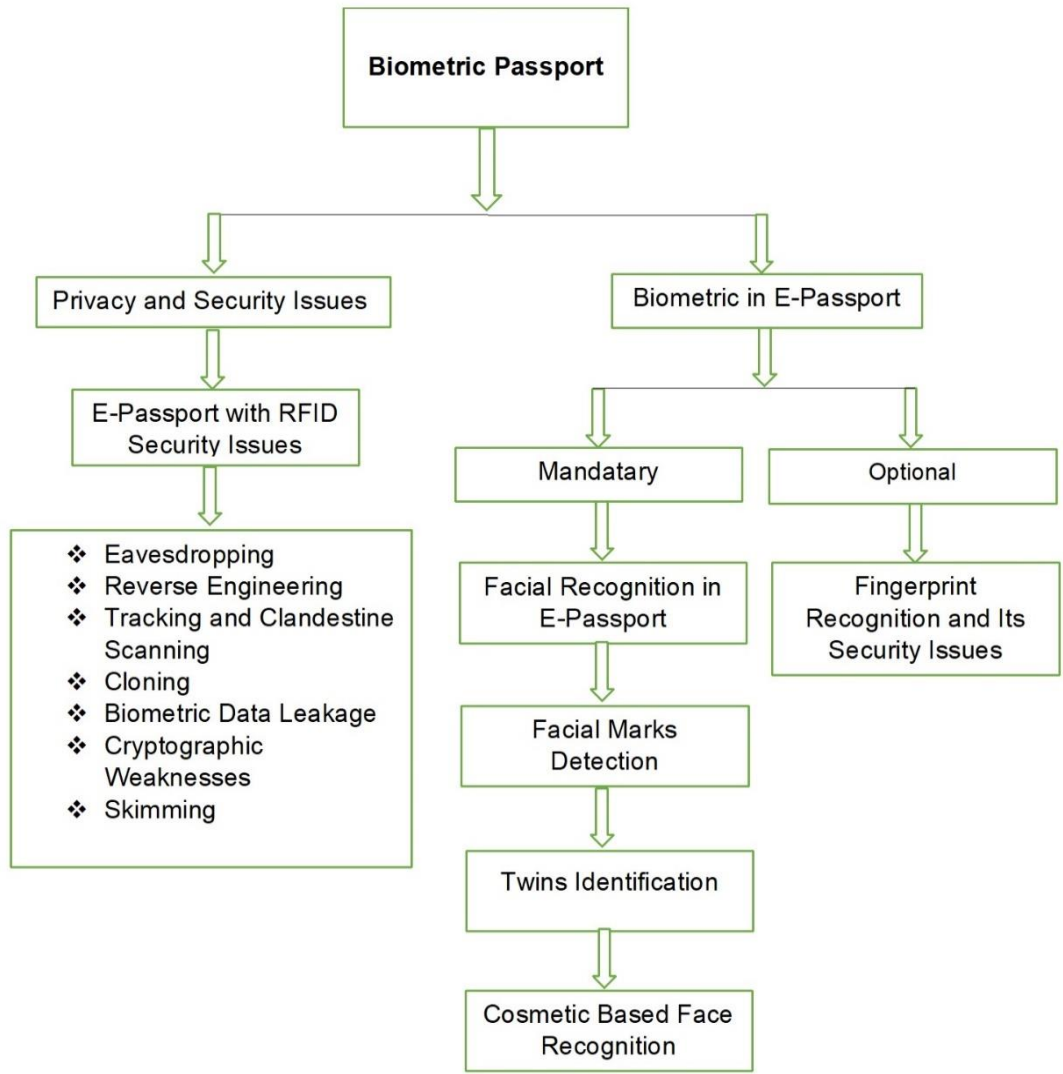

**Figure 2.1: E-passport and its security issues**

In March 2016, the Heads of State of the East African Community (EAC) ordered the implementation of biometric passports in all Member States, with an individual having a one-year phase-out of the existing community and national passports with acquiring electronic-passports [4]. However, the worries and fears of consumers, such as the exposure of personal information and the manipulation of data (misuse), remain a major research topic. This is because biometric technology has been clarified by a small number of studies from the viewpoint of consumer issues and acceptance [5]. Figure 2.1 displayed the e-passport and its security issues. One such analysis obtained that most consumers are concerned, skeptical, or disrupted by this technology as they see it as a way of probable invasions of their privacy. It is important to consider these users' issues, as well as risks and technology attacks. The feelings and opinions of individuals can enhance the hazard of denial and lead to dissatisfaction with the application of biometrics [6,7].

In addition to conducting a deeper understanding of the concerns and awareness of users [9] regarding biometric passport protection and data protection throughout the biometric issuing procedure, a framework for biometric passport applications was proposed for users to monitor information disclosure and unauthorized transmission of data. The study is against the claim of Ng-Kruelle et al. [8], which is against the violation of technical state power. Therefore, the technological creator must develop a biometric document that is very hard to replicate/duplicate. Develop a system that properly ensures the reference libraries and template database to establish and render the biometric application appropriate for the public and build trust with those running the applications in the framework itself.

### 2.1.1 Privacy and Security Issues

Due to the implementation of the RFID scheme, protection in the e-passport scheme has become of critical importance. RFID chips employ wireless communication that is prone to remote exposure. Attacks may be carried out on the contact network, chip, or backend device. Established some cryptographic tools and ensures e-passport confidentiality, accuracy, and authenticity of data. However, compared to others, the biometric framework offers an efficient and better part for user authentication, but it is not completely secure. Some vulnerabilities associated with data privacy problems have been emerging in recent years. Many attacks are possible at the hardware and software levels. Eavesdropping, covert scanning, reverse engineering, tracking, biometric data-leakage, cloning, cryptographic vulnerability, and skimming are the most common attacks.

## 2.1.2 E-Passport with Radio Frequency Identification (RFID) Security Issues

Different challenges to the threat posed by the biometric passport were addressed in terms of privacy and security [32]. They demonstrated that while electronic passport content without the knowledge of a user can be read in the short-range, even the e-passport booklet persists closed. Since there is an RFID chip in the biometric passport that is inserted within the biometric passport. In their study discussed regarding threats that the RFID tag contributes to data leakage because of eavesdropping or skimming, it can secretly capture the information. Laurie demonstrates the brute force attacks on the low entropy and hidden keys contained in RFID [33]. Monar, D. [136] talked about security and protection rises that utilize biometric passports. The communication bear on that, the contactless chip inserted in a biometric passport permits the substance to be read and authenticate without direct contact with a passport reader machine, and most significantly, the biometric passport booklet is unopened. The author debated that data stored in the RFID could be clandestinely gathered by methods for "eavesdropping" or "skimming". Horne, R., & Mauw, S., explained how privacy and security issues in the ICAO 9303 standard for e-passport. ICAO confirmed the vulnerabilities which permit an e-passport owner who recently passed through a checkpoint to be reidentified without opening their e-passport. Their study illustrated how bisimilarity was applied to discover these vulnerabilities that exploit the BAC protocol [160].

If the issuer of the biometric passport supplies the biometric passport, as stressed by the administrative officer or hotel clerk [34], to the biometric passport holder, an electronic-passport can be prone to splicing or fake finger attacks and other similar attacks. Johnson et. al. clarified that the biometric passport has had extensive press coverage of safety concerns, and studies indicate that there could be scope for biometric passport replication [35, 36, 37]. The issuance of travel documents [38] is correlated with some of the threat scenarios mentioned in the literature. Risks of data leakage, monitoring, host listing, and identity theft are discussed as complications that arise when electronic passports are issued [39]. Two levels of attack are susceptible to the RFID method, namely on the transport layers and network layers [40]. Cloning, impersonation, spoofing, eavesdropping, unauthorized tag scanning, application layer, and tag alteration can be used to attack the RFID tags. Four attack tactics are used by opponents to defend a computer using RFID, for instance, cloning, skimming, eavesdropping, and relaying [41].

Ryan publicly released a video in the Tech Insider as of lately, which the organization shot with the online security consulting company Red Team Security. This video shows how simple it is not just for hackers to replicate data on an RFID tag [42]. Nevertheless, to replicate the data to another card in order to generate an exclusively practicable clone. The seven forms of RFID protection attacks were discussed by Smiley: man-in-the-middle attack, power analysis, reverse engineering, sniffing, cloning & spoofing, replay and eavesdropping, viruses, and denial of service [43]. Ensuring a system that is fully secure, which ensures that the entire system is safe and functioning [44]. Due to the numerous sources of attacks and threats that may occur together in some cases, such as eavesdropping, denial of service, tracking, and cloning that are mentioned in their report, it remains a problematic task. It shows that for both manufacturers and users, the RFID tag has many protection and privacy issues. The following problems were identified: replay assault, eavesdropping, spoofing, deactivation, cloning, man-in-the-middle attack, and tag detachment [45].

Privacy is the greatest problem facing RFID technology, Jung & Lee explained and is of the most significant issues. The RFID tag comprises a large proportion of the user's private information, which, when leaked, may disturb his / her privacy. To prevent unauthorized access, cryptography has been established, but some problems remain that necessitate being covered [46]. In conclusion, in terms of technical concerns, the new RFID technology has not been developed to assure accomplished privacy. There are plenty of inadequacies in the technology itself and hackers will find vulnerabilities that kill the security mechanism of RFID. Rajaraman points to the illegal duplication of security-hampering RFID tags. Security experts will investigate RFID chips in London embedded in subway tickets, top-up any prepaid amount and make copies showing the security fault in the tag [47].

RFID tag deployment and usage of its rapid growth through multiple producers. Developers have used the chip for security services such as e-passports and built-in credit cards with RFID chips. Extensive research papers have been published for almost a decade discussing the security effects of RFID tags [48]. In relation to the RFID chip proposing some of the areas that need to be further studied, this study focuses on safety consequences. It applies not only to the management of RFID tag security issues but also to security-related problems such as jamming, replays attack, tag detaching, man-in-the-middle attack, eavesdropping, deactivation, cloning, and spoofing when considering RFID and its security. Wing shows that RFID poses security risks, and before implementation, other triggers should be warded off. Denial of service, tracking, RFID sniffing, insert attack, spoofing, repudiation, replay attack, physical attack, and viruses [49] are a few important RFID-related security problems.

In addition to some novel attacks, Avoine et. al. address the most important security and privacy outcomes of the contactless chip protocols that are introduced in e-Passports and present all related literature. RFID chip elaborates the protection of biometric passports and discloses them to new privacy issues and unregistered access or attack as a consequence of its contactless feature [50]. The author demonstrates that the use of an RFID device is a valid security issue, as maintaining the safety of RFID chips is a difficult job. The main issue is that there are multiple outlets that misinterpret data protection and readers, RFID chips, and networks can generate the attacks. Nevertheless, from a security point of view, privacy persists an essential measure to ensure data protection for an individual [51]. Pal demonstrates that RFID tags in supply chain automation are subject to different protection and related to privacy risks. The technological basics of RFID tags and their protection risks are outlined in this report. Besides, the latest privacy and security risks are defined by those that concentrate on the RFID scheme, for instance, the communication channel, tag, and the general risks to the system [52].

While Personal Computer (PC) chips and RFID chips were already employed to store information of the biometric, they are not suitable for affordable schemes. Besides, the RFID tag is readable, risking information leaks without anterior cognition [53]. The existing study discussed that ICAO's standard RFID-based e-passport has a significant amount of protection of security concerns and issues related to privacy violations [54] [32]. RFID found that tags are dangerous to demands illegal methods that provide access to sensitive information for the assassin. This is a reality that threatens the confidentiality of data security and privacy. An RFID reader that can generate a precise Electronic Product Code (EPC) can block the RFID system, and hackers who approach RFID readers or an EPC can block and read sensitive tag data [55]. Tag cloning, signal interference, tag killing, eavesdropping, denial of service attacks, and jamming will tamper with the RFID tag. A hacker will disrupt the frequency in any situation and prevent the message from arriving at the intended recipient. Although RFID is certainly an assuring technology, it proved that it has several technological demands related to it, namely the protection and privacy issues, collision issues, interface problems, and miscellaneous challenges [56].

**2.1.2.1 Eavesdropping**

This is the method by which the attacker collects data secretly to the contact channel and intercepts the data during the conversation between the legitimate reader and the chip by employing an unauthorized machine [57]. This is a kind of passive assault and very difficult to identify since driven signals do not emit. It has been documented that an intruder can eavesdrop on the RFID card communication network for up to at least 2 meters. The preparation of a biometric passport is not determining in airports only, but also in several commercial applications that enable a large eavesdropping scenario for the attacker.

### 2.1.2.2 Reverse Engineering

It is the procedure of bringing apart a device, entity, or system's technical principles to find out how it functions. A specific ID is employed as a private key by a biometric chip that is hardcoded in the development phase of a chip, so it is very hard to reverse engineer. However, if the attacker has effective proficient expertise and has admittance to instrumentation not widely available in the commercial market, the attacker may be able to reverse engineer. The ability to reverse biometric models which could lead to the rebuilding of physical characteristic images is a significant issue. Although it has been deemed unlikely, research uses the prototype to minimize the collection of theoretically matching images of the physical function.

### 2.1.2.3 Tracking and Clandestine Scanning

Clandestine scanning is outlined as the private mode of interpretation of an identity card's electronic data without the card holder's permission. Anyone with a reader can easily retrieve private details such as name, address, date of birth, biometric information, and nationality. Clandestine tracking is an opportunity to locate a person and the privacy of the place can be easily exposed. In contrast to clandestine scanning, clandestine tracking is more dangerous since the intruder will maintain global data monitoring without physical appearance. The Faraday cage was proposed to protect e-passports [32] in order to minimize clandestine scanning, but still, the problem remains in an e-passport.

### 2.1.2.4 Cloning

Chip cloning is the way to extract information from an authorized identification card and to create a new chip with an unauthorized copy of the captured taste. The researchers reported the cloning outcomes of a cryptographically safe Texas Instruments Digital Signature Transponder (DST) at RSA Labs and Johns Hopkins University that was employed to purchase gasoline and trigger the ignition of a car [58]. Active authentication is employed as a countermeasure to the hazard of cloning but can be bypassed by improving the biometric passport chip's EF.COM register.

**2.1.2.5 Biometric Data-Leakage**

The majority of biometric information is consistent. Once biometric information has been determined, it is not possible to replace it. It is commonly employed in electronic-passports to improve protection and privacy issues because of this property. The use of data-hiding is one of the strategies for rising information protection. Watermarking-based multimodal biometric methods are commonly used to secure biometric information from attacks and alteration. A multimodal biometric framework based on watermarking with two security levels is proposed and person verification and safety of biometric templates [11]. There are still the problem persists in e-passport due to data leakage threats.

**2.1.2.6 Cryptographic Weaknesses**

To securely store and transfer data, cryptographic methods have been employed in biometric-based electronic cards. Based on physical or behavioral characteristics, biometrics are commonly utilized for the automatic identification of a person. It suggested a new fingerprint-based authentication preservation scheme by applying the El Gamal cryptosystem for biometric comparison in the encrypted area [59]. The strategy does not have a key sharing facility and is only employed for authentication. Without experimental evaluation, the authentication protocol for a biometric passport established on elliptic curve encryption is more theoretical [60]. Finally, it is concluded that cryptography relied on by ICAO has some vulnerabilities.

**2.1.2.7 Skimming**

It is the process of accessing encoded information by means of electronic storage devices without the permission of users. E-passport data can be retrieved within a few inches or at most a few feet by transmitting power to the passport. However, the range can be extended if the reader transmits the signal with increased power. There are more chances of vulnerabilities due to skimming.

**2.1.3 Biometrics in E-Passport**

Techniques based on biometric characteristics are commonly used, particularly in areas where a higher level of protection or accurate identification is required. Nevertheless, for more ordinary purposes, all developments are becoming affordable as well. In the following decades, therefore, we should foresee a huge application of biometric-based products [61]. For each individual, a right biometric characteristic should be singular and it should be constant in time (generally from a particular age); it is an unambiguous identifier of an individual provided in the simplest possible way. In addition, few of the biometric characteristics have been well established and have also been specially employed for a long period of time, such as criminalistics fingerprints. Many of the biometric features, on the other hand, have been explored relatively recently. Since a detailed overview of biometrics is not feasible, let us concentrate on the features that are critical for the implementation of modern-day passports: two-dimensional facial photographs and fingerprints (in the near future, the employ of iris can be expected) [62].

In addition to the basic MRTD requirements laid down in Sections 3, 4, 5, 6, and 7 of Doc 9303 [27], Part 9 specifies the requirements to be applied for states that bid to provide an electronic-machine readable travel document (eMRTD) can apply to read, authenticate, and verify the information related to eMRTD itself from any adequately equipped receiving state. This requires compulsory biometric data that are interoperable internationally and can be employed mandatory as input facial recognition systems and optionally in iris recognition or fingerprint recognition systems.

**2.1.3.1 Facial Recognition in E-Passport**

As a basic safety feature, facial photography of an applicant is used. This form of protection is also well understood from older types of documents. The face picture mainly serves for visual recognition by officers in classic paper records. Despite the training of the officers and their ability to recognize a person the case of similar individuals such as siblings, twins, or even doubles can extend to identity mismatch even if there is any change in the appearance of an applicant, such as haircut, beard, glasses, etc. The face carried out data that is constant in time and can be evaluated, such as the chin position, distance between eyes, nose position, and so on, if the facial photo is treated from a biometric point of view. The identification process can be influenced by these factors by supplying the officer with additional details. The twins, however, will also look identical. This is why a completely different component of defense is required [62].

A safe multi-biometric system based on the Convolution Neural Network (CNN) merging with fingerprint and Electrocardiogram (ECG). They claimed that their study was the first to combine fingerprint and ECG for human authentication by applying CNN [162]. An ECG-based authentication scheme for human authentication by the use of CNN and an efficient algorithm for feature detection is discussed [161]. Yang, Y. et.al. talked about BehaveScense, a precise and powerful continuous authentication technique for sensitive-security mobile apps by applying touch-based behavioral biometrics to protect user security and save energy [163]. To provide a technique for improving the security of the image on the cloud. In their study introduced ensuring images on the cloud platform by applying biometric authentication [164]. A fast and accurate ECG authentication using two steps as classification and ECG beat detection. Feature extraction and ECG signal preprocessing are done by minimizing time-consuming. For ECG beat detection, Hamilton's technique was applied and the Residual Depthwise Separable Convolutional Neural Network (RDSCNN) technique was applied for classification [165]. These methods can also be used for passport authentication.

Although there is growing interest in supporting this approach with different biometric markers, verification of photo-ID is the most common means of identification, and we depend on trained specialists to reliably perform this function. Hence, analysis systematically demonstrates that spectators are inadequate to fit images of unknown faces [63, 68], making a surprisingly large amount of errors even though high-quality side-by-side photographs taken on the same day are shown. Moreover, it is no simpler to fit a live person to a picture [66-68], a finding that calls into question the employ of photo-ID. First, experience executing unknown face-matching tasks may improve precision as part of daily work. It is well known that people are extremely precise in matching known faces [69], making their ineffective success with unknown faces all the more striking [70]. Perhaps one aspect that contributes to the complexity of known face matching is that people in their everyday experience rarely encounter this task: the immense majority of face processing is oriented towards faces [71].

Experimental participants are often surprises by the complexity of unfamiliar face matching [63], pointing that poor execution of the test may stem from the originality of the works. This novelty is lost in occupational environments. As part of their jobs, the passport workers checked had all obtained training in facial image comparison. The goal of this training is to equip passport officers with more efficient facial image comparison strategies. There are rare cases of effective training for unfamiliar face matching tasks, and there have been accounts of some null outcomes [72, 73]. However, recent work has shown that certain forms of training can increase the performance of face matching [74].

**2.1.3.1.1 Facial Marks Detection**

By improving the representation structures of features, the current studies evaluated the better shape of face recognition technology. These recognize that on the face, salient skin regions grow for example a mole, wrinkles, scars, and freckles, etc. [75]. It shows in the literature that by applying the standard data collection of face images, facial marks are focused on measuring the efficacy of face recognition. Another study indicated that by employing the semi-automatic process, for soft biometrics and twin recognition, facial markings are feasible [76]. They also incorporated demographic data such as ethnicity and gender recognition in their process, for cosmetically applied faces, however, it is not such a successful technique. They employed the Active Appearance Model (AAM) algorithm and used the Laplace of Gaussian (LoG) technique to detect facial features.

In the existing literature, facial marks were obtained to have been used for the purpose of face recognition, which was very unusual in [76] [77]. It showed that facial probe has allowed for the recognition of 'individual' and 'class' features and the 'class' also involves the existence of hair, overall facial form, the color of hair, presence of marks and nose shape, so on. [75, 78]. Similarly, scars, the location and number of tattoos, and the location of wrinkles on the face are individual characteristics. The Scale Invariant Feature Transform (SIFT) algorithm had used [79], and to combine them to remove facial irregularities with global face matchers [80]. Nevertheless, it does not seem to decide the particular form of facial marks. Therefore, for facial database indexing, their proposed technique is not suitable. In the tattoo-based image recovery system, taking in Scars, Marks, and Tattoos (SMT), tattoos may appear in any part of an individual and are more definitive [81]. The appearance of markings directly on the facial parts that indicate simple morphologies is of interest to us.

A method to recognize moles by employing the techniques of normalized cross-correlation and a proposed morphable model [77]. The authors argued that a 3D morphable structure was eventually added to the process and lighting invariant and that various types of facial marks were not taken into account except for moles. Previous studies focused on the identification of facial markings, extended to biometric passport protection with advanced cosmetic applied face detection techniques [82, 83]. The identification of scars and acne is presented on the basis of automatic face classification [110, 111]. The existing study depends on the detection of facial marks since the authors add to previous studies by familiarizing systems with the recognition of the facial mark size. This offers more precision, and it will increase the precise and verifiable identification of the face without false recognition.

**2.1.3.1.2 Twins Identification**

Researchers have begun to focus on the issues involved in order to differentiate between identical twins. A palm print feature was employed to classify the associated characteristics of the twins and was able to distinguish between them [84]. Another study found the same fingerprints, and they noted that the presence of fingerprints was similar to and could differentiate between identical twins [85]. It obtained that genetically identical twins with unrelated individuals are uncorrelated to iris patterns [86]. By employing 2-D face recognition, analyzed ten sets of identical twins [87]. In an analysis of the difference of biometric features in identical twins, Sun et al. implemented the application of iris, fingerprint, and face biometrics. Fingerprint and Irish displayed abjection in results in their experiment and face matchers had difficulty differentiating between identical twins [88].

Their studies have been carried out and measured on the very small twin biometric dataset by using previous commercial matchers. The first elaborate study of the discrimination of identical twins was presented by Phillips et al., using the face recognition algorithm [89]. They equated their unalike face recognition technology to the identical twin dataset in their research. It contains images developed under facial illumination, pose, and facial expression. They also found that differentiating between identical twins is more promiscuous. Srinivas. N. et al. discussed in their study that identical twins can be distinguished by the utility of facial markings as a signature for biometrics [90].

**2.1.3.1.3 Cosmetic Based Face Recognition**

Another study on the assessment of facial beauty, but also limited their research related to the identification of facial makeup [91]. A method for testimonial was developed, employed candidate, and synthesis makeup on the face for makeup and hairstyle; they promote makeup to users and the highest beauty score is the result [92]. Their method creates invoking outcomes; however, certain drawbacks such as their technique can only impact a non-makeup face. The main work addresses facial makeup specifically in order to identify a face [93]. Two separate datasets were introduced: The YouTube Makeup dataset and the Virtual Makeup dataset, thus analyzing the efficiency of face recognition before and after makeup with three kinds of face recognition characteristics for instance, Gabor wavelets, the commercial Verilook Face Toolkit, and Local Binary Pattern. The technique defined the makeup that maintained a feature vector that contains shape, texture, and color, details on the face images [94].

To detect the face before and after applying makeup from the face pictures, the Support Vector Machine (SVM) and Canonical Correlation Analysis (CCA) classifier were employed [95], with the verification problem method carried out at [96]. Their method is capable of extracting the features from the face of make-up and non-make-up, based on correlation mapping, so they match the face. Before and after the makeup, the mapping of features acquired confronts images to trim the distance between the face images to be compared. By using the techniques of CCA, Partial Least Squares (PLS), and rCCA (regularized CCA) the mapping was defined.

Cunjian Chen et al. implemented a patch-based ensemble learning technique that used multiple subspaces created from after and before the makeup face image by sampling patches [97]. A unique automated makeup recognition and remover system was proposed [98]. To detect and obtain the application of cosmetics, a topically-filtered modest-rank dictionary learning method is employed. The probability of employing makeup for spoofing an individuality was investigated in [99]. They initially compile a series of facial images with the faces of cosmetically applied celebrities from the internet. With the two separate facial matchers, they obtain the effect of this and after applying makeup, more spoofed faces are better balanced. To demonstrate the identification of iris biometrics and cosmetics employed face by combining the images' shape, texture, and color type [100].

**2.1.3.2 Fingerprints and Its Security Issues**

The advent of fingerprints has addressed the requirement for a new effective method of proof identification concerning evidence. And the fingerprints of monozygous twins have been shown to be slightly different. Comparing the appropriate fingerprint with its digital image held ensures that the two identities of twins will certainly be separated. Even so, prospects for fingerprint counterfeiting still exist. However, scanners are most often equipped with advanced animosity detection, especially in the event of a safety risk, fraudsters face problems with fingerprint scanners. Often, because of the presence of an officer, it's almost inconceivable to fraud the fingerprint search. Naturally, following this measure does not result in an entirely flawless safeguard against inappropriate behavior. Nevertheless, with the addition of a fingerprint scan, the security standard has improved rapidly [62, 61].

In order to bypass the fingerprint reader here, it should be considered that finger fakes may be used by a potential intruder. One of the most important and daunting tasks in real-world environments is to secure the unsupervised and automated fingerprint recognition systems used for access control. Repudiation, coercion, contamination, and circumvention are the basic threats to a fingerprint recognition device [102]. Based on automatic fingerprint recognition, several techniques may be employed to gain unauthorized access to a device. Constructing an artificial fingerprint is one of the simplest ways to utilizing gum, soft silicone, similar substances, or plastic material if algorithm, hardware, and data transport attacks are ignored [102-104], where, compared to the real fingerprint, the fingerprint can be seen from a rubber stamp. There is no distinction between them for a very large number of sensors, i.e. The artificial fingerprint from the database is litigated and recognized as one particular enrolled employer. To stop possible attackers from displaying a false finger or, worse yet, to damage an individual in order to gain access, a component of liveness detection must be added to the system [102, 105, 106]. There are still a number of open concerns in fingerprint recognition, despite the increase in recognition accuracy under non-ideal conditions and recent developments in biometric template security [166].

The second issue that is sometimes overlooked is skin diseases and their effect on the identification of fingerprints [107-109]. Such skin diseases (in general, attacking the fingers or hands) can be divided into three major classes [107, 108]: skin discoloration, histopathological changes, a mixture of histopathological changes, and skin discoloration. Many sensors are based on physical concepts that do not permit a histopathological skin disorder to acquire a fingerprint. The second category involves decoloration of the skin, i.e. there is just a variation in the color of the skin, but the structure of the papillary lines remains the same. The majority of fingerprint acquisition sensors are not vulnerable to this form of skin disease. Both previous forms are merged in the last group. For almost all fingerprint sensors, this category is very complicated, since the combination of a structural change and a color change also leads to structural changes and color that cannot be recognized as a fingerprint for further processing.

### 2.1.4 SUMMARY

It is noticed that RFID chips are indefensible to requests for unauthorized approaches that provide access to sensitive data for the hacker [55]. This is a fact that is undermining the security and privacy of information. The RFID chip can be blocked by an RFID reader which can produce an exact Electronic Product Code (EPC), and sensitive tag information can be read and blocked by intruders who admittance to RFID readers or EPCs. Another concern is a person's facial image is used as a maximum-security component. There are opportunities for weakness in face recognition, however, because of poor matching photos of unknown faces from terrorism, illegal immigration, fake passport holders, and criminals [101]. Subsequently, incorrect authentication is another concern due to makeup or cosmetic skin, twins, and similar faces. There are some reasons why terrorists and illegal immigrants will clone and exploit a biometric passport. It is important to concentrate on safe biometric authentication, especially face recognition, to avoid such activities. The most traditional biometric data applied to authenticate an individual are facial images. It is a non-intrusive type of knowledge that, for covert identification, is publicly recognized.

From the above study, it is observed that numerous drawbacks in RFID tag security and privacy issues and chances of failure in face recognition. Also, very few works are available in the literature on the investigation of RFID and Face recognition. In this study, face recognition based on facial mark size recognition, hand geometry recognition, facial blemishes detection, cosmetic-based facial marks detection is introduced. Biometric data encrypted by applying the AES algorithm, SHA 256 algorithm, and SF algorithm. The encrypted data encoded into the QR code and HCC2D code are introduced to increase the privacy and security of an electronic passport.

# CHAPTER 3

# QR CODE BASED BIOMETRIC PASSPORT FOR NATIONAL SECURITY USING MULTI-BIOMETRICS AND ENCRYPTED BIOMETRIC DATA

## 3.1 INTRODUCTION

In 2004-2005, numerous nations implemented biometric-empowered passports in compliance with the guidelines of the ICAO to improve state-of-the-art safety and protection at the border control point [54]. By incorporating a wireless-based micro-chip with the MRTD standard, the biometric passport structured according to ICAO rules indicates that it will improve the biometric passport. For almost all persons, an electronic passport is required to be employed for global travel documents. It is designed by adhering a micro-chip known as the RFID to a biometric passport. This RFID accepts the data obtained by the electronic passport and allows the required biometric comprehensibility system installed at the border crossing point to be examined [11,12].

The distinctive protection and security develop with various threat challenges contained in the biometric passport were examined [32]. In their report, they explained that while the electronic passport substance can be perused within the short-range without the knowledge of a client, even the electronic passport booklet persists closed. The RFID tag found in the biometric passport is the primary reason for this problem. The researchers debate that the RFID tag prompts information leakage threat as it can furtively gather the data by eavesdropping or skimming. There are a few threats situations saw in the existing study which are identified with the issuance of travel documents [38]. Identity theft, data leakage risks, host listing, and monitoring are addressed as the difficulties that arise in issuing biometric passports [1]. The RFID chip is inclined to two types of attacks to be specific such as transport layers and network layers [40]. RFID chips can be attacked using spoofing, cloning, application layer, impersonation, eavesdropping, tag modification, and unauthorized tag reading. It clarifies four strategies of attacks employed by hackers to establish the security of a system that utilizes RFID-ISO/IEC 14443, for instance, cloning, skimming, eavesdropping, and relaying [41].

The primary challenges concerning risks are:
- Information leakage threats are disclosed by the RFID tag as it can covertly gather the data by "skimming" or "eavesdropping".
- The RFID tag has countless security emerges which prompts protection violation issues.

A new framework is introduced in this chapter where encrypted biometric passport information in the QR code will update the state-of-the-art security features during authentication as it is not an active component. This chapter focuses on multi-biometric authentication, which constitutes one of the least intrusive biometric frameworks which is based on the size of the facial marks detection and hand geometry recognition. Facial marks size as a signature of an individual for identification and it will enhance the authentication accuracy. Besides, the AES algorithm and the SHA-256 algorithm are employing for the encryption of personal data and biometric information. The encrypted data is encoded into the QR code that can be effectively printed on a biometric passport. Malicious tampering can be prevented by this method of encrypting and encoding into the QR code.

The proposed technique gives a more secure alternative utilizing the encrypted QR code when contrast with RFID. Data misfortune is not possible, since an active element is not an encrypted QR code. Hence, neither a remote repository is required nor a network connection is essential for retrieval of the biometric data which is encoded into the QR code. Nevertheless, the number of data is stored in the QR code relies on its memory storage capabilities, and its biometric data depends on the minimum data required to discriminate against a single individual. If the bearer data is permitted access to biometric information, the encrypted information encoded into the QR code from the electronic-passport can automatically be checked for authentication of the owner's credentials. In particular, the mechanism illustrated in the proposed system takes into account the faster verification process, without human intercession, of the claimed identity of the biometric passport owner. The verification is completed by comparing the biometric details of the person with the data stored in the database and with the biometric passport during the local scan. Thus a more reliable biometric passport can be upgraded to QR code by the encrypted biometric information while conducting the verification as it has stored encrypted data and is not an active feature. The proposed system would strengthen the biometric passport protection system worldwide.

The QR code has a biometric data storage capability and cannot be employed as an active part. In addition, it is cheaper and does not necessitate explicit data recovery equipment. QR codes are efficient and passive read-only which unchangeable and decodable by the individual devices.

This chapter summarizes the contributions and primary novelty as follows:

- Encryption of biometrics and encoded into the QR code.
- In order to improve the protection of an electronic passport, this chapter adopted multi-biometrics authentication such as facial mark size recognition and hand geometry recognition.
- Personal data and multi-biometric information are encrypted by employing the AES and SHA 256 algorithms.
- To endure passive and insure the information by encrypting it into the QR code, and the information that is not disclosed absent of its users' knowledge or consent must be protected.
- During authentication, encrypted biometric passport information in the QR code will update the state-of-the-art security features as it is not an active component.

## 3.2 PROPOSED FRAMEWORK

In this work, a framework is introduced to secure the biometric passport by encrypting the personal data, biometric information, and encoded into the QR code. Figure 3.1 represents the proposed framework for biometric passport data encryption and decryption. This technique acquires multi-biometric authentication such as identification of face and hand geometry recognition. Face recognition focuses on detecting the size of the facial marks and identifying the geometry of the hand, which is encrypted to solve the hazard challenges in the QR code. Firstly, using the Extended Profile - Local Binary Patterns (EP-LBP), a Canny edge detector, and the Scale Invariant Feature Transform (SIFT) algorithm with Image File Information (IMFINFO) process, the facial mark size recognition is initially achieved. Secondly, by using the Active Shape Model (ASM) into Active Appearance Model (AAM) to follow the hand and infusion the hand geometry characteristics for verification and identification, hand geometry recognition is achieved. Thirdly, the encrypted biometric passport information that is publicly accessible is encoded into the QR code and inserted into the electronic passport to improve protection. The major components of the proposed framework is given in the below diagram.

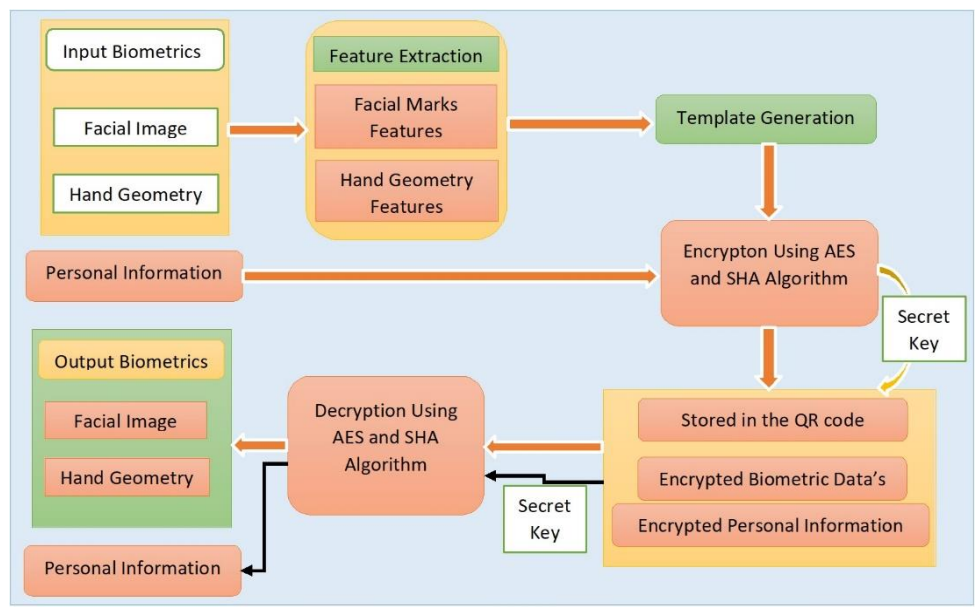

**Figure 3.1: Framework for the encryption and decryption of biometric passport information**

By employing AES and SHA-256 algorithms, the biometric passport data encryption is achieved. The primary objective of this study is to passively withstand and protect the information contained in the QR code and data cannot be identified without a biometric passport owner being recognized or granted.

### 3.3  BIOMETRIC PASSPORT AUTHENTICATION

The inspection procedure has been done into two sections to distinguish an individual. The first section is for distinguishing proof dependent on 1: n and the other is 1:1 verification, which relies upon hand geometry recognition and face detection. Subsequently, the second component is based on 1:1 confirmation for an individual's approval and both can be carried out at the same time.

### 3.4  FACE RECOGNITION METHODS

In facial recognition, biometric features are used to enhance safety in order to verify the face of an individual. It is a state of the art secure technique that utilizations publicly acceptable frameworks for distinguishing proof application. However, due to poor matching images of new faces/unknown faces, there are various possibilities of failure in face recognition from illicit migration, terrorism, fake passport holders, and criminals [101]. This issue can be comprehended by giving a superior facial recognition technique. Each individual has remarkable anatomy of his face. The face carried out wrinkles, marks, moles, etc. that could not effectively be modified. It is possible to identify a person with these facial marks and to forestall the masquerades. This chapter, therefore, focuses on identifying the size of the facial mark for a biometric passport to improve national security with a border control system.

### 3.5  FACIAL MARKS ON THE FACE

Facial marks are generally located in the face of salient regions that can be available in the forehead, cheek, and chin. This thesis introduced an edge detection algorithm to recognize the size of facial marks. The facial feature extraction technique is employed to distinguish various classes of facial marks and examines marks classifications that are available on the face. It recognizes all categories of marks present on the face, for example, a scar, spot, mole, freckle, brightening, acne, pockmarks, dark spot skin, and so on. Along with areas of the eyes, mouth, nose, and eyebrows, a feature extractor technique may also produce some false positive outcomes. We subsequently dismiss unwanted features in the face image.

## 3.6   THE DETECTION OF FACIAL MARKS SIZE TECHNIQUE

To implement the facial mark size detection algorithm, the proposed method uses the EP-LBP algorithm [112], the Canny edge detection algorithm [113], and the SIFT algorithm [114]. The facial mark size detection with sequencing steps are shown in Figure 3.2. The systems for the recognition of facial mark size detection techniques are delineated beneath.

### 3.6.1   Facial Features Detection

The proposed technique for recognizing the facial mark depends on the automatic recognition of the face marks in this study. The comprehensive flow of facial mark detection is shown in Figure 3.2. The recognition of the size of facial marks is attained with the following steps:

1. Using the Extended Profile - Local Binary Patterns algorithm to recognize ninety points of the landmark on the face which characterize the shape of the face image and the important facial characteristics are distinguished. The undesirable facial highlights are deducted from the face, for instance, the mouth, eyes, nose, and eyebrows.

2. The masking process is carried out using the Canny Edge detector algorithm.

3. Using the SIFT algorithm, the facial marks are identified and the size is calculated using the IMFINFO technique.

Using the essential facial feature recognition procedure, the size of the facial mark is extracted and delimitates the procedure of location of essential facial features on mask generation and detection of marks. The algorithm EP-LBP [112] is used to differentiate important features of face, for instance, nose, eyes, mouth, and eyebrows from the face. Depending on the training database, when an EP-LBP is trained, it frames another new contour to match the new face picture. Compared to the facial component's contour, it recognizes 90 landmark points and recognizes the shape parameters. Subsequently, each model's profile is like the pre-learned profile, to promote facial recognition.

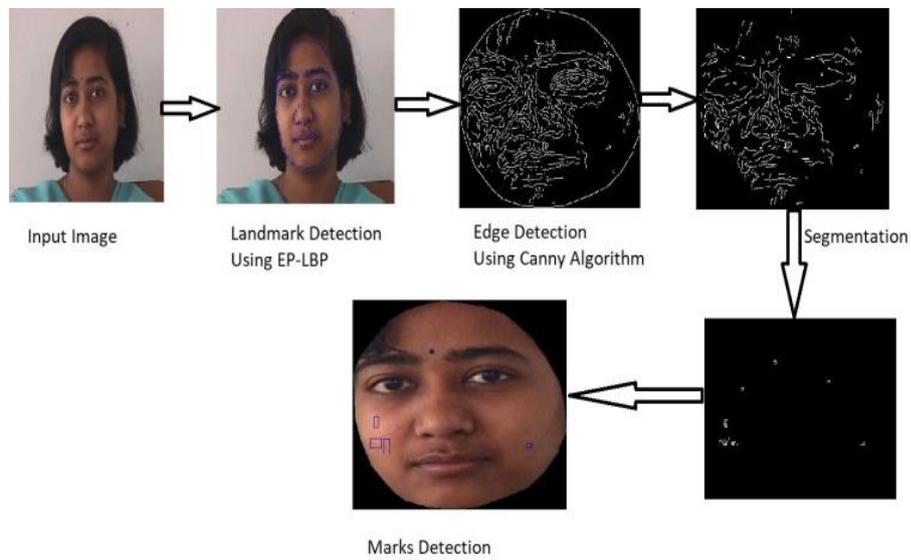

**Figure 3.2: Diagram of the detection process of facial mark size**

### 3.6.2 Mask Construction and Designing Mean Shape

To locate the landmarks of the facial image, 90 landmarks points to construct facial marks size recognition simpler used EP-LBP [112] technique. The scale and rotation are uniforms to identify the size of the facial mark. In this manner, each face picture is thus mapped in the mean form and $S_i$ is conceived, where $I = 1, 2, 3, 4,.....$ N functions as a state with 90 points of landmarks reference in the database of each N facial image. It is determined by using the following condition,

$$S_\mu = \sum_{i=1}^{N} S_i \qquad (3.1)$$

Consider, $S_i$ is an individual gallery face image by applying the Barycentric coordinate-based algorithm and $S_\mu$ is the mean form [115]. A standardized mask has been created and it is also an applicant's unique mask to restrict the identification of false-positive facial marks from the facial image. The facial mask may carry out unwanted facial features, hence it takes out the false positive to recognize the facial marks precisely. By using the Canny edge detection algorithm [113], we structure a user-specific mask and obtain the exact edges on the facial image.

### 3.6.3 Facial Marks Size Detection

In the isolated and prominent location of the face, facial marks are commonly situated concerning a measurable variety of severity. The Laplacian of Gaussian (LoG) [116] has chosen to define as a second-order derivative the edge projected in [76]. The LoG technique filters the edges of the images to deduct the user-specific mask $M_s$ with a progression of threshold value $t_i$, where *l = 1,2, .... k* by reducing order. As a result, the threshold value $t_i$ is effectively employed to the sum of the relevant segment's subsequent estimates, which is more prominent than $C_c$ (predetermined value). If $C_c$ = *10*, the identification of facial marks is most intense at that point, and the size of the facial marks considered to be three pixels (height and width) are taken away to extinguish the noisy facial edges produced by false-positive regions. Therefore, with a circle or a bounding box at its place, the facial mark size is seen and the facial mark size is tested. In this analysis, the approximate facial mark size and distance were determined using the Bhattacharyya algorithm [117].

### 3.6.4 Facial Marks Size Representation

Two methods are used in this chapter to constitute the size of the facial marks:

1. The initial technique is to classify the size of a facial mark by obtaining a quantity of the intensity of the pixel having a position with the bounding box. In the subsequent methodology, by using the SIFT algorithm, each size of the marks is recognized in an image. Facial marks might be available in different places in the face. A few facial marks, in particular, scars, moles, and so on can occur in a similar facial area. If a mark on the forehead that is accessible and located on the chin is not conceived fallaciously, it is significant. All sorts of facial markings on the face are substantial to note. The coordinate points x and y of the bounding box were used to constitute a facial mark and the qualities (0,1) were normalized individually by the image width and height partitions.

2. The second approach to detect facial marks is described by the SIFT technique's vector appreciation to represent facial marks and facial features. Therefore, to provide safe confirmation, it evaluated the size of the facial mark. The extraction of facial features makes, it possible to determine the size of each mark; the quantity of pixels is employed to determine the area. The facial marks detection is estimated using the IMFINFO technique. Using a histogram, the size dispersion is shown and a distinctive statistical value is estimated (median, mean, and standard deviation) in figure 3.3. The portrayal of the size of facial marks would increase the relation among detection of facial marks size and the accuracy of recognition. In Figure 3.4, the histogram of the facial marks present on the face is shown.

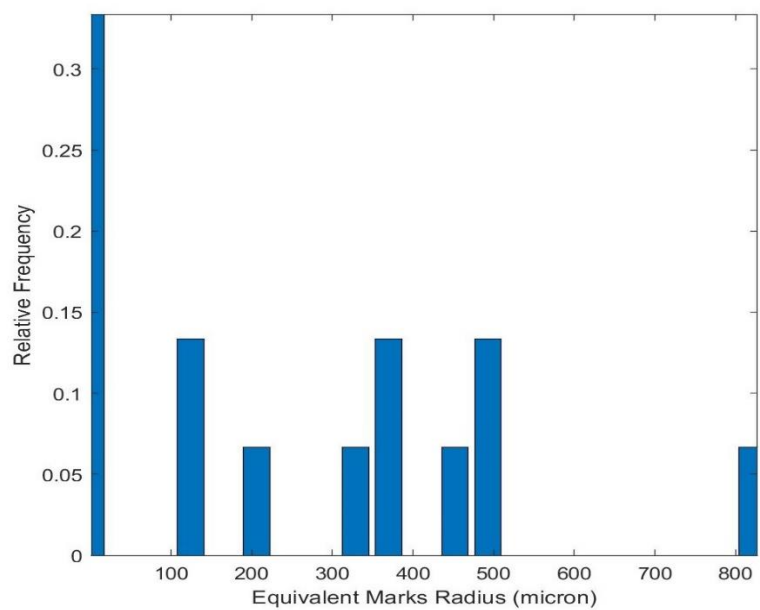

**Figure 3.3: Distribution of facial marks size**

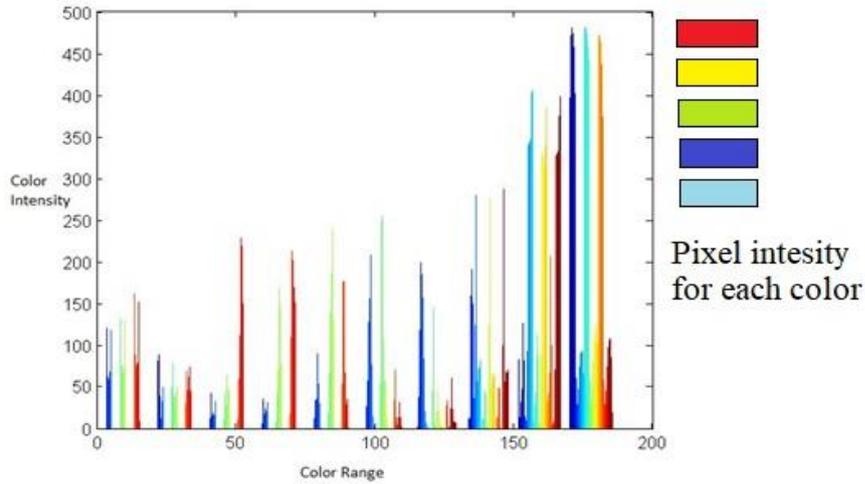

**Figure 3.4: Histogram of the facial marks in the color face image**

### 3.6.5 Matching the Facial Marks

Conceived the two facial images $I_1$ and $I_2$, the set of their facial marks detection are $N_1$ and $N_2$. The recognized marks on the face and the similarity of the two facial pictures $I_1$ and $I_2$, we call Facial Marks Matching (*FMM*) are determined by the employ of the specified condition (3.2).

$$FMM = \frac{\sum_{i=0}^{|N_1|} \min D(n_i, n_j)}{|N_1|}, \quad \forall n_j \in N_2 \mid (x_j, y_j) \in R_i \qquad (3.2)$$

If each facial mark is conceived by $n_j \in N_2$ it's pertaining to central coordinate points are $x_j$ and $y_j$. Each facial mark is $n_i \in N_1$, and $R_i$ is a rectangular area built over its central $I_2$ coordinates as part of the enormous matching of the marks $n_j \in N_2$, applied in $R_i$, which is considered throughout the matching procedure of the facial mark. Hence, D is built to calculate the length among facial marks, where the Bhattacharyya distance measurement algorithm [117] calculates it and obtains the best consequence. In this study, the proposed method proves the utmost outcomes when contrasting and getting histograms during the performance of the observational procedure.

## 3.7 RESULTS AND DISCUSSIONS

The experimentation is conducted on the Indian Institute of Technology, Kanpur (IITK) face database [118], and the Fundacao Educacional Inaciana (FEI) database [119] to evaluate the projected facial mark size detection algorithm. In this experiment, the Indian facial database with 40 participants, accumulated 500 face pictures of IITK. The Brazilian face database subsequently collected 2800 face pictures from the FEI face database of 200 subjects. Instead of a splendid homogeneous background, the Indian face database was selected; the face images are in the JPEG format, with 640x480 pixels for each face image size. All face images are set up in two genders which are male and female. A homogeneous white background is colored in the FEI database face pictures. The genuine size of the image of each face is 640x480 pixels. All FEI database essences are age-dependent on Brazilians between 19 to 40 years of age.

To execute the proposed algorithm, it is established manually annotated facial marks, whereas a normal of 5 to 6 facial marks were identified. Thus, birthmarks, moles, pockmarks, grains, and freckles are facial marks. Figure 3.5(a) displays the scale of the facial mark recognition from the IITK dataset and Figure 3.5(b) displays the facial mark size recognition from the FEI dataset.

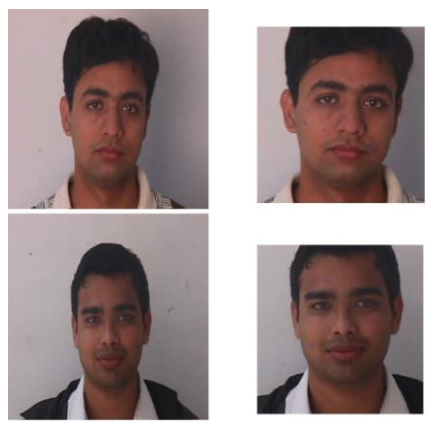

**Figure 3.5: (a) The facial marks size recognition**

**from the IITK dataset**

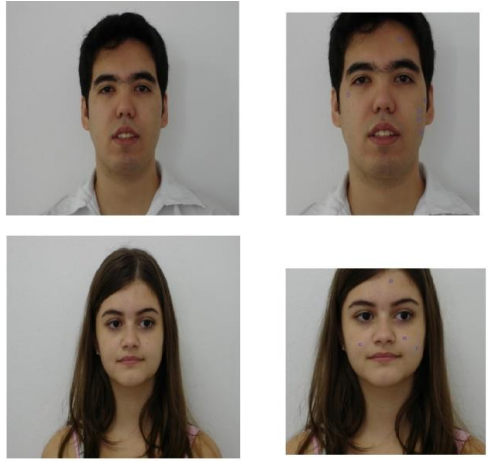

**Figure 3.5: (b) The facial marks size recognition from the FEI dataset**

In the face image, some of the sizes of the facial markings are then not stabilized and problems occur during face recognition, such as acne, zits, or pimples, and so on. The long-lasting facial markings are calculated on the facial image to improve the facial recognition accuracy; the facial marks are scar, moles, pockmark, freckle, dark spot skin, birthmarks, whitening, and so on. In order to make the correct identification, we calculate the facial mark size and evaluated the facial mark distance.

This ameliorates the recognition and identification of a specific individual. In facial mark size recognition, the proposed method attained 92 percent (IITK database) and 93 percent (FEI database) accuracy. The proposed algorithm is executed employing Matlab (R2017a). Thus the validation and evaluation of the detection of the size of facial marks are shown in the accompanying sections.

### 3.7.1 Evaluation of the Facial Marks Size Detection

The proposed technique of recognition of facial marks (named C-SIFT) prompts the formalization of the accuracy of the mark's recognition. To build up the similarity among detected facial marks and annotated facial marks, it sums up the standard measurement to evaluate face recognition of the issue of facial marks identification [120]. Conceive, an image *I* and *A* is a facial mark annotated on the face, $n_i$ is the recognized facial marks were considered as true if $\exists\, n_a \in A$.

Therefore, we calculate $\quad \dfrac{area(n_i) \cap area(n_a)}{area(n_i) \cup area(n_a)} \geq t_0 \quad$ (3.3)

The threshold value $t_0$ was empirically obtained in the form of 0.4.

In the existing research, it was unable to obtain any implementation of the facial mark size recognition techniques. Thus, a few of the techniques based on facial mark detection algorithms were chosen in this experiment to contrast with the outcome of the proposed facial mark size detection algorithm on a similar database of face images. The proposed facial marks detection technique is utilized by altering a few of the pipelines from the previous techniques, for instance, canny edge detector with SIFT (We call C_SIFT) rather than Sobel operator with LoG (We call S_LoG) [76]. The present technique is contrasted with antecedent published work Canny with SURF [82] (referred to hereinafter as C-SURF) and Canny with HOG [121] (referred to hereinafter as C_HOG). Aside from that contrasted with the proposed technique outcomes with Viola and Jones object detector [122] (referred to hereinafter as V_J). The outcomes for the FEI database and IITK database have appeared in the following table 3.1 and table 3.2.

Table 3.1  Compared to the proposed algorithm C_SIFT and other detection techniques using the IITK dataset

| Algorithm | Precision (%) | Recall (%) |
|---|---|---|
| C_SIFT | 75.33 | 62.23 |
| C_SURF | 71.34 | 57.01 |
| V_J | 29.01 | 16.11 |
| C_HOG | 71.99 | 59.09 |
| S_LoG | 27.89 | 53.87 |

For face recognition, this technique was proposed, and it can be formulated to detect the object's variety in the face image. To identify the facial marks, a series of 100 positive sample images with facial marks and 300 negative sample images without a mark was trained on the skin area. All of the above techniques were evaluated with various configurations of parameters and a superior result was found in the tested version in each case.

Table 3.2 Proposed technique C_SIFT compared with other detection methods using the FEI dataset

| Algorithm | Precision (%) | Recall (%) |
|---|---|---|
| C_SIFT | 73.65 | 59.22 |
| C_SURF | 58.84 | 41.11 |
| V_J | 21.09 | 11.21 |
| C_HOG | 69.91 | 51.12 |
| S_LoG | 22.19 | 19.34 |

This analysis shows precise results for recall and precision for the proposed algorithm of facial mark detection (C_SIFT) than the C_SURF, S_LoG, C_HOG, and V_J algorithms. Thus the proposed algorithm for detecting facial marks size can be more reliable (recall) and fewer false-positive marks (precision).

### 3.7.2 Validation of Face Verification

False Recognition Rate (FRR) and False Acceptance Rate (FAR) are estimated when the Operation Point (OP) is designed to be 0.1 percent of FAR, we call it as (FM_OPFRR) to evaluate the accuracy of the facial marks detection algorithm.

Therefore, by using a facial mark matching technique, the verification test is executed on the IITK dataset. To contrast the facial mark recognition algorithms from the existing techniques with respect to verification accuracy, the experiment from this face database was carried out in this study because it has more prominent facial mark variability. Depending on manual annotation (M_Annotation), this analysis is carried out and the outcomes are equalized on the face image with manually annotated facial marks. Experimental findings are presented for FM_OPFRR and EER in table 3.3 and the M_Annotation outcomes have been included. Therefore, in the outcome analysis, as compared to previous facial mark detector algorithms such as C_HOG, V_J, and S_LoG, the proposed facial mark detector has obtained better results. Analysis of the importance of manual annotations as a result of its fewer error rates than automated techniques acquired.

Table 3.3 The proposed algorithm C_SIFT compared with other algorithms using the IITK dataset

| Algorithm | EER(%) | FM_OPFRR(%) |
|---|---|---|
| M_Mannual | 30.04 | 99.13 |
| C_SIFT | 31.34 | 99.31 |
| V_J | 33.01 | 99.52 |
| C_HOG | 41.99 | 99.81 |
| S_LoG | 37.89 | 99.69 |

To measure the facial marks detection performance in face verification and face identification, it aggregated with the proposed facial mark detection algorithm with extremely well known two methods (i) EP-LBP (Extended Profile – Local Binary Pattern) operator [112] and (ii) Fisher Vector (FV) Faces [123].

The experimentation is carried on one vs. one for verification evaluation and applied on both FEI and IITK databases. The automatic detection (FMM_Auto) is calculated and contrasted with the results of the EP-LBP and FV technique, including their compounding and facial mark matching. The experimentations are carried out under comparable circumstances by employing the manual annotation (M_Annotation) of marks on the face. It will have the advantages of the face verification experiment that is expected. The outcomes dependent on FM_OPFRR and EER are performed utilizing the IITK database and FEI database severally. Tables 3.4 and 3.5 showed the outcomes.

**Table 3.4** Comparison of proposed face verification experiment and its combinations using the IITK dataset

| Algorithm | EER(%) | FM_OPFRR(%) |
|---|---|---|
| FMM_Auto | 30.04 | 99.13 |
| M_Annotation | 21.34 | 98.11 |
| EP-LBP | 3..01 | 7.11 |
| EP-LBP + FMM_Auto | 1.67 | 6.09 |
| EP-LBP + M_Annotation | 1.59 | 6.01 |
| FV | 1.6 | 6.11 |
| FV + FMM_Auto | 1.12 | 4.22 |
| FV + M_Annotation | 0.99 | 3.12 |

From the estimates of the initial two rows of tables 3.4 and 3.5, It is noted that facial marks are inadequate for inclusion in the process of face verification. However, the accuracy of face verification is increased when they are combined with previous traditional FV and EP-LBP face recognition techniques. The progress in the EER is significant and shown in table 3.4. In this experiment, FV + M_Annotation increases accuracy as compared with FV alone. The error made in the automated recognition of facial marks does not refer to the last results because of the merger with the EP-LBP technique. EP_LBP + M_Annotation and EP_LBP + FMM_Auto were modest improvements, but with the mixing with FV technique, there has been confirmed improvement in the detection technique for facial marks. Fewer changes in the identification of automatic facial marks are observed in table 3.5.

Other interventions are applied to decrease the false detection, and the selection of the weights $W_{fmm}$ and $W_{fr}$ are the products of the marks equating execution for the mix of algorithms. It is apparent that the effect of detecting facial marks is less disclosing. Notwithstanding, with higher estimates of $W_{fr}$ (0.8 and 0.9) and lower estimates of $W_{fmm}$ (0.1 and 0.2), the blend accomplishes the most beneficial outcomes.

**Table 3.5** Comparison of proposed face verification experiment and its combinations using the FEI dataset

| Algorithm | EER(%) | FM_OPFRR(%) |
|---|---|---|
| FMM_Auto | 28.14 | 99.33 |
| M_Annotation | 20.34 | 91.01 |
| EP-LBP | 4.01 | 26.11 |
| EP-LBP + FMM_Auto | 1.99 | 8.09 |
| EP-LBP + M_Annotation | 1.89 | 6.17 |
| FV | 2 | 17.11 |
| FV + FMM_Auto | 1.34 | 5.33 |
| FV + M_Annotation | 1.23 | 4.89 |

### 3.7.3 Validation of Face Identification

Using a 5-fold cross-validation technique, the critical distinguishing proof procedure was led by inevitably using one face picture of each individual from the gallery and the remainder of them as an examine. The IITK database is utilized for this validation in this analysis and the findings of face matching are compared with the automated recognition of marks on the face and the manual detection of facial marks. Hence, with the combination of the FV method, the most valuable results are obtained in the previous section. The attained results with respect to the Average Recognition Rate at Rank1 (ARR_R1) and Rank 5 (ARR_R5) have appeared in table 3.6.

**Table 3.6** Proposed method and combination of FV algorithm in an identification analysis using the IITK dataset

| Algorithm | ARR_R1 | ARR_R5 |
|---|---|---|
| FMM_Auto | 24.11 | 46.99 |
| M_Annotation | 39.91 | 57.45 |
| FV | 88.01 | 96.12 |
| FV+FMM_Auto | 90.05 | 98.1 |
| FV+M_Annotation | 92 | 98.2 |

The above table demonstrates that the utilization of facial marks alone doesn't accomplish a higher rate of facial recognition, but it improves the results of facial identification by mixing with a face recognition method. Here, conceived Rank1 to Rank10 by utilizing only the FV method and obtained 88.01 percent of the ARR, as matching facial marks are designed for differences in both manual detection and automatic detection, this extends the ARR. The experimental outcome until Rank 10 is shown in Figure 3.6.

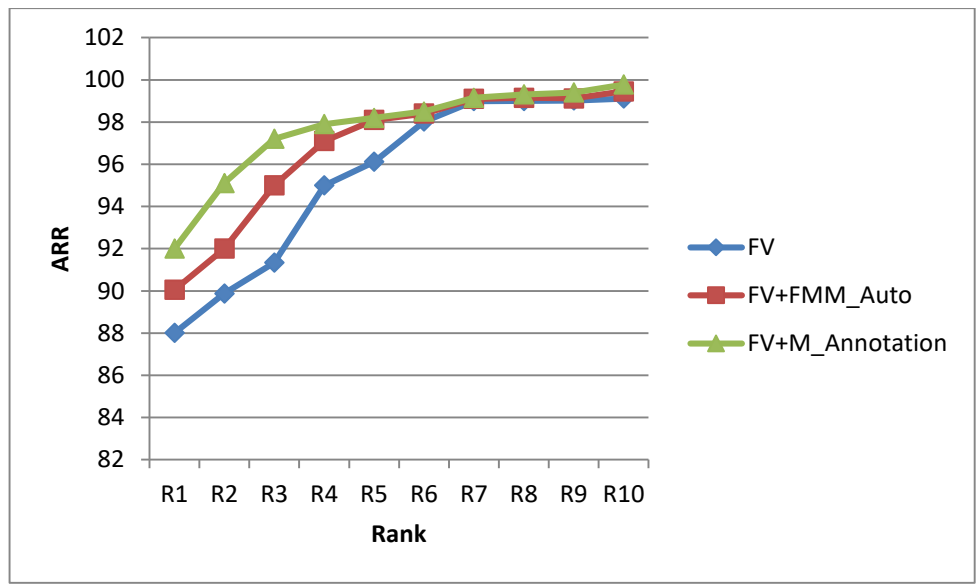

**Figure 3.6: The cumulative average recognition rate curve (in percent)**

The above graph shows the cumulative average recognition rate curve in percentage. FV technique alone, a combination of FV with automatic detection, and a combination of FV with manual annotation are shown in the graph. It improved the results when combined with FV and manual annotation. As compared to the results of R1, R10 gives a more proficient outcome.

### 3.8 HAND GEOMETRY DETECTION METHOD

It was developed in the 1980s and is the most prospicient enforced in a biometric system. Such systems are broadly incorporated into the application, particularly in terms of consolidation potential and public recognition. The main approach to applying hand geometry is to measure the characteristics of the hand and register them, such as the length, width, thickness, and surface area of an individual hand. To extract and track features of hand geometry for identification and authentication in this experiment employed the ASM into AAM algorithm [124]. Hence, by employing ASM in the AAM technique to acknowledge the area of the landmark on the hand to eliminate the landmark point, and the distance characteristics are taken out to differentiate proof. The execution of ASM into AAM fitting method accelerates and improves strong efficiency.

### 3.8.1 Enrolment

The procedure of adding clients to the database is known as enrolment. Hand geometry pictures of 120 and 30 subjects have been accumulated in this experiment. The images incorporate either the left or right hand of the clients in the dataset. A scanner is used to obtain the hand image and it captures colored hand geometry. Furthermore, the dataset carried out data on the age and sex of individuals.

### 3.8.2 Features Selection for the Candidate

The examination was carried out depending on the implementation of ASM into AAM [124] technique and as a valuable hand feature characterizes 14 distances on hand geometry. The characteristics are the length of the palm in width and finger length. The individual function is defined in this study depending on ASM into AAM of the model. Henceforth, it has now been presented ASM into AAM algorithm, and to represent the features is precisely defined, and among other model fitting algorithms, it is outstanding.

### 3.8.3 Feature Selection

The feature extraction procedure is performed depend on the technique [125], the 14 distance features on which is concerned are steady worth. In accordance with these requirements, the characteristics are not sensitive to hand geometry forms. The standard deviation of imparted characteristics considers feasible details for dissimilar hand geometry images of a comparative subject. A medium-large standard deviation is positive for a massive distance function used. It may, in any case, exclude a small measure of distance. Using the function quality metric, it is carried out,

$$rms = \frac{1/n \sum_{j=0}^{n} \mu^n j(f_i(Sub_j))}{1/n \sum_{j=0}^{n} \sigma^n j(f_i(Sub_j))} \tag{3.4}$$

where $f_i(Sub_j)$ is the i-th feature of subject j.

## 3.9 RESULTS AND DISCUSSIONS

In this analysis, the experimentation was conducted on the basis of performance evaluation, which considers two important segments of techniques that are ASM into AAM fitting algorithm and the categorization of the hand dataset. The first elaboration in this chapter focuses on the outcomes received by using ASM in the AAM fitting model development experimentation, and the resulting elaboration focuses on evaluating a classifier's output by using the features of matching ASM into the AAM model.

### 3.9.1 Model Fitting

ASM into AAM algorithm is used to construct the model in the hand pictures dataset and labels in the hand dataset. It used the refitting technique to limit the noise in the manually specified labels [126]. The ASM into AAM fitting technique is presented and operates until convergence with the ground truth marks. To build another model, the vertex region of the fitted mesh is used as ground truth here. It is shown that on-hand data is used to refit and boost refitting accuracy [126]. The ground truth pattern and similarity alter arguments that initialize the ASM into the AAM fitting algorithm are arbitrarily disordered to calculate the algorithm's accuracy. At that point, using Euclidean distance measurement, it measures the distance between the ground truth vertex region and the equipped vertex region. The instance of a hand picture from the database is shown in Figure 3.7.

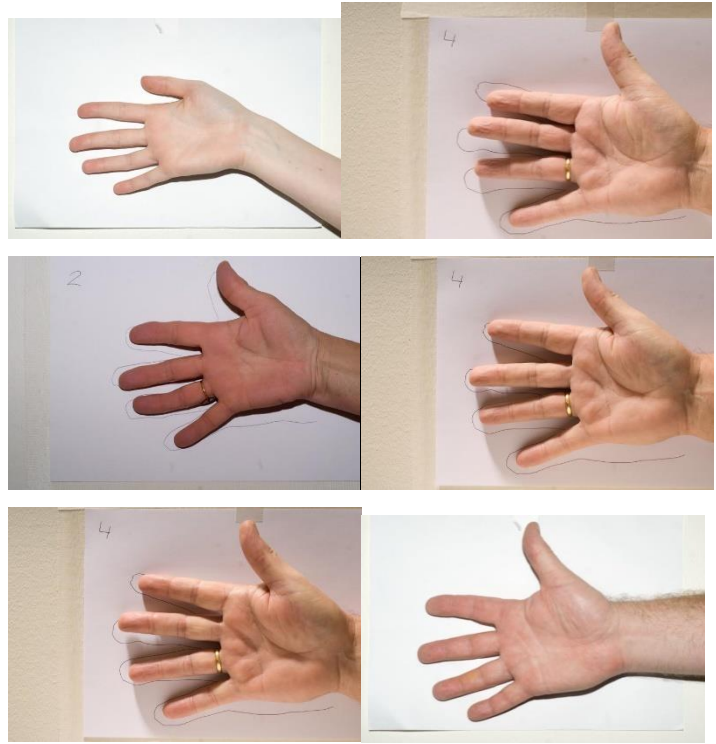

**Figure 3.7: Example of hand pictures from the database**

### 3.9.2 Classification Accuracy

The test employing to execute the experiment with the cross-validation of 10 folds to assess the accuracy of hand geometry recognition, as the amount of folds is constant to 10 for constituting increasing the standard. Witten et. al. [127] illustrated that 10 – fold is the perfect amount of folds necessitated to obtain a more beneficial error rate. In their work, they have subsequently demonstrated an encompassing test on multiple datasets utilizing various classifiers. Therefore, in this experiment, the technique of performance was compared by using the accuracy of the classification. This evaluates the total amount of cases tested and the number of cases ordered correctly by each classifier as shown in table 3.7 beneath.

**Table 3.7 Comparison of the proposed ASM into AAM technology and the AAM algorithm**

| Methods | Instances(Total hand geometry) | Correct classification | Accuracy (%) |
|---|---|---|---|
| AAM | 120 | 112 | 93.33 |
| ASM into AAM | 120 | 116 | 96.67 |

The proposed method is the comparison with the existing method Gross et al. [125] AAM fitting algorithm. Therefore, since the integration of ASM into the AAM algorithm of fitting provides more agility and accuracy, the proposed algorithm has achieved precise precision. Intel Core i7, 6600U processor, 16 GB RAM, 3.4GHz CPU clock speed, and Windows 10 pro were analyzed. The hand geometry results are presented in Figure 3.8.

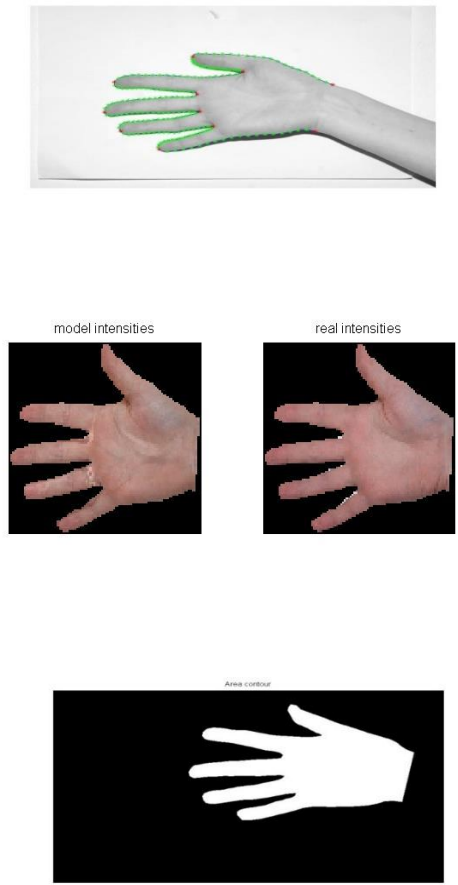

**Figure 3.8: The output of the hand geometry sample**

## 3.10 ENCRYPTED BIOMETRIC DATA AND QR CODE

The QR code is a type of barcode matrix trademark, also referred to as a two-dimensional barcode (2-D). It was first established for the automative industry in Japan. A machine-discernible optical imprint is an institutionalized standardized barcode and the QR code details on the matter that is accomplished [128, 129]. It contains square dots arranged on a white background in a square pattern, and four specific types such as modes of information numeric, alphanumeric, and byte/binary can be generated for the encoded information. It is also noted that a semiconductor image sensor digitalizes images depends on a two-dimensional image and is then carefully examined by a programmed device. The e-passport information is encrypted in this research work and stored into the QR code. Figure 3.10 shows the biometric-passport data encoded in the QR code.

### 3.10.1 Biometric Encryption

Biometric encryption uses an individual's physical characteristics as a means of code or decryption, and of granting or refusing access to a Personal Computer (PC) device. The need to authorize the predilection and proposal of security and data protection experts for the use of biometrics to affirm individuality rather than the goal of recognition alone is genuinely addressed by biometric encoding methods. Because of its inconsistency, biometric information in itself cannot be represented as a cryptographic key. However, the quantity of information that includes biometric information is enormous. The technique immovably secures biometric data with a cryptographic key in order to prevent the key or biometric data from being retrieved from the stored prototype. Only if a particular live biometric test is provided for verification is it conceivable to replicate the key.

Throughout enrollment, the digital key is created indiscriminately; subsequently, the client is not aware of the key. The biometric key is purely autonomous and for the most part, it can be modified. In addition, a biometric specification is evolved and the key is hold by a biometric encryption algorithm to the efficient and safe development of a biometric template and is often referred to as a private template. With the biometric encryption template, the biometric key is encrypted at a fundamental stage, which provides excellent faith in protection. In the QR code, the produced biometric encryption is stored.

When the client presents his/her ongoing biometric sample for the validation or verification section, the legitimate biometric encryption template is verified. This helps biometric encryption as a decryption key to recall the comparable key and the biometric aids. The biometric will be discarded at the end of the validation. The template cannot recover the key because of an attacker whose biometric information is different. Because each time the biometric template is different, this method of decryption and encryption is now muzzy, and the distinctive key encryption is created employing conventional cryptography. Then the digital key is retrieved and can be used as the base for any logical or physical application. Hence, biometric encryption for biometric key administration is a productive, secure, and protective process. We proposed an image encryption method in this chapter by applying the AES and SHA-256 algorithms that are stored in the QR code. This approach would improve the protection of biometric passport data.

**3.10.2 Encrypted Data into the QR Code:**

Using the AES algorithm [130, 131] and the SHA-256, the electronic passport demographic and biometric information is encrypted and the encrypted information is encoded into the QR code. AES algorithm which is otherwise known as the Rijndael cipher block and it was introduced by Joan Daemen and Vincent Rijmen. It is often referred to as an asymmetric key algorithm, as the same key is employed for data decryption and encryption. With distinct block sizes and key sizes, it is a kind of cipher. Therefore, it is possible to access a block size of 128 bits and three kinds of key sizes, which are 128 bit, 192 bit, and 256-bit files. The key size and plain text are chosen separately, the key size and plaintext size are selected by the total amount of rounds to be performed. The amount of keys sizes and plaintext decides the number of rounds, a 128-bit key has a minimum round of 10, and a 256-bit key has up to 14 rounds. In this chapter, it is encrypted the demographic information and biometric data carried out inside the front page of the biometric passport and encoded into QR code. In this study, the facial image concealed from the interloper [132] was encrypted. Furthermore, two encrypted QR codes are employed, one for demographic data and the other for the encryption of biometric information. The demographic information and biometric information encryption approach using the algorithms AES and SHA 256 is given below:

### 3.10.2.1 Demographic Data Encryption Using AES

By employing the AES algorithm encrypted the demographic information such as name, gender, date of birth, address, and so on. This information is the same content from the passport booklet and encrypted to enhance security.

**Encryption Algorithm:**
  (I) Execute the procedure below (Once)
   a. Elaborate the keys to 16 bytes.
   b. Do it once in the 16-byte plain text initialization process.
   c. The condition of XOR with the main block.
  (II) For the single round
   a. Employing the S-BOX in plain text.
   b. Circumvolve the K row of the block in plaintext with K bytes.
   c. Conduct the protocol for the Mix Column.
   d. The condition of XOR with the main block.

**Decryption Algorithm:**

It is the reverse process of encryption.

### 3.10.2.2 Biometric Information Encryption Employing AES

In this chapter, the biometric data encryption is obtained using the AES algorithm and the SHA-256 technique is discussed. The biometric information is the cipher; it was executed using the key-based biometric encryption algorithm. The AES algorithm is used to create a key for biometric information encryption.

### 3.10.2.2.1 Encryption Phase

Two inputs are required for the biometric encryption. Initially, conceive the input biometric $I_1$ which is encrypted. Secondly, $k_1$ is the hidden key, and by using the SHA-256 algorithm, $k_1$ is altered to $SHk_1$. Consequently, $I_1$ is encoded in string $B_1$ of Base64. Consequently, the AES-256 algorithm is fed into $B_1$ and $SHk_1$ to produce the ciphertext $C_t$.

**Encryption Algorithm (with N=2):**

1. Use the biometric information input and encode it utilizing the base64 standard.

2. Using the SHA-256 algorithm to generate the key, peruse the key file, and begin the biometric encryption by applying the AES-256-bit key.

3. By applying the base 64 encoded content and hash generated in step 1 and step 2, respectively, biometric information is encrypted.

4. Generate an additional biometric $I_2$ size ($S_1$, $S_2$) with $P_i$ pixel data.

   Assume, i. $S_1$ – countenance of characters of document for the key (Default: 255).

   ii. $S_2$ – Character amount of the key document.

   iii. $P_i$ – The data for the pixel to occupy (Default: 0).

5. Each row $I_R$ of the image repetition height:

   i. Obtain $J_A$ to be an $I_R^{th}$ character's *ASCII* code in the document of the key.

   ii. Initially, fulfill the $J_A$ pixels in the $I_R^{th}$ row with black color from the image. From $I_R$. $I_2$ [ $I_R$ ][ $J_A$ ] = 0 for each $I_R$, $J_A$ in $S_1$, $S_2$ with the final objective of $J_A < ASCII(key[I_R])$.

6. Create $N (= 2)$ pictures ($S$, $T$) of an alike size ($S_1$, $S_2$) and pixel data to an extent such that,

   i. For the 1st image $S$, pixel data is randomly generated and inclines to either be $0$ (black) or $1$ (white) $I_R$. $S [ I_R ][ J_A ] = $ random($0, 1$).

ii. 2nd image pixel data $T[I_R][J_A]$ is characterized such that $I_R$. $T[I_R][J_A] = S[I_R][J_A]$ xor $I_2[I_R][J_A]$ for each $I_R, J_A$ in $(S_1, S_2)$.

7. The output of the encrypted encoding is $C_t$, S, and T biometric data severally.

### 3.10.2.2.2 Decryption Phase

The biometric decoding stage consists of two biometric inputs; the primary ciphertext is the $C_t$ to decode the biometric data. The array of hidden key $k_2$ shares is given in the second. Therefore, using the master key $K_3$, the biometric data is constructed. In addition, by using ASCII, $k_2$ is deciphered by $k_1$. Therefore, using the SHA-256 algorithm, $k_1$ is altered to $SHk_1$. To create the Base64 encryption of the biometric data $B_1$, the $SHk_1$ and $C_t$ are fed into the AES-256 decoding. Additional $B_1$ is altered to the definitive outcome $I_1$.

**Decryption Algorithm (with N = 2):**

1. Calculate the ciphertext input $C_t$ from the biometric data.
2. $K_2$ is the key of the biometric data; it is necessary to load $K_3$ from the input $K_2$.
3. Generate other biometric data $I_2K_1$ of size $(S_1, S_2)$ such as $K_2, K_3$ with the final objective that

   i. $I_2K_1[I_R][J_A] = K_2[I_R][J_A]$ xor $K_3[I_R][J_A]$ for each $I_R, J_A$ in $(S_1, S_2)$.

4. Initiate $K_1$ is the key as an array of characters with sizes such as picture height $I_2K_1(S_2)$.
5. In the image height of $I_2K_1$ repeat, $I_R{}^{th}$ is an individual row
   i. Let *count = 0*
   ii. $J_A$ is an individual pixel of the black color picture in the $I_R{}^{th}$ row. From $I_R$ count increase by 1
   iii. Use the ASCII code of the count subsequently generated b, obtain the character $K_1I_R$. i.e., $K_1I_R = char\ (count)$
   iv. Obtain $K_1[I_R] = K_1I_R$
6. $K_1$ initializes the AES-256 with a hash $(K_1)$ using the key (SHA-256)
7. Decrypt the $C_t$ ciphertext and save the base64 encoding decrypted as image $I_1$.

8. The result $I_1$ is the image that is decrypted.

### 3.10.3 Image Encryption and Decryption Output

The input image is encrypted with AES and SHA-256 algorithm for biometric security prospect. Figure 3.9 shows the image encryption and decryption.

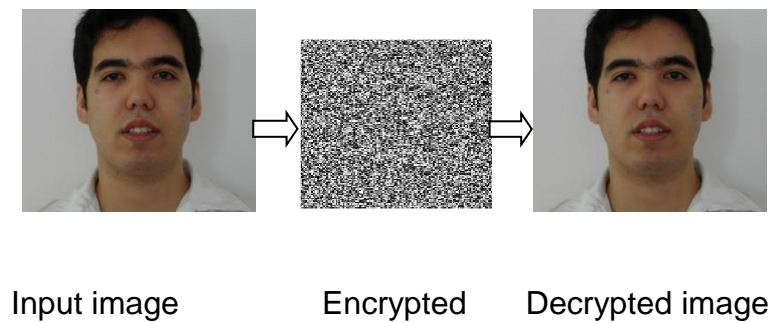

Input image      Encrypted      Decrypted image

**Figure 3.9: Encryption and decryption of image**

### 3.10.4 QR Code Generation

In this chapter, first build the biometric passport data string bits to construct the QR code. The string translates the data into the QR code that needs to be encoded. To create an error correction code for the QR code, it applies the Reed-Solomon error recognition technique. The resulting data is used to render eight compositions of the QR code. There is a replacement disguise form for each QR code and the bits move from each disguise form according to their headings in the QR matrix. It aids to construct the least required QR code for a QR scanner to be used. The length of the characters reaches 1264 characters, and the comparative scheme is rehashed until the entire encoded message. The biometric data encrypted is stored or encoded using the QR Code Generator online software [133]. (a) The encrypted personal information and (b) encrypted biometric information are stored in the QR code are shown in Figure 3.10.

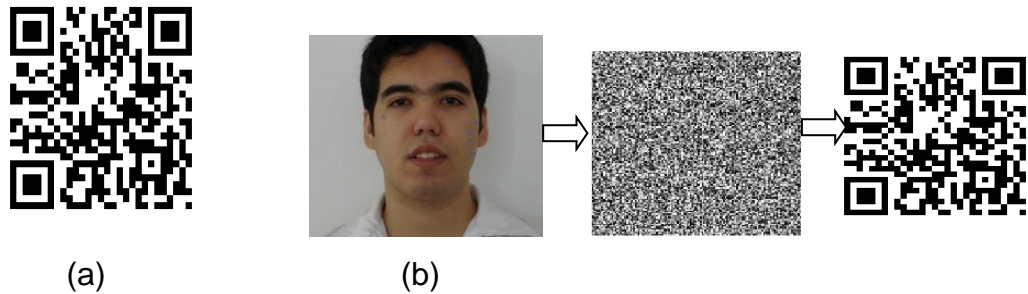

          (a)                    (b)

**Figure 3.10: (a) Encrypted personal information stored and (b) Biometric information encrypted and encoded**

## 3.11 CONCLUSIONS

In this chapter, a new framework for demographic information encryption, biometric data encryption, and QR code encoding has been proposed for secure biometric passport authentication. It focuses on detecting the size of facial marks, recognition of hand geometry and encrypted demographic information, encrypting biometric data encoded into QR code for national security, and the application of border control. By employing the AES with the SHA-256 method, the biometric passport information is encrypted to protect the information from the decipherable threat of information leakage. The inspection process is carried out in this study in two stages: 1:1 verification and 1: n identification, which emphasizes the recognition of the size of the facial mark with an emphasis on different sizes of the facial mark, and the other is 1:1 resolved verification. The proposed technology has improved the protection of biometric passports because the encoded biometric passport information in the QR code is not an active feature. They are much cheaper and do not require particular data retrieval instrumentation. QR codes are read-only components that are prudent and passive which can not be altered and decoded with specific gadgets. By concentrating on securing RFID and improving its protection, this study contributes to the current literature. The proposed method attained 92 percent (IITK database), 93 percent (FEI database) accuracy, and 96.67 percent accuracy attained in hand geometry recognition.

# CHAPTER 4

# FACIAL BLEMISHES DETECTION AND ENCRYPTION USING SECURE FORCE ALGORITHM AND HCC2D CODE FOR SECURE BIOMETRIC PASSPORT

## 4.1 INTRODUCTION

The significance of border security, national security, and protection is rapidly developing from a security viewpoint because of high interconnectivity throughout the world. Reasonable distinguishing evidence is needed by the traveler for national security, homeland security, and border control systems. Afterwards, the importance of including new components of their identity 'Biometrics' has been seen by numerous national bodies in the post-September 11 [134, 30] elaborated the number of biometric applications is faster developing with an important part of the biometric development being regulatory applications searching for more prominent accuracy and provide the security of the identification of citizens. Monar, D. [136] talked about security and protection rises that utilize biometric passports. They communicated bear on that, the contactless chip inserted in a biometric passport permits the substance to be read and authenticate without direct contact with a passport reader machine, and most significantly, the biometric passport booklet is unopened. They debated that data stored in the RFID could be clandestinely gathered by the methods of "eavesdropping" or "skimming".

Some of the threat scenarios found in the current study are connected to the issuance of travel documents [38]. It addresses the problems that occur in issuing biometric passports, for instance, identity theft, host listing, risks of data leakage, and tracking [39]. Two levels of attack are susceptible to the RFID method, namely on the transport and network layers [40]. The assault on RFID tags can be carried out by spoofing, cloning, impersonation, application layer, eavesdropping, tag modification, and unauthorized tag reading. Four attack methods are used by antagonists to determine the protection of a device using ISO/IEC 14443 RFID, for example, cloning, relaying, skimming, and eavesdropping [41]. An RFID reader that could generate an accurate Electronic Product Code (EPC) can block the RFID computer, and intruders could admittance the chip readers or it can be blocked by an EPCs and read sensitive data from the chip. Tag cloning, signal interference, tag killing, denial of service attacks, eavesdropping, and jamming will tamper with RFID. A hacker will disrupt the frequency in any situation and prevent the message from reaching the destined recipient. Although RFID is undoubtedly a promising technology, it has several technological disputes assorted with it, namely problems with collisions, issues with privacy and protection, problems with interfaces, and miscellaneous challenges [56].

In this chapter, a novel technique for facial blemishes detection has been proposed, where the biometric information is encrypted in the High Capacity Color Two Dimensional (HCC2D) code act as a significant character in biometric data. Using the Secure Force (SF) cryptography algorithm, biometric features are encrypted here and the biometric information is stored into the HCC2D code, which can be printed on the electronic passport. Malicious tampering would prevent this method of encrypting the biometric information in the HCC2D code.

The proposed solution is a better substitute for the encrypted HCC2D code compared to RFID since there is no risk of leakage of information, as the encrypted HCC2D code is not an active feature. As the data is encrypted and encoded in the HCC2D code, no network connection nor a distant database for recovery is required. The amount of information in the HCC2D code is nevertheless dependent on its space constraint and the minimum details required to discriminate against persons on the basis of biometrical properties. If the bearer data is permitted to access biometric information, the data encrypted and encoded in the HCC2D code implanted in the electronic passport will be automatically checked in order to authenticate the credentials of the owner of the e-passport. More specifically, the process outlined in the proposed system permits for safer and faster verification without human intervention of the claimed identity of the biometric passport holder. Authentication is accomplished by contrasting the local scan of a person's biometric details with the information contained in his electronic passport and the database. This biometric information encrypted and encoded into the HCC2D tag would improve the security of the electronic passport when doing verification and authentication since it is not an active function. This study would strengthen the protection of biometric passports globally. Figure 4.1 demonstrates the proposed structure of the framework for biometric feature encryption.

The HCC2D tags cannot be employed as an active element and have the capability to encode the biometric features. Besides, it is not over-priced and does not require specialized data recovery hardware. HCC2D tags are effective passive read-only tags whose information can not be altered by a specific machine and can be decoded.

The main contribution of this analysis are summarized as follows:

- By applying the SF algorithm, the biometric information is encrypted and encoded into the HCC2D code.
- Without the user's awareness or consent, the encrypted biometric information contained in the HCC2D code should not be exposed.
- During the verification process, the encrypted biometric information stored in the HCC2D code would improve the security of the biometric passport, since it is not an active part of the verification process.

## 4.2 PROPOSED METHODS

The face recognition technique focused on the detection of facial blemishes and encrypted by applying the SF algorithm into the HCC2D code is introduced in the chapter. The facial blemishes detection is achieved by employing the algorithms AAM utilizing PCA and Canny edge detector with SURF detector. This cryptographic mechanism will ensure the biometric data to be secure without involving peculiar hardware for border control applications and national security. We could preferably imprint a 'publicly readable' HCC2D code on the biometric passport to protect the data since it is not an active element. Therefore, the biometric passport should hold on persisting passive, this intends that it should not reveal data without its bearer's cognizance and permit. The RFID labels were decipherable generally without such awareness. In our system, the HCC2D code technique can be utilized as storing the data for a biometric passport. Furthermore, we present face recognition techniques based on facial blemishes recognition to identify authorized persons. The facial blemishes as a signature of a person to identify and verify the biometric passport. The proposed method for the control of biometric passports should be useful and practical. Figure 4.1 shows the architecture of biometric encryption.

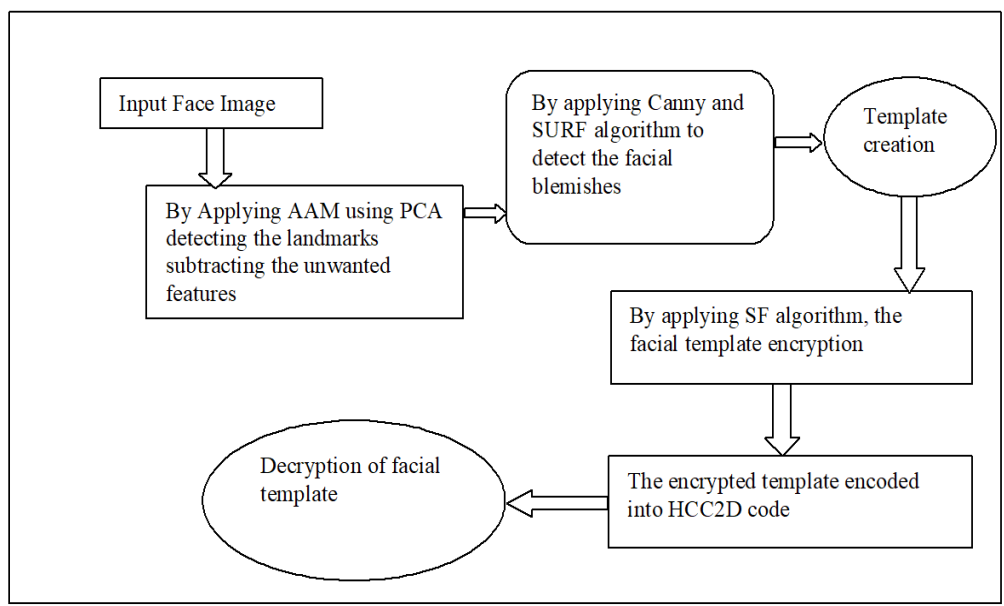

Figure 4.1: The architecture of biometric encryption in the HCC2D code

## 4.3 FACE RECOGNITION METHOD

Facial images are the most regular biometric characteristic utilized by people to make singular validation. Therefore, the estimate to apply this biometrics method for security purposes. It is a nonintrusive technique and publically acceptable system for identification applications. The facial image of a person is utilized as an utmost security component. nevertheless, due to poor matching images of unknown faces [101] from terrorism, criminals, illegal immigration, and fake passport holders, there were chances of failure of identification. In another scenario, similar faces such as twins, siblings, similar faces, or even doubles could head to an individuality mismatch. Therefore, if a biometric point of view deals with the facial image, the face retains data that is fixed in time and can be analyzed over the decades. Biometric attributes are unique to the person, and characteristics are constant in time. Generally, it is an unambiguous or an independent identifier of an individual from a specific period. This chapter discusses the biometric passport authentication approach based on face recognition, focusing on facial blemishes as a signature and encrypted by employing the SF algorithm, in order to prevent threat scenarios from false identities. To identify a person, facial blemishes may distinguish between individuals.

### 4.3.1 Facial Blemishes on the Face

A facial blemish is any form of marks, discoloration, spot, or flaw that present on the face skin. Facial Blemishes are situated on dissimilar areas on the face. To determine the various classes of blemishes appearing on the face, it requires analyzing the various kind of blemishes. Different categories of facial blemishes are papules, nodules, Age spots, pustules, birthmarks, scars, whitening, dark skin, and hyperpigmentation.

- Papules: Papules are small skin lesions of varying types. They are typically around one centimeter in diameter.
- Nodules: Nodules are a collection of tissues and usually 1 to 2 centimeters in diameter.
- Age spots: These small dark spots can form on any area of the face or body.
- Pustules: pustules are fluid or pus filled bumps.
- Birthmarks: Birthmarks typically occur either at birth or shortly afterward. They can range in appearance, size, shape, and color.
- Scars: Scarring of the skin occurs when the dermis layer becomes damaged; Scar plays the discolored area in the skin.

- Whitening: Whitening acts as a skin area that appears to be brighter in contrast with the surrounding area.
- Hyperpigmentation: An overproduction of melanin can cause uneven skin tone or dark patches.

If there is any kind of thick fix of light or dark spots found, it is labeled in a bounding box. We obtained into condition the wrinkles that are bigger and prohibited the little wrinkles close to the mouth and the eyes. We additionally ignored facial hair and the beard in structure the ground truth. Different classes of facial blemishes that are excluded previously mentioned are named under the "others" classification. Figure 4.2, demonstrates the different facial blemishes in a face picture. The normal number of facial blemishes that are identified in the dataset is 6 for every subject. Every one of the pictures in the dataset presents at any rate 3 to 4 facial blemishes for each subject.

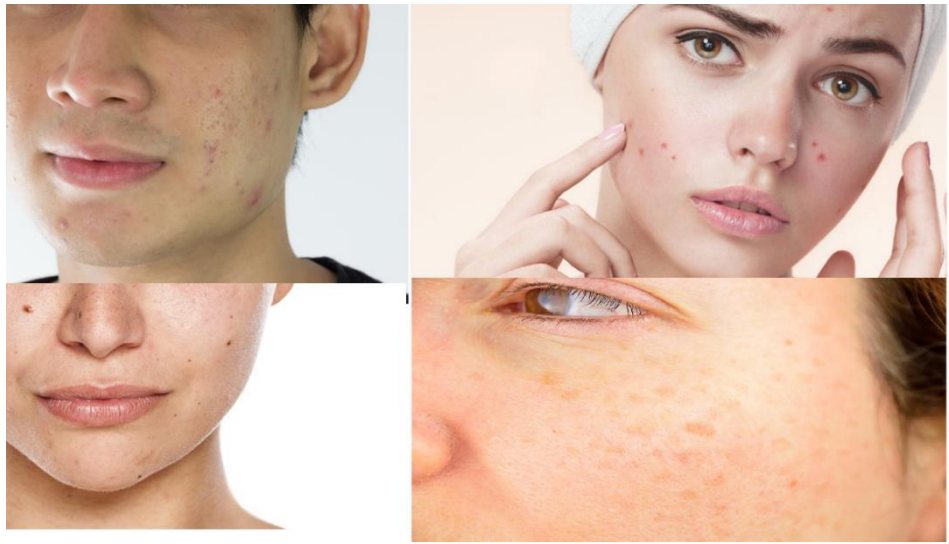

**Figure 4.2: Different kinds of facial blemishes on the face**

### 4.3.2 Facial Blemishes Detection Algorithm

The proposed facial blemishes detection algorithm is depending on automated facial blemishes detection techniques present an elaborated description in figure 4.3. The blemishes detection technique is attained by applying the Canny edge detector [113] and Speeded Up Robust Features

(SURF) [137]. The following technique has been used to identify facial blemishes:

Step 1. The main facial features are found through the application of the Active Appearance Model (AAM) using PCA [138] [139]. The features of eyebrows, eyes, mouth, and nose are eliminated from the face to reduce the false positive detection.

Step 2. The masking procedure is designed by employing the Canny Edge detector.

Step 3. The SURF algorithm is used to detect facial blemish's features.

The facial blemishes are detected in the feature extraction process. The facial blemishes are extracted in the succeeding sections and elaborate on the procedure of localization of primary facial features, mask generation, and facial blemishes detection.

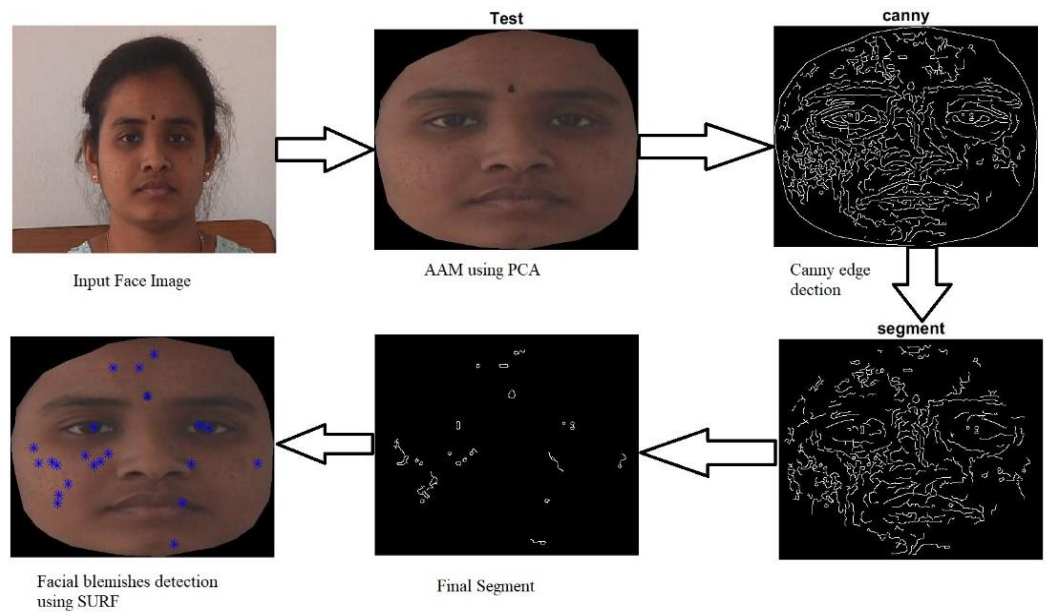

**Figure 4.3: Automatic facial blemishes extraction process**

### 4.3.3 Facial Feature Extraction

The facial features like eyes, eyebrows, mouth, and nose are recognized by applying the AAM using PCA procedure and subtracted the undesirable features. Therefore, facial features are masked before the location procedure to keep away from false detection amid the identification procedure. Depending on the preparation database, an AAM using PCA is prepared, which shapes a new contour to match the new picture. It authenticates the landmark points of 110 relating to the facial feature's form. AAM using PCA distinguishes both the texture and shape of the face images and senses the shape parameters to render the profile like the pre-learned profile in each model. Here the set of landmark dots in the training image data fixed by applying them manually. A set of image shapes data $I=\{i_1,....,i_\mathcal{N}\}$ and corresponding textures $T = \{t_1,......,t_\mathcal{N}\}$ are received. By applying PCA to get eigenvalues on $I$ and $T$, the shape and texture of the principal components, $A_i$ and $A_t$ are found. The facial image shape and texture, $I_{new}$ and $T_{new}$, of a new face image can be explicit as $I_{new} = I_\mu + A_x b_x$ and $T_{new}=t_\mu + A_G b_G$, where $I_\mu$ and $T_\mu$ ( $b_X$ and $b_G$) are means (weight vectors) of $I$ and $T$, individually. The 110 landmark dots are labeled on a subset of the face image database for training.

### 4.3.4 Designing Mean Shape and Mask Construction

AAM using PCA has been used to recognize the facial landmarks to simplifies the facial blemishes detection from the face. Therefore, every face image is mapped into an average form. Let $F_i$ is each face image, here $i=1,2,3,…, N$ constitutes the form of individual of the $N$ facial pictures in the dataset established on the 110 landmarks. So, the mean shape is computed by applying the equation,

$$F_\mu= \sum_{i=0}^{N} F_i \quad \text{……………………………………… (4.1)}$$

Therefore, by utilizing the Barycentric coordinate-based [140] texture mapping method, $F_i$ is mapped to the mean form $F_\mu$. The generic mask is constructed and derives a user-specific mask is derived to suppress false-positive recognition across the facial blemishes feature. The user-specific mask covers little abnormal landmarks throughout the facial blemishes feature. Then suppress the false-positive beard connected to the facial element. Thus, a user-specific mask is produced from the edge detected by employing the Canny edge detector.

### 4.3.5 Facial Blemishes Detection

Facial blemishes are present on the face which are extracted by employing the Canny edge detector and SURF feature detection algorithm. The facial feature extraction steps are given below,

Step 1. Calculate all the edges which present on the face by utilizing the Canny edge detector algorithm.

Step 2. Filtered the facial image with the two-dimensional Gaussian filter, to suppress the noise considered σ = √2. The Gaussian function is given below,

$$G(x,y,\sigma) = \frac{1}{2\pi\sigma^2} \exp(-\frac{1}{2\sigma^2}(x^2+y^2)) \quad \text{............................ (4.2)}$$

Where σ is a variance of the Gaussian Function, the distance from the horizontal axis is *x*, and the distance from the vertical axis *y*.

Step 3. An edge detector operator is applied to find out the edges in the vertical and horizontal direction of each one and combined to determine the overall directions.

Step 4. Determine the maximum gradient magnitude in the direction perpendicular. The gradient magnitude is calculated with the below equation,

$$D(x,y) = \sqrt{F_x^2(x,y) + F_y^2(x,y)} \quad \text{............................................. (4.3)}$$

Step 5. Apply a large threshold to find all definite edges and trace uninterrupted on these edges with the same threshold value to develop the final edge.

Step 6. Finally, the facial blemishes' features are detected by applying the SURF descriptor. Figure 4.4, shows the detected facial blemishes.

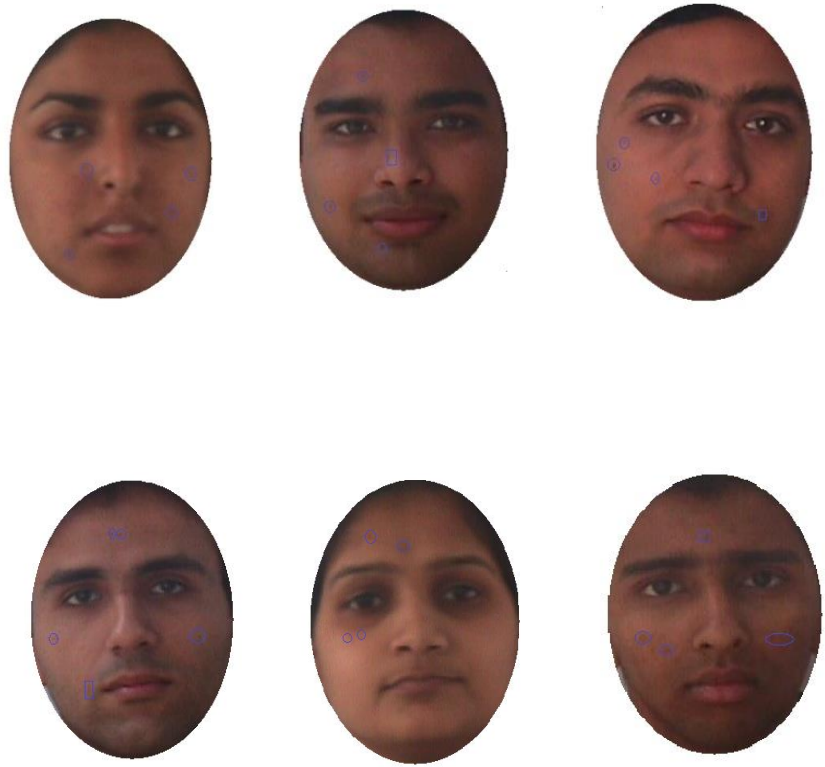

**Figure 4.4: Facial blemishes are detected from the face image**

### 4.3.6 Matching the Facial Blemishes

The two sets of images are known to be $I_1$ and $I_2$, with $N_1$ and $N_2$ set of their observed facial blemishes. The recognized facial blemishes and the similarity set of two images $I_1$ and $I_2$ (We call *FB*) are measured employing the equation given (4.4).

$$FB = \frac{\sum_{i=0}^{|N_1|} \min D(n_i, n_j)}{|N_1|}, \forall n_j \in N_2 | (x_j, y_j) \in R_i \quad \ldots\ldots\ldots\ldots(4.4)$$

When each facial blemish is assumed to be $n_j \in N_2$, the spacial central coordinate points are $x_j$ and $y_j$. Each $n_i \in N_1$ is facial blemishes, and $R_i$ is a rectangular region constructed as an area of highly matching facial blemishes $n_j \in N_2$, across its central $I_2$ coordinates, held in $R_i$, which is accepted into account during the matching blemishes skin detection process. Consider D is to calculate the length among facial blemishes, where the distance is measured employing the algorithm Bhattacharyya distance [117] and obtained the best result. When contrasting and finding histograms during the implementation of the experimental procedure, our approach shows dramatic results.

## 4.4 EXPERIMENTAL RESULTS AND DISCUSSIONS

The facial images with manually annotated facial blemishes were created to perform the experiment in order to verify the findings. We were unable to find any related references to literature based on the identification database for facial blemishes. Hence, we have collected the face image database from IITK (Indian Institute of Technology, Kanpur) [118] and manually annotated the facial blemishes to perform the experiment. The annotation process was done based on the 11 different face images of individual forty distinct subjects (440 images). The images were taken bright homogenous with the subjects in a frontal and upright position. The face images are categorized as males and females in a JPEG file format. We manually annotated the blemishes to perform the facial blemishes detection and an average of 5 to 10 was found. Figure 4.5 shows the manually annotated facial images. The facial blemishes are classified such as papules, nodules, Age spots, pustules, birthmarks, scars, whitening, dark skin, and hyperpigmentation. In this chapter, all kinds of facial blemishes to match the individual identity are constructed. At last to match the identity in this experiment considered 3 to 4 blemishes must match. Hence, this technique will improve identification accuracy. The proposed facial blemishes detection technique achieved a 93.03% accuracy level for the dataset IITK and implemented using MATLAB (R2018a), intel core i7 CPU, 6600U, 16 GB RAM, 1TB SSD.

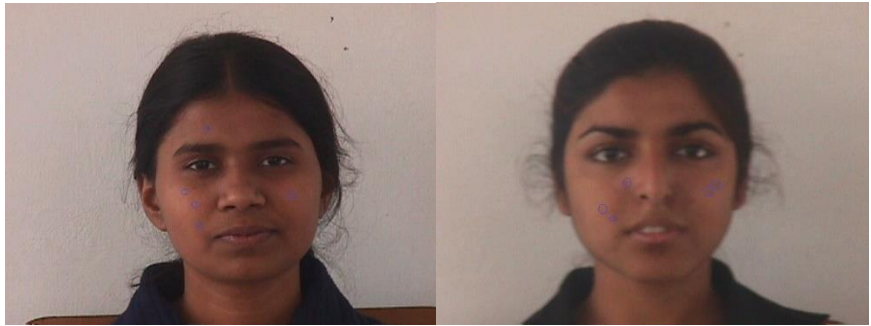

**Figure 4.5: Manually annotated facial images**

### 4.4.1 Evaluation of Facial Blemishes Detection

The validation of the accuracy of the proposed facial blemishes recognition technique using the IITK face dataset has been experimented taking into account the measurement of Precision and Recall. We generalized the standard measure to evaluate facial blemishes detection to demonstrate the resemblance among the observed facial blemishes and annotated ones. Consider the image $I_m$ with $A_i$ annotated facial blemishes, the detected facial blemishes $f_b$ was considered accurate if $\exists f_a \in A_i$ such that

$$\frac{area(f_b) \cap area(f_a)}{area(f_b) \cup area(f_a)} \geq t_0 \qquad (4.5)$$

The threshold $t_0$ was established empirically as 0.4

No valid implementations of the algorithm for detecting facial blemishes are found in the literature. Therefore, to compare the proposed algorithm and facial blemishes results we contrasted with Viola and Jones object detector [122], (we call VJ). This algorithm was trained to detect facial blemishes with a collection of 220 positive facial blemish samples and 440 negative sample images without facial blemishes. The outcome of the facial blemishes is shown in table 4.1. It was observed that the proposed facial blemishes detection algorithm attains higher values of precision and recall. Therefore, it is able to detect more accurate facial blemishes and less false positives.

**Table 4.1 The proposed P_SURF technique contrast with VJ using the IITK database**

| Algorithm | Precision (%) | Recall(%) |
|---|---|---|
| P_SURF | 73.12 | 65.99 |
| VJ | 32.01 | 19.11 |

Note. VJ= Viola and Jones, SURF= Speed Up Robust Feature proposed a method for facial blemishes detection.

### 4.4.2 Validation of Face Verification

To calculate the accuracy of the identification of facial blemishes: False Acceptance Rate (FAR) and False Recognition Rate (FRR) is calculated when the Operation Point (OP) is 0.1 percent conceived as False Acceptance Rate (FAR), we named as (FB-OPFRR).

In this experiment, the verification is performed based on the IITK database by employing the procedure of matching the facial blemishes. Here experimented with this database to compare the facial blemish detector algorithms with the established verification accuracy methods, as it has a larger variance in facial blemishes. The experiment is conducted on the basis of manual annotation (M_Annotation) and the outcomes match the image of the face with manually annotated facial blemishes. Table 4.2 displays the experimental results for EER and FB_OPFRR and provides the results of the M_Annotation. The proposed facial blemish detector has obtained better results compared to the VJ detector on the basis of the outcome analysis. It is observed that fewer error rates with the manual annotation experiment than the automated detection technique, hence manual annotation is more beneficial.

**Table 4.2 To contrast the proposed P_SURF algorithm with other algorithms using the IITK dataset**

| Algorithm | EER(%) | FB_OPFRR(%) |
|---|---|---|
| M_Annotation | 31.10 | 98.77 |
| P_SURF | 32.33 | 98.89 |
| VJ | 34.09 | 99.07 |

It combines two very common techniques with the proposed technique of matching facial blemishes to assess the efficacy of face recognition for face detection and face verification (i) AAM (Active Appearance Model) and (ii) Fisher Vector (FV) method [123].

The experiment is carried out one versus one for verification validation on the IITK database. For matching the facial blemishes, automatic detection (FB_Auto) is calculated and compared with the results of AAM and FV techniques, similar to their combinations. The experiments are performed under identical conditions by employing the manual annotation of facial blemishes (M_Annotation). It will have the advantages of the proposed face verification experiment. The FM-OPFRR and EER dependent outcomes are done employing the IITK database. Table 4.3 displays the results. From the values of the first two rows in table 4.3, it is observed that facial blemishes are inadequate to be included in the face verification steps. Hence, it enhances the accuracy of face verification when combined with existing popular face recognition techniques FV and AAM. The improvement in the EER (Table 4.3) is increased in FV + M_Annotation contrast with FV alone, it is reduced by more than half error rate. The error produced in the automatic detection of facial blemishes is not applicable to the last result when combined with AAM, and minor alters happened in AAM + FB_Auto and AAM + M_Annotation, but with the combination with FV, the detection phase of facial blemishes shows improvement.

**Table 4.3   Compare amongst FV, AAM, in proposed face verification experiments using the IITK dataset**

| Algorithm | EER(%) | FB_OPFRR(%) |
|---|---|---|
| FB_Auto | 31.73 | 98.11 |
| M_Annotation | 20.57 | 97.67 |
| AAM | 3.88 | 6.99 |
| AAM + FB_Auto | 1.98 | 5.99 |
| AAM + M_Annotation | 1.67 | 5.66 |
| FV | 1.55 | 5.59 |
| FV + FB_Auto | 1.34 | 4.88 |
| FV + M_Annotation | 1.02 | 3.69 |

Other processes are taken to eliminate mistakes, and the effects of the facial blemishes matching output are the selection of the weights $W_{fr}$ and $W_{fmm}$ for the combination of the algorithms. The effect of the identification of facial blemishes is less revealing. Hence, with higher $W_{fr}$ values (0.8 and 0.9) and lower $W_{fmm}$ values (0.1 and 0.2) combination produces the best outcome.

### 4.4.3   Validation of Face Identification

To perform the basic recognition method, a 5-fold cross-validation technique was used by using one face image at each time of an individual from the dataset and the remaining of them as a examine. The IITK database is employed for the validation in this experiment and compares the effects of facial matching with automatic detection of facial blemish, manual detection of facial blemish, and the FV technique with the combination. In the previous segment, it has proven the most beneficial of the findings. Table 4.4 indicates the attained results with regard to the average face recognition levels at Rank 1 (ARR_R1) and Rank 5 (ARR_R5).

**Table 4.4 The proposed technique and FV combination of identification experiments**

| Algorithm | ARR_R1 | ARR_R5 |
|---|---|---|
| FB_Auto | 29.72 | 51.44 |
| M_Annotation | 41.33 | 59.76 |
| FV | 90.22 | 97.23 |
| FV+FB_Auto | 91.99 | 98.67 |
| FV+M_Annotation | 93.06 | 98.99 |

The table above shows that a high identification rate is not reached by the use of facial blemishes alone, but it increases the results of face detection by compounding a face recognition technique. By implementing the FV method alone achieving 90.22 percent of the Average Recognition Rate (ARR). Hence, considered Rank 1 to Rank 10, which boosts the ARR when matching facial blemishes is conceived for both manual and automatic forms. The experimental outcome up to Rank 10 is shown in Figure 4.6.

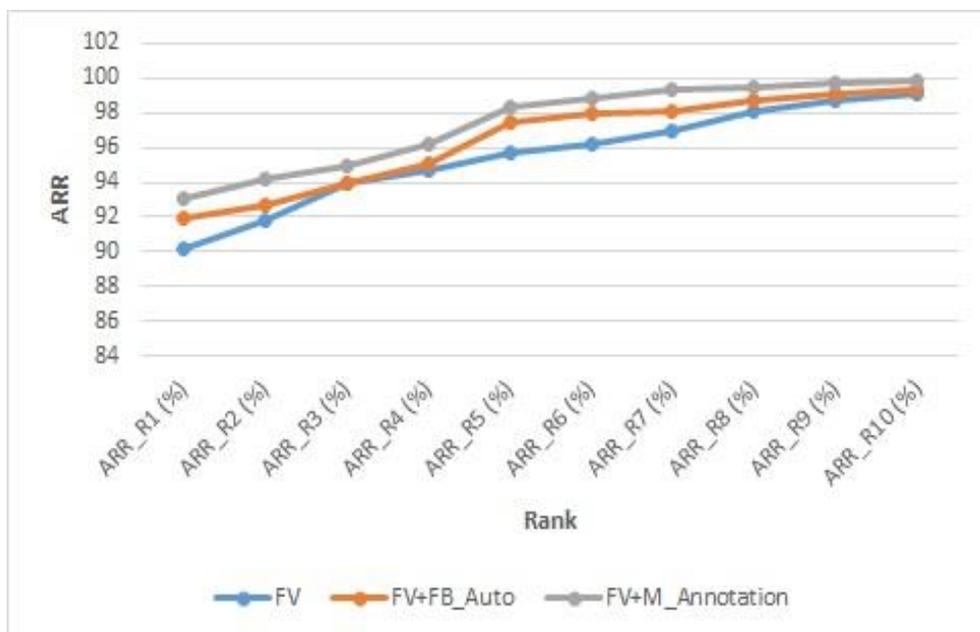

**Figure 4.6: The average recognition rate R1 to R10 in percent**

## 4.5  BIOMETRIC ENCRYPTION

Biometrics is a skill that estimates a person's physical features uniquely. Hence, biometrics assists as a technique to supplant the unwieldy utilization of complex passwords. Our exploration utilizes the highlights of biometrics to proficiently actualize a biometric encryption framework with an utmost state of security. Biometric encryption is a technique that adheres securely in a biometric template to a cryptographic key, so it is not possible to retrieve either the cryptographic key or the biometric key from the stored biometric template. A new key is generated merely if the accurate live biometric information is established for confirmation. The cryptographic key is arbitrarily generated on enrolment, thus, the hacker incognizant of the key information. The key can always be altered and the key is ultimately independent. Since a biometric taste is produced, the encrypted biometric firmly and systematically adhere the key to the image to produce a secure biometric encryption template and known as a private template. The biometric encryption template allows a splendid privacy assurance and it can store in a database or a token, smart card, etc. In this proposed system the biometric encryption is stored in HCC2D code.

For the part of authentication, it is checked through the legalized biometric encrypted prototype when the individual contributes his/her recent biometric sample. This helps the encryption algorithm to recover a decryption key and the biometric aids. The biometric pattern will finally be disposed by the authentication section. In the case of an intruder whose biometric information is different, the cryptographic key won't be retrieved by the template. This method of biometric encryption and biometric decryption is also fuzzy because every time the biometric prototype is different, and standard cryptography is used to construct the dissimilar encryption key. The cryptographic key is then recovered and can be very well used for some logical or physical use as a justification. Thus an image encryption technique is, therefore, an effective, privacy-friendly, and secure biometric key management method. An image encryption technique using SF algorithms stored into the HCC2D code is proposed in this study. This technique would ensure the biometric information of a biometric passport is secured.

### 4.5.1 Encryption of Biometric Data into the HCC2D Code

The biometric information encryption is developed by using the SF algorithm [141]. The SF algorithm is established on the Feistel framework where the technique of encryption and decryption are almost similar and minimizes the code size in great magnitude. The main aim of the SF algorithm is to render a less complicated architecture. In the SF algorithm, the encryption process took place only five rounds of encryption. To provide better security, individual encryption round covers 6 uncomplicated mathematical functions and operating on 4-bit data. It is to establish a decent number of diffusion and confusion of information to face various cases of attacks. The SF key expansion procedure is carried out by complex mathematical functions such as rotation, permutation, multiplication, and transposition to create the key for the biometric encryption procedure and enforced in the decoder. It moved the computational weight to the decoder and by implication. Thus, for the biometric encryption method, the established keys are firmly carried out to the encoder. The Localized Encryption and Authentication Protocol (LEAP) employed in [142] is adopted in this paper. The SF algorithm carried out four major blocks and a detailed description available in [143]. In this chapter, the e-passport biometric information is encrypted which includes biometric information encoded into an HCC2D code. The facial image template is hidden from the attacker since it has been encrypted. The procedure of biometric encryption and decryption steps are illustrated below. Figure 4.5 shows the biometric encryption and decryption.

### 4.5.2 Key Expansion Block

The key expansion method that is employed to create various keys for biometric encryption and biometric decryption has been carried out by the SF algorithm. In an order that causes diffusion and confusion, dissimilar processes are carried out. In addition to improving the key quality, it is to reduce the hypothesis of a debile key. $R_k$ is the round keys that are acquired depending on the key schedule from the input cipher. This method is carried out by two techniques: round key selection, and key expansions. The key expansion is executed by applying Exclusive OR (*XOR*), Not Exclusive OR *(XNOR)*, Left Shift (LS), matrix multiplication by applying Fix Matrix (FM), permutation by applying P-table, and transposition by applying Truth Table (T- table).

The k is the input cipher linear array of 64-bits and isolated into four portions of 16-bits. These 16-bits are allocated into 4*4 matrix row-wise in LS function and then the resultant is arranged matrix column-wise of 4*4 bits. *XOR* and *XNOR* are the logical operators that are executed. Hence, the outcome of these procedures is mixed to create a linear array of 64-bits. The received 64-bits are transferred through P- table and 4*4 matrix row-wise arranged on LS operation are executed later. The multiplication of LS matrix and FM that transform the 16-bits to 64-bits data. Therefore, 64 bits are ordered row-wise and LS is executed. Individual LS 64-bits are divided with 4 columns of 16-bit blocks, AND and XOR operations are executed to convert into a single 16-bits block. Hence, the created 16-bits block is split into 4-bits organized XOR operation and column-wise are used to create the 4-bits key. These keys are applied transposition and substitution methods on the 16-bits block to generate the 4 subkeys such as k1, k2, k3, and k4, these four keys are from 16-bits and applied for the initial four encryption round. The k5 is the fifth subkey created by XOR the four subkeys and it will be applied in the 5$^{th}$ encryption round.

### 4.5.3 Key Management Protocol

The LEAP algorithm employed by Ebrahim, M. [143] key management is utilized in this encryption technique.

### 4.5.4 Encryption Process

After generating the key by the expansion block is firmly obtained through the communication channel by the encoder. To encrypt the biometric data sample operations carried out OR, AND, XNOR, XOR, left shift, substitution, and swapping functions executed to develop confusion function and diffusion function. The data is a linear array of 64-bits and divided into two half 32-bits. Every 32-bits are promoted to subdivided into two half 16-bits. Every round swapping of 16-bits data is performed to create a more complex cipher. K1, K2, K3, K4, and K5 are subkeys that are XNOR with the left and right half of individual round respectively.

F=OR (S-boxes (AND (LS (16 bits/4)))) ……………………(4.6)
Where F is the F function.

The calculation of the encrypted and original image entropy elapsed time is 1.646663 seconds. The encrypted image histogram plot diagram is given in figure 4.6.

### 4.5.5  Decryption

Decryption is the reverse procedure of the encryption technique.

### 4.5.6 Image Encryption and Decryption Output

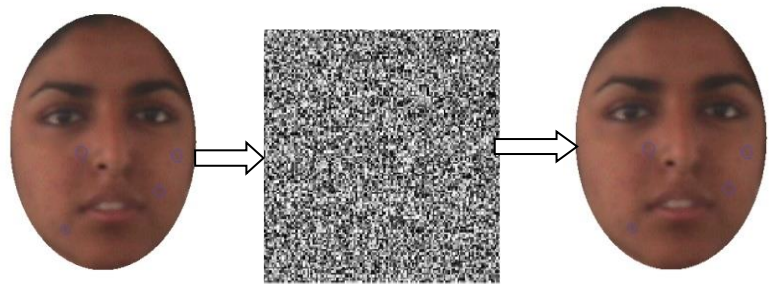

Input Image template    Encrypted        Decrypted template

**Figure 4.5: Image encryption and decryption using the SF algorithm**

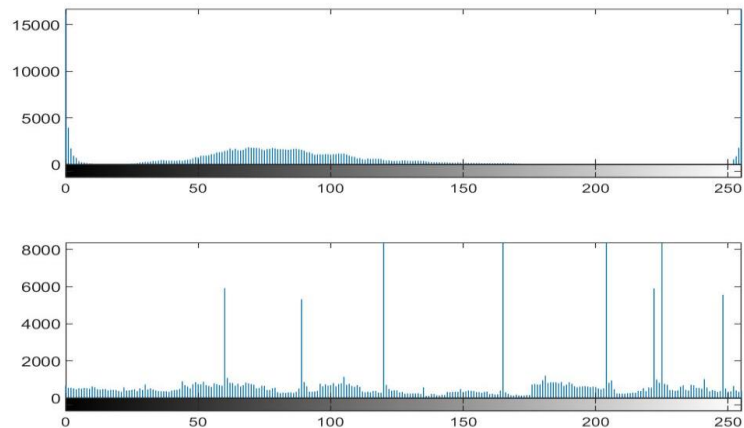

**Figure 4.6: The encrypted image histogram plot**

### 4.5.7 Generation of an HCC2D Code

The primary objective of the HCC2D code [144] to enhance the space available for information storage to continue robustness error correction properties like a QR code. By having each module in the information region with a color selected from a color palette, HCC2D code improves the storage space. HCC2D code contains some data, for instance, the symbol version, the total count, the error correction level and the Reed Solomon block type, etc. The encoder and decoder for creating and developing HCC2D code were recognized with the aid of libqrencode [145] and a C library for encoding information. For the decoder was constructed with the aid of [146] an open-source java project. Figure 4.7 shows the biometric data encryption and stored into the HCC2D code as well as the decryption of the facial template.

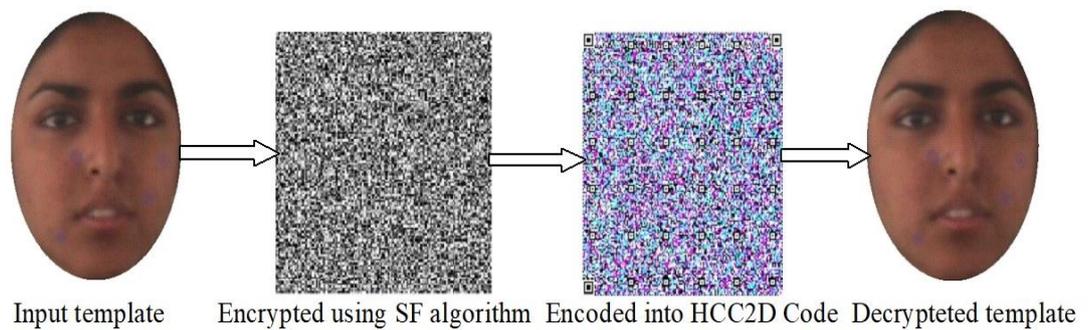

**Figure 4.7: Encryption and encoding into the HCC2D Code with decrypted template**

### 4.5.8 Verification of the Person

The object of the verification stage is to confirm the character of a given individual. Given facial blemishes as a template, we target checking the asserted personality of the individual in the scene utilizing just facial biometrics facial features stored in the HCC2D tag embedded in a biometric passport. Such a record should be introduced by the individual who professes to claim the record. Alternatively, given a paper report which constitutes the individual exhibiting the record, the target of the verification step is to consequently acknowledge or dismiss the asserted personality of the individual. The procedure begins with the accompanying sources of info: face recognition and facial blemishes matching of the individual to confirm and the HCC2D tag biometric data stored secure facial features of the claimed individuality.

## 4.6 CONCLUSIONS

In this chapter, a novel face recognition technique is proposed based on the identification of facial blemishes using the SURF algorithm and encrypted using the SF algorithm. Hence, this paper demonstrated facial blemishes detection by applying AAM using PCA to detect the facial landmarks, a Canny edge detector to detect the edges of the facial blemishes, and the SURF algorithm is employed to detect the facial blemishes features. Facial blemishes features are considered as a signature to authenticate a person. These facial blemishes features are encrypted by applying a cryptographic SF algorithm to assure the biometric data. The encrypted biometric data is encoded into HCC2D code and embedded into a biometric passport to keep secure the information. It could print ensured biometric information on a biometric passport through the HCC2D code. Due to the encrypted biometric information encoded in the HCC2D code. This method will ensure the privacy and protection of the e-passport and cannot be used as an active element. Certainly, HCC2D codes are economical, read-only passive, and it is not possible to alter the information. To expand the data size to store the information, require to concentrate on privacy and security for future improvement employing 2D barcodes such as QR code, HCC2D code, etc.

# CHAPTER 5

# COSMETIC APPLIED BASED FACE RECOGNITION FOR BIOMETRIC PASSPORT

## 5.1 INTRODUCTION

There is a substantial gain in the exercises of terrorists and criminals in the private and government property because of unconventional verification of the electronic passport. Thus, unconventional authentication because of cosmetic or makeup employed faces, identical faces, and twins are other issues raised. There are a few reasons for the abuse and cloning of biometric passports by criminals, illegal immigrants, and terrorists. Focusing on safe biometric authentication particularly on face recognition, it is significant to avoid such criminal acts. The most mainstream biometric information used to verify a person is facial images. It is a non-intrusive type of information for covert detection that is freely acknowledged.

Digital records and biometric features replace traditional identity proof archives that empower the system to easily ensure the character of the person. Biometric details and electronic information have emerged universally for identity verification purposes [147, 148]. A special working group has been set up by the ICAO to predict the proper strategies for the unique encoding of a person's physical characteristics into a biometric picture, with the machine-readable format, this can easily be tested to validate an individual's personality [54], [149], [27].

The ISO/IEC 19794-5 standard FCD19794-5[152] and the ICAO rules Doc 9303 ICAO [27] explained the setting of the guidelines for the recording, encoding, and transmission of biometric data to establish the scene specifications for digital image properties and face image photographic attributes. The ISO/IEC standard has been incorporated in legitimate arrange, an amendment [151] explained with the consideration for capturing photographs and two corrections [152], that have been absolved a portion of the limits mentioned in the first edition. To explain the terms and conditions associated with the use, testing and stipulation of face image character measurements, the following additional document [153] was released to indicate the interpretation, aim, and function of face image quality scores. In this report, it was stated that quality algorithm output computation and quality algorithm standardization are beyond the reach of this investigation.

One of the consistency profiles for facial achievement is ANSI/NIST-ITL 1-2-11 standard [154]. The ISO/IEC 19794-5 standard provides conventional guidelines which are alternative satisfactory and inadmissible for face images but a reasonable and definitive depiction is as yet absent. The task two-dimensional facial image quality [155] was performed concerning the ISO standard. It is noted from the existing research that a biometric passport faces different disadvantages in terms of face recognition and no one studies on the basis of the cosmetic applied face, identical face, and twins to date. In this study, cosmetic-based face recognition is introduced and focused on allowing an advanced biometric passport protection system to maintain a strategic distance from illegal migrants. In order to differentiate the individual and distinguish the twins and related faces, a face recognition technique is proposed based on the detection of facial marks from cosmetic or make-up faces. Figure 5.1. (a) presents identical faces, and (b) the twins are shown.

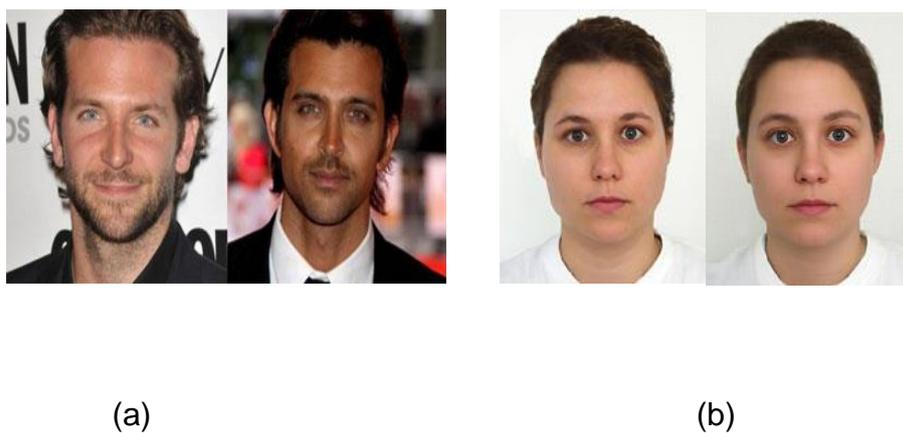

(a)                               (b)

**Figure 5.1: (a) Presents the example of similar faces and (b) Example of twins**

An up-to-date national security authentication procedure tends to be the most critical consequence of insuring a border control application with respect to a biometric passport. It is noticed that [159] verification techniques acknowledge cloned or forged a biometric passport. These days, cosmetic applied or facial makeup faces presents a challenge in recognizing an individual's personality. As an important safety factor, a facial picture of a candidate is used. A greater focus on physical and electronic inspection is required. From a biometric point of view, the facial image has been disputed, the face contains time-consistent data and can be estimated for up to a decade. Biometric characteristics are one of a kind to the distinct individual and ought to be steady in time.

An algorithm is proposed in this chapter to detect the facial mark from the makeup or cosmetic face used for the authentication of an individual distinguishing to maintain a strategic gap from dangerous false identity situations of an electronic passport. The validation of biometric passports depends on the identification of the face based on a signature of permanent facial marks. Therefore, identifying individual facial marks may distinguish a person. To identify individuality, facial markings are exceptional. The truth is that they can distinguish the identification if the individuals are very similar in appearance. Facial marks are available on the face at an irregular location. They are eliminated from the surrounding skin as noticeable alters between the face and completely distinct in color, structure, and texture.

The primary aim of this research is to recognize the various features of the facial mark situated on the face. To determine the various types of facial marks, each face image of the cosmetic used dataset is manually annotated so that both position and category can easily be defined. The approval of face images depends on automated identification and the feasibility of this proposed technique is demonstrated by manually annotated facial marks. The detection of facial marks from cosmetic applied face eliminates the error in both the process of authentication and identification. In its ability to fine-tune the candidate's faces from a dataset, the proposed approach shows the best outcome in face recognition.

## 5.2 PROPOSED METHODS

Facial marks which can be consistent marks or temporary marks are found on the face at various places. To extrapolate the various types of facial marks situated on the face, it includes analyzing the distinct types of facial marks. The signs conceived are dark spots, moles, whitening, pockmarks, scars, freckles, pimples, and birthmarks. The different marks found on the face are shown in figure 5.2. Because of the cosmetic face added, the marks were examined with numerous difficulties.

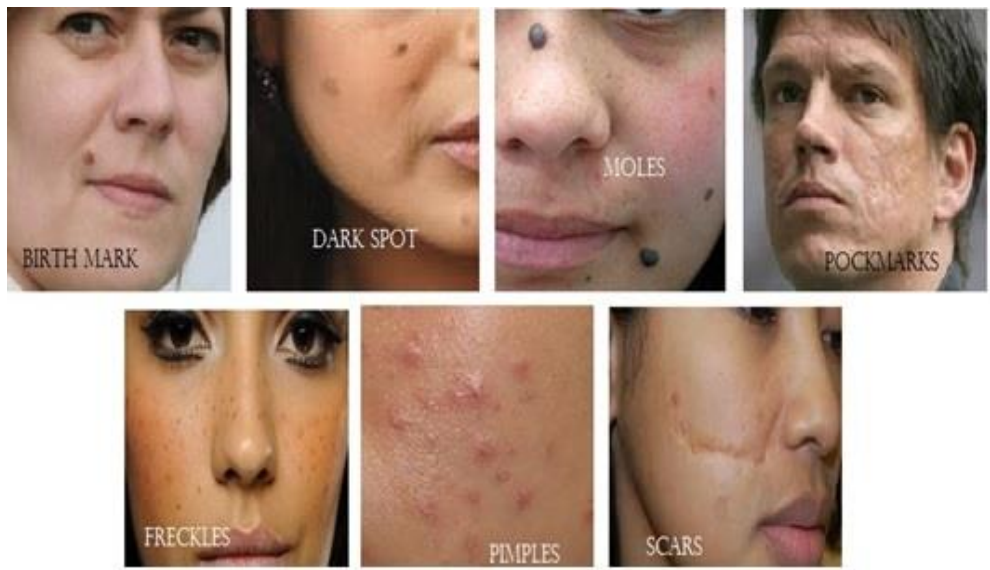

**Figure 5.2: Examples of the various forms of facial marks on the face**

For biometric passport security purposes, an algorithm was developed for the identification of facial mars from the cosmetic or make-up based facial image. The proposed facial mark detection algorithm based on makeup is shown in Figure 5.3.

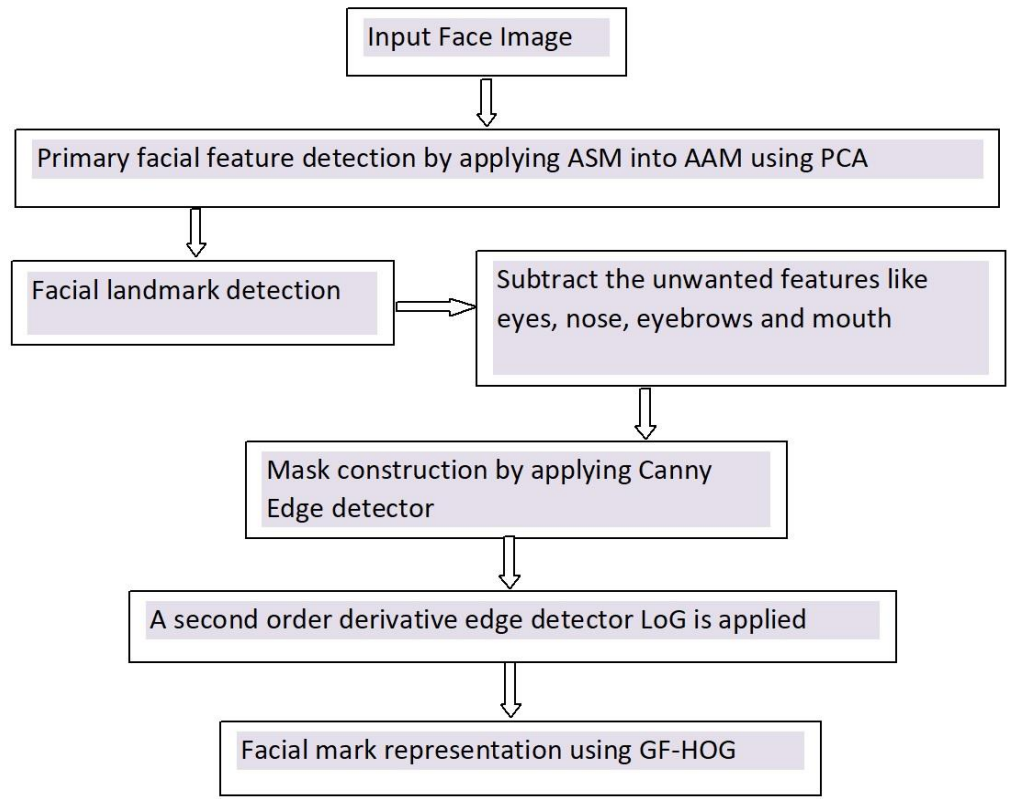

Figure 5.3: The proposed makeup-based facial mark detection algorithm

### 5.2.1 Facial Feature Detection and Mapping to Mean Shape

In this study, ASM into AAM with PCA algorithm [124] is used to find the 120 landmark points that take out the false positive of the facial features present, for instance, eyebrows, nose, eyes, and mouth from the face. These false positives can generate unnecessary features from the face and omit the accompanying unnecessary features to detect the facial marks from the facial mark detection process on cosmetic or makeup face. AAM and ASM converged to avoid the error rate observed by the cosmetic or make-up facials and to define the specific points of landmarks. Hence, the 120 landmarks were detected to make face boundary and other locations such as eyes, mouth, nose, and eyebrows. In terms of rotation and scale, the face images are uniform, admitting the representation of the person mark in a distinctive focus on the coordinate scheme. Thus any facial image on cosmetics is mapped into the median structure. Where, $S_i$ reflects the form of individual of the *N* face pictures centered on the one hundred and twenty landmark points in the database (*where i=1,2..., N*). Hence, the mean shape of the following equation is determined.

$$S_\mu = \sum_{1=n}^{N} S_i \tag{5.1}$$

Consider, $S_i$ is an individual make-up face image that is mapped using the Barycentric coordinate algorithm [115] to the mean form $S_\mu$. Thus, $S_i$ and $S_\mu$ are distinguished into a triangle that is presented in figure 5.4. Assume that *T* is a triangle in $S_i$, and that *T'* is found in $S_\mu$ by its corresponding triangle. Conceive that the $r'_1$, $r'_2$, and $r'_3$ of *T(T')* are $r_1$, $r_2$, and $r_3$ and their three parallel vertices. Thus *p* is the point inside the *T* expanded as $p = \alpha r_1 + \beta r_2 + \gamma r_3$, its parallel point p' and *T'* is expanded as $p' = \alpha r'_1 + \beta r'_2 + \gamma r'_3$, In this case conceived as $\alpha + \beta + \gamma = 1$.

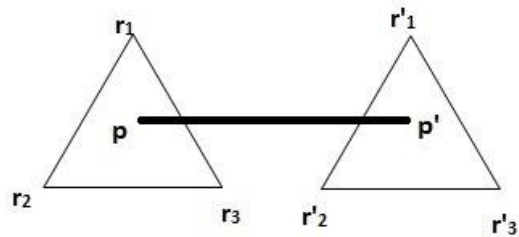

**Figure 5.4: Presents the mean shape mapping**

By using the Barycentric mapping technique [115], the value of the pixel is chosen and mapped into a mean shape (Figure 5.4). Thus, the entire points that are situated within the triangle, the mapping technique has chosen across and the texture $S_i$ is mapped to $S_\mu$. While the mapping technique is being created, entire makeup face images are tempered with regard to rotation and scale. The average face structure of the face image is shown in figure 5.5.

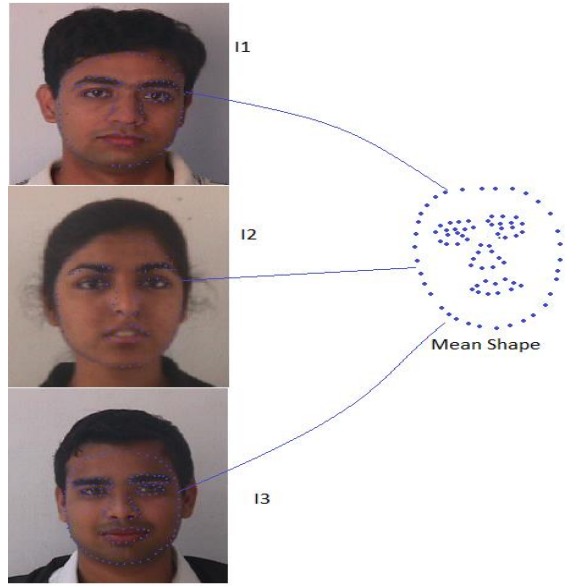

**Figure 5.5: The mean face construction**

### 5.2.2 Construction of the Mask

To eliminate the undesired features recognition of false positives from the cosmetic face or make-up face, the required mask is extracted from the mean form $S_μ$ and a user-specific mask. The false-positive will wrap the facial image around tiny irregular facial features. $M_g$ is referred to as the generic mask created from the mean form. For example, the false-positive beard and tiny wrinkles connected to the face characteristic are populated. A user-specific $M_s$ mask is then produced from the cosmetic face. The primary goal is to acquire the facial region to distinguish from the cosmetic or makeup used face with the permanent facial marks. Therefore, a user-specific mask was built using the Canny edge detector algorithm [113] from the facial edge image obtained. One of the most efficient ways of detecting the edges with low error rates is the canny edge detector algorithm. $M_s$ is improved by a total number of edges of $M_g$ and face image associated with $M_g$, which aided to transfer the most undesirable false positive out of the face image. The complete process of facial mark detection based on makeup is shown in figure 5.6.

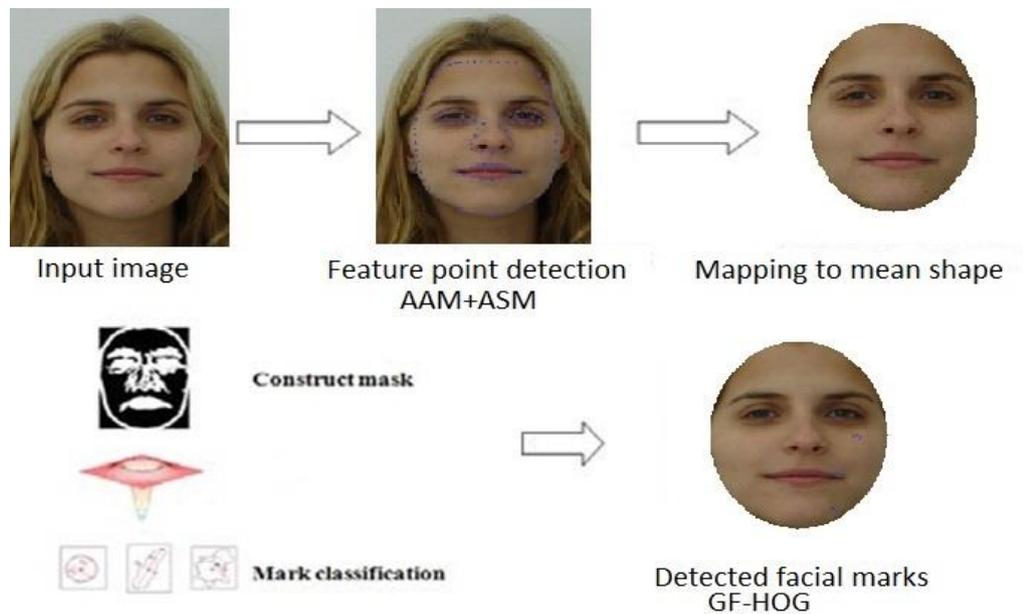

**Figure 5.6: The process of facial marks detection**

### 5.2.3 Facial Marks Detection from Makeup Face

To locate the facial marks on an obscured and prominent area of face, a second-order derivative edge detector LoG algorithm [76] was used. Using the LoG operator, the face image is filtered and the user-specific mask $M_s$ deduced, using binarization operation so as to it incorporates a sequence of threshold measurements with a diminishing order $t_i(i=1,2....,k)$. Hence, $t_i$ is then applied for sequencing until the predetermined value of $C_c$ is greater than the amount of the linked variable that follows. Once $C_c=10$ has been declared, the unneeded associated areas are released to identify the facial marks. At last, the facial markings are identified by the bounding box (rectangular or circular) and its location and size are shown. Detecting facial marks is shown in Figure 5.7.

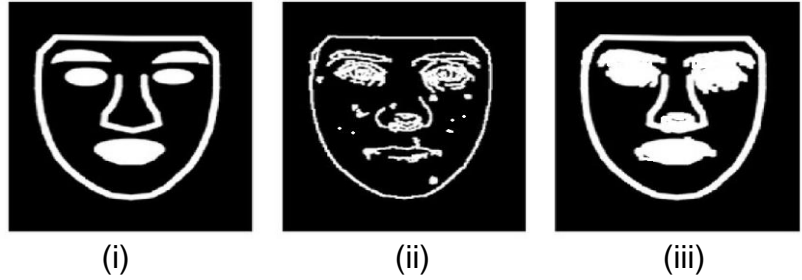

(i)　　　　　　　(ii)　　　　　　　(iii)

**Figure 5.7: (i) Generic mask (ii) Edges detection using Canny algorithm and (iii) A user-specific mask**

### 5.2.4 Facial Marks Representation

In this case, two methods were conceived to represent the facial marks based on the make-up or cosmetic face: This initial method involves specifying a bounding box for a facial mark; binaries with a threshold value determined by the mean value of the bounding pixels are the pixels inside the bounding box. Few encompassing parts are brighter or darker, the mean value has chosen consequence for the bounding box. Therefore, it classifies a label as linear to all, succeeded by circular to irregular points. In the second approach, the facial marks which are obtained from the cosmetic or makeup face image are converted by utilizing the algorithm Gradient Field - Histogram Oriented Gradients (GF-HOG) [156] with the destination to obtain a predefined amount of orientation intervals. An 8 x 8-dimensional block that carries one cell of the same size and acquires eight bins is then inserted into each histogram.

The primary purpose is to constitute a facial mark on the make-up face or cosmetic face; often in a separate region of the face, it presents two identical facial marks. Significantly, a facial mark situated on the chin and forehead is not considered wrongly. The facial markings on the face image are a major factor in the conception. Figure 5.8 demonstrates identified marks using the GF-HOG algorithm on the makeup face.

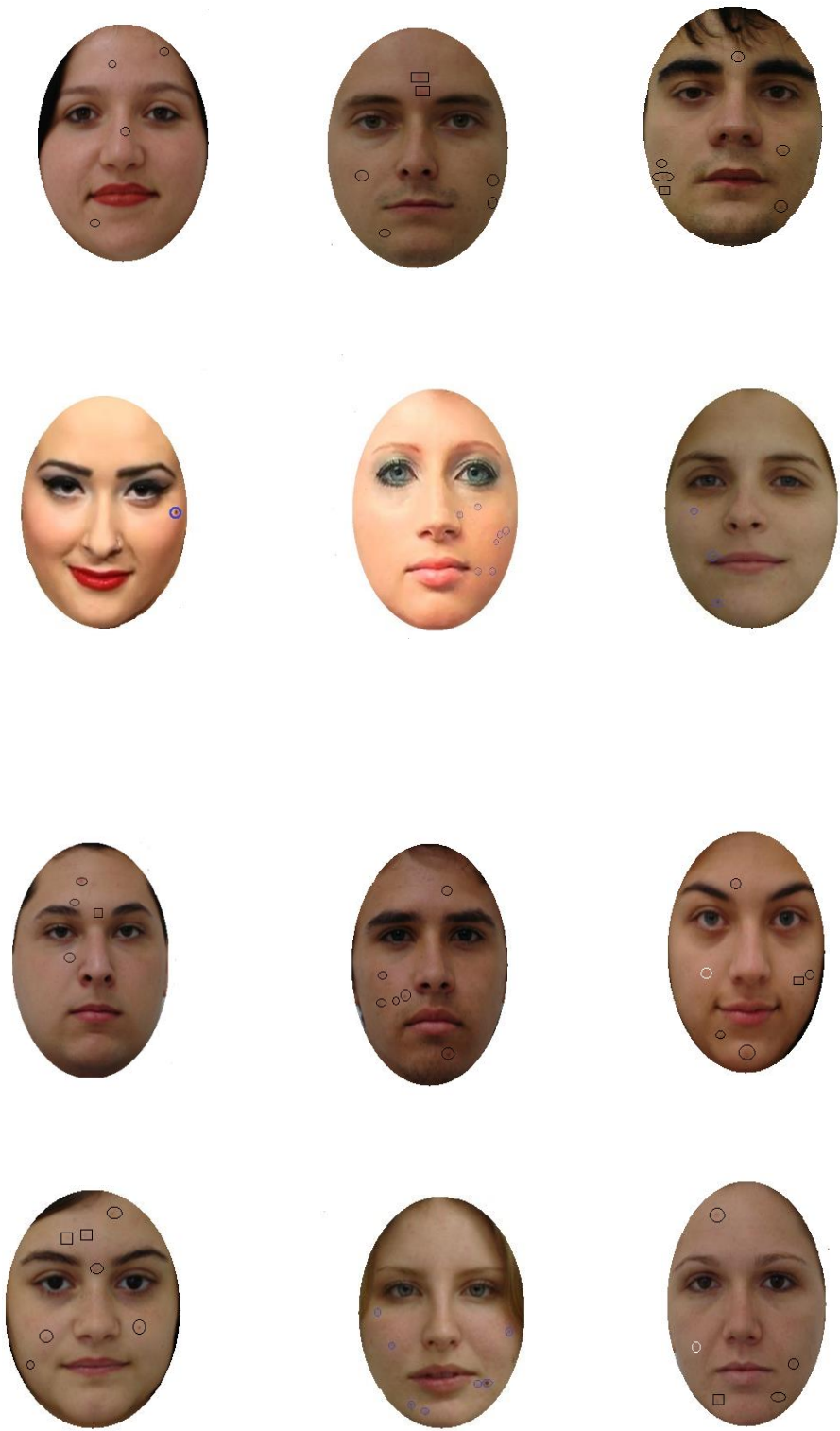

**Figure 5.8: Detection of facial marks from cosmetic applied face images**

### 5.2.5 Matching the Facial Marks

Consider $I_1$, $I_2$ are two face images, and with two different types of approaches. The collection of detected facial marks from cosmetic or makeup face $N_1$, $N_2$ with the similarity between $I_1$ and $I_2$ (here called *CAF*) was developed. The resemblance is defined in the initial approach when conceiving only the facial marks obtained in $I_1$ and undertaking to define them in $I_2$. Thus the representation of facial marks added depends on the strength of the pixel value. It is therefore possible to determine the resemblance among $I_1$ and $I_2$ using the following equation.

$$CAF_1 = \frac{\sum_{i=0}^{|N_1|} B(n_i, R_i)}{|N_1|} \tag{5.2}$$

Where each facial mark is $n_i \in N_1$ and $R_i$ is the triangular field, the area of possible matching was constructed around the central co-ordinates of image $I_2$. *B* establishes the similarity score concerned with the most efficient matching of the facial mark $n_i$ in the $R_i$ region, obtained by a Normalized Cross-Correlation (NCC) [157]. In the second method, the GF-HOG algorithm-based standard representation is used. It is developed with a lot of accuracies and matching the facial marks set $N_1$ and $N_2$ can be determined with the following equation.

$$CAF_2 = \frac{\sum_{i=0}^{|N_1|} \min D(n_i, n_j)}{|N_1|}, \forall n_j \in N_2 | (x_j, y_j) \in R_i \tag{5.3}$$

Where $x_j$ and $y_j$ are each facial mark $n_j \in N_2$, and the central coordinate points. $R_i$ is also the region on the face employed to ensure that only certain $n_j \in N_2$ in $R_i$ are well selected in the matching phase and ensures that the facial marks match improvements. The distance D among the facial marks is measured by applying the algorithm [117], which showed the most satisfactory outcome in the experimental process when comparing the histogram.

## 5.3 RESULTS AND DISCUSSIONS

For analysis, cosmetic face images from the Makeup Induced Face Spoofing (MIFS) dataset [158] were obtained to perform the experiment. The MIFS database, which corresponds to 107 female subjects is collected from YouTube makeup tutorials in 2013. Before makeup and after makeup applied faces were captured, the facial pictures are contained with the subjects. In this case, the individual subject includes four samples that are used before the utilization of makeup and after the usage of makeup. Two samples contain the facial images of an individual: after the use of makeup and before the application of makeup. In the eyes region, some modifications occurred due to makeup or cosmetic applications since the eyes have been accentuated by different kinds of cosmetics or makeup on the eyes. It is additionally, a change that happened on the skin on account of the utilization of cosmetics on the lips. An average of three facial marks are obtained as a result of the annotation technique, e.g. birthmarks, moles, and grains are the preeminent often obtained. Different marks, for instance, darkened, freckle, enlightened, warts, scars, tattoos, and pockmarks do not appear to be noticeable apparently because of the makeup utilized facial image.

Several cosmetic face images with manually annotated facial marks were created to verify the experimental findings are depicted in figure 5.10. Hence, different sort of facial marks is manually labeled as a ground truth. This method allows us to quantify the proposed cosmetic or makeup based facial marks detection method. Concerning computational complications during the experiment, the most consuming procedure is facial marks as feature extraction. Taking into account this recognition, the features extraction process is remarked to be the most common part in choosing generally run-time performance. In this study, the experiment is done utilizing a computer with Intel Core i7, a 7600U processor, 3.5 GHz, and 16 GB RAM.

No database has focused upon cosmetics or makeup for twins and identical faces, it could not be obtained from the existing study in order to analyze makeup-related facial mark recognition. Hence, this study produced the databases from the internet source, which picked up for the identical face 500 face images (we call MFS) and for the twins 450 face images (we call MFT). In figure 5.9. (a) similar face in the MFS dataset and (b) twin's dataset, some of the accumulated face images are given. This database contains variants in expression, aging, illuminations, and pose from the manual, and automatically annotated marks on the facial were obtained. The database gives manual annotations of facial marks that acknowledges their assortment and place of birthmarks, moles, and grains, etc. As each person has unique facial marks, the twin faces and identical faces are sorted by facial marks. In this chapter, the pattern used for individual analysis is explained by measuring their techniques in this database.

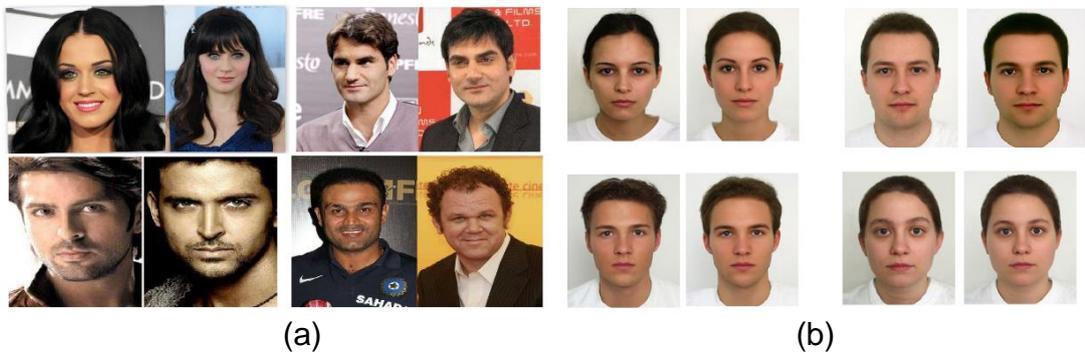

(a)           (b)

**Figure 5.9: (a) Similar faces in the MFS dataset and**
**(b) Twins in the MFT dataset**

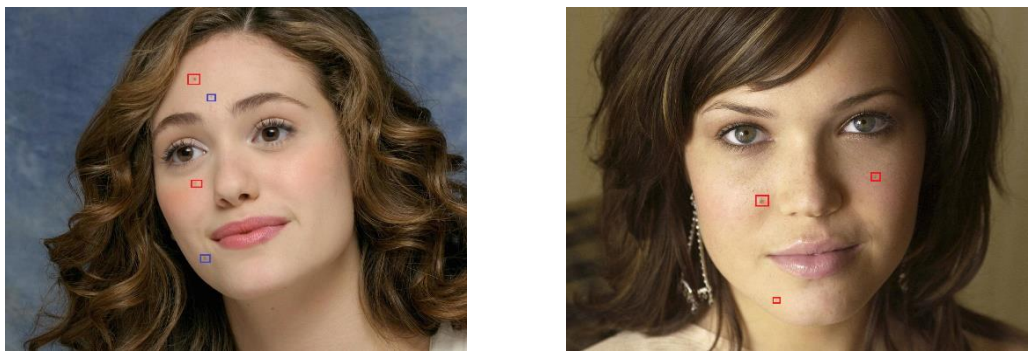

**Figure 5.10: Manually annotated face images from the MIFS dataset**

### 5.3.1 Evaluation of Facial Marks Detection from Makeup Face

In this experiment, the proposed facial marks recognition algorithm for cosmetic face or makeup face employed (here we named as C-GHOG), the accuracy validation was performed by evaluating Precision and recall with the stated 107 image database (MIFS dataset), 450 images (MFT dataset) and 500 images (MFS dataset). We summed up the prime approach to evaluate facial mark detection so that correlation can be build up among detected facial mark and an annotated face. Consider, an image *I* with annotation marks *A*; the facial marks detection $n_i$ is conceived as precise if $\exists n_a \in A$.

We calculate, $\frac{area(n_i) \cap area(n_a)}{area(n_i) \cup area(n_a)} \geq t_0$ (5.4)

The value of the threshold $t_0$ was generated empirically as 0.4.

Table 5.1 Comparison of C_GHOG with C_S, V_J detector using the MFIS dataset

| Algorithm | Precision (%) | Recall (%) |
|---|---|---|
| C_GHOG | 77.1 | 61.6 |
| V_J | 28.4 | 9.1 |
| C_S | 61.9 | 44.3 |

In the latest study of the algorithm for facial mark detection for makeup or cosmetically used face pictures, it was impractical to obtain the already implemented techniques. Henceforth, to equate the proposed facial marks detection algorithm for the cosmetic applied face with other facial marks detection algorithms on the same cosmetic used face database. In this experiment, compared with the Viola and Jones object detector [122] (we call as V_J) and Canny with SURF algorithm [82] (we call as C_S). These algorithms are built up for facial marks recognition and trained to locate various sorts of facial marks.

**Table 5.2  Comparison of C_GHOG with C_S, V_J techniques using the MFT dataset**

| Algorithm | Precision (%) | Recall (%) |
|---|---|---|
| C_GHOG | 76.23 | 61.12 |
| V_J | 28.99 | 8.34 |
| C_S | 59.28 | 42.99 |

There are a collection of 100 positive samples with facial marks of the trained subject and 100 negative samples in the image region without marks, was developed to obtain the makeup-based facial marks. The acquired outcomes have appeared in table 5.1, 5.2, and 5.3. The proposed algorithm for cosmetic-based face detection of facial marks performs a higher precision and recall estimate than the algorithm of Viola-Jones and Canny with the SURF algorithm.  It also shows that with fewer false positives, it can more reliably recognize facial marks from make-up or cosmetic faces.

**Table 5.3  Comparison of C_GHOG with C_S, V_J methods using the MFS dataset**

| Algorithm | Precision (%) | Recall (%) |
|---|---|---|
| C_GHOG | 79.89 | 63.20 |
| V_J | 27.11 | 10.79 |
| C_S | 61.34 | 43.99 |

### 5.3.2 Validation of Face Verification

The measurement parameters for face verification are the False Recognition Rate (FRR) and False Acceptance Rate (FAR). FAR is the indicator of the probability that an access assault by an unauthorized individual is wrongly recognized by biometric authentication. It is equivalent to the amount of incorrect adaptations separated by the aggregate amount of efforts at identification. FRR is the likeliness ratio that an entry attempt by a known client would be wrongly ignored by biometric authentication. It is the proportion of the number of incorrect identifications divided by the number of attempted identification. The Equal Error Rate (ERR) is conceived as the error when FAR(i) = FRR(i). Operating Point (OP) is another way to test a biometric authentication technique. Hence, evaluating the number of admittance attempts by a legitimate client asserted when a few illegal clients are admitted. In this study, the OP is conceived at a constant value, FAR = 0.1%. Thus it presents FRR (FM_OP_FRR) throughout this experiment.

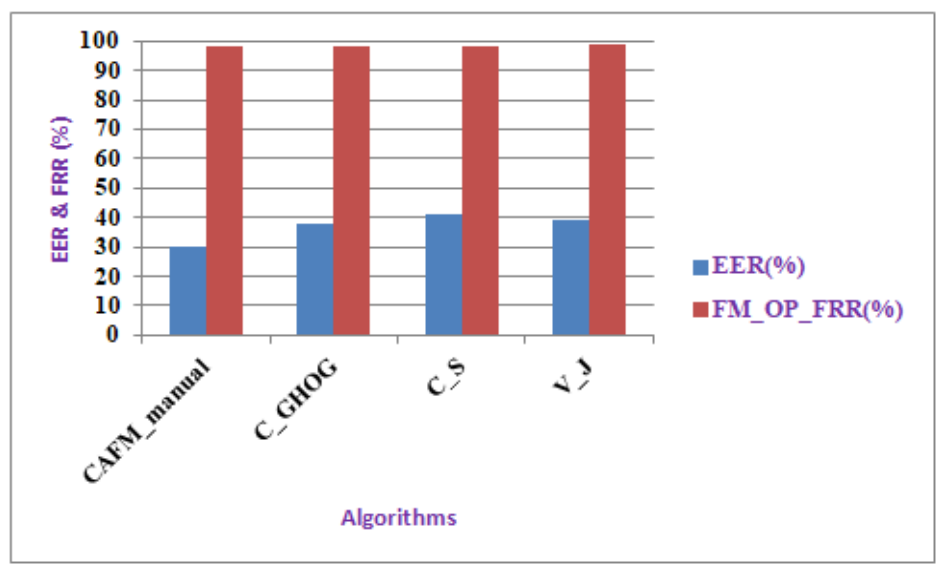

**Figure 5.11: The proposed C_GHOG algorithm compared with the various algorithms using the MIFS dataset**

In this analysis, the MIFS database was taken into account by employing facial mark matching to contrast facial mark detection algorithms based on the accuracy of the verification. Figure 5.11 below presents the outcomes. The table includes the results of the matching of the facial marks by manual annotation (CAFM_mannual). The proposed facial mark detection algorithm in this experiment obtained more advantageous outcomes contrasted to other algorithms. Manual annotations contribute to lower error rates than automated detection methods. It shows that the best results can be achieved by manually annotating the algorithm for face mark detection.

To validate the advantages of utilizing facial marks, the proposed facial mark matching algorithm is combined with two common algorithms to obtain the best result for individual face validation and identification in face recognition execution. A) The nearest neighbor classifier with ASM into AAM using PCA [124] and B) Fisher Vector (FV) [123].

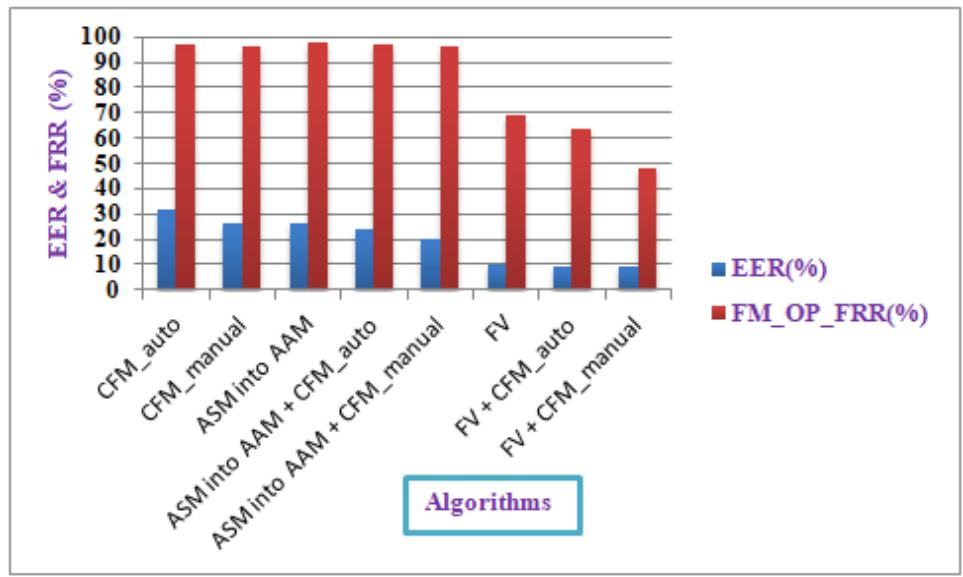

**Figure 5.12: Comparison among ASM into AAM, FV, and the relevant variant using the MIFS dataset**

In this chapter, one to one face verification is used to conduct the experiment and employed the databases of MFT dataset, MIFS dataset, and MFS dataset. Facial mark detection from the cosmetic applied face with automatic facial mark detection (CFM_auto) is then calculated and equated with the results of the FV algorithm and ASM into AAM algorithm with their combined algorithms. In this case, similar perceptions under comparable conditions are admitted by utilizing the facial marks of manual annotation (CFM_manual). This provides the benefits of the suggested verification algorithm. The outcome depends on FM_OP_FRR and EER for MFT, MIFS, and MFS datasets that have appeared in Figures 5.12, 5.13, and 5.14 respectively. If the first two rows of facial mark detection tend to be not adequately suited to be used in overcoming verification technology in combination with replacement classical face recognition algorithms such as the FV algorithm and ASM in the AAM algorithm that increase the accuracy of verification techniques. The increase in accuracy in the EER is significant yet on account of FV+ FM_manual, it shows that EER is diminished, which appeared in figure 5.14.

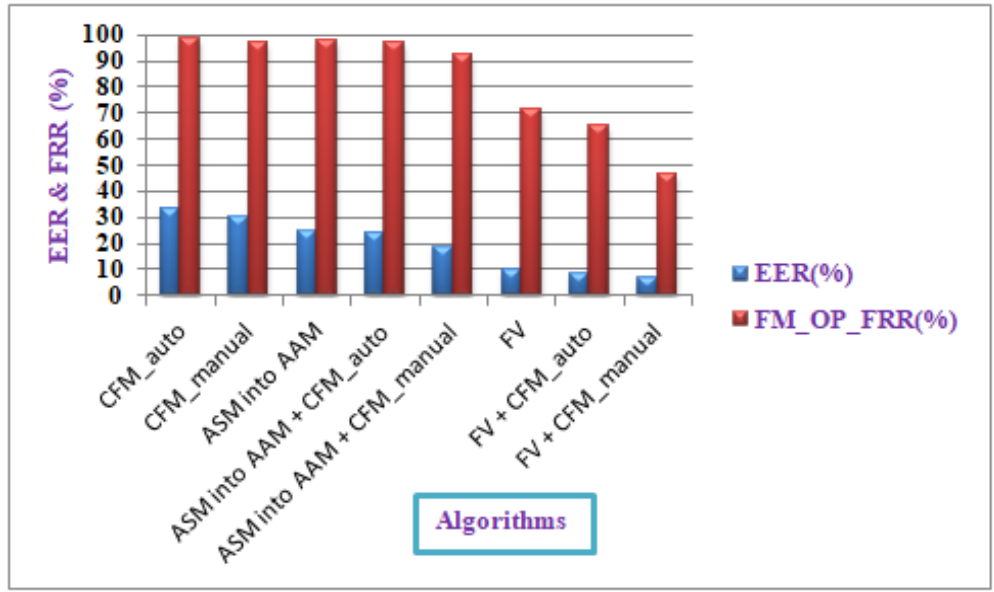

**Figure 5.13: Compare among ASM and AAM, FV, and the proposed variants using the MFT dataset**

The error produced in the automatic facial marks detection is unrelated to the final results due to the combination of ASM with AAM algorithm. Taking into account the difference between ASM into AAM + FM-manual and ASM into AAM + FM-auto. It enhanced the facial mark detection procedure when combined with the FV algorithm, it could achieve a greater benefit on the outcome. In Figures 5.13 and 5.14, the progress of the automatically facial marks detected is a lower number which is questionable as a result of the trouble in the MFT dataset and MFS dataset because of the makeup face image. Therefore, ameliorated using the manual annotation of facial marks.

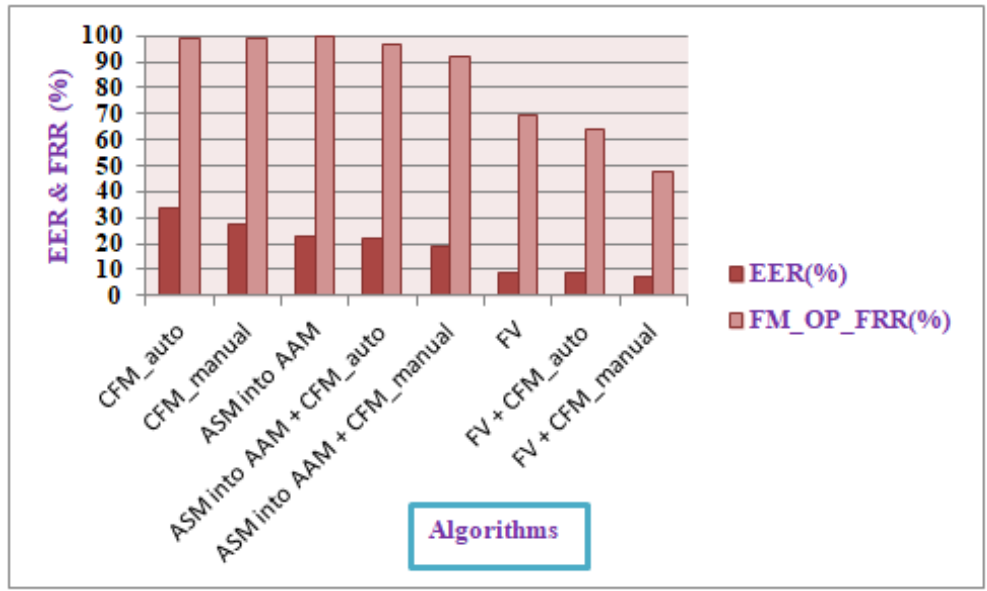

**Figure 5.14: Comparison among ASM into AAM, FV, and the proposed variants using the MFS dataset**

### 5.3.3 Validation of Face Identification

In this research work, a validation experiment has been conducted based on the MFIS database, and fivefold cross-validation is executed by using one facial picture of every individual as a gallery, and the rest of the facial pictures are examined. The effects of the matching of facial marks with automated recognition of facial marks and manual recognition of facial marks will be distinguished in the investigation. Therefore the combination with the FV algorithm was also compared and the most beneficial results were proven. The detected outcomes with respect to ARR at Rank1 (AAR_R1) and Rank 5 (AAR_R5) appears in figure 5.15. The use of facial marks alone is not achievable, henceforth the best results for face distinguishing evidence are improved by mixing with a face recognition algorithm, figure 5.15 shows the outcome. The FV algorithm attained a 88% Recognition Rate (RR) in rank 1 and shows improvement when facial marks indistinguishable is considered on both manual and automatic facial marks variations. The finding of these results up to Rank 10 has appeared in figure 5.16.

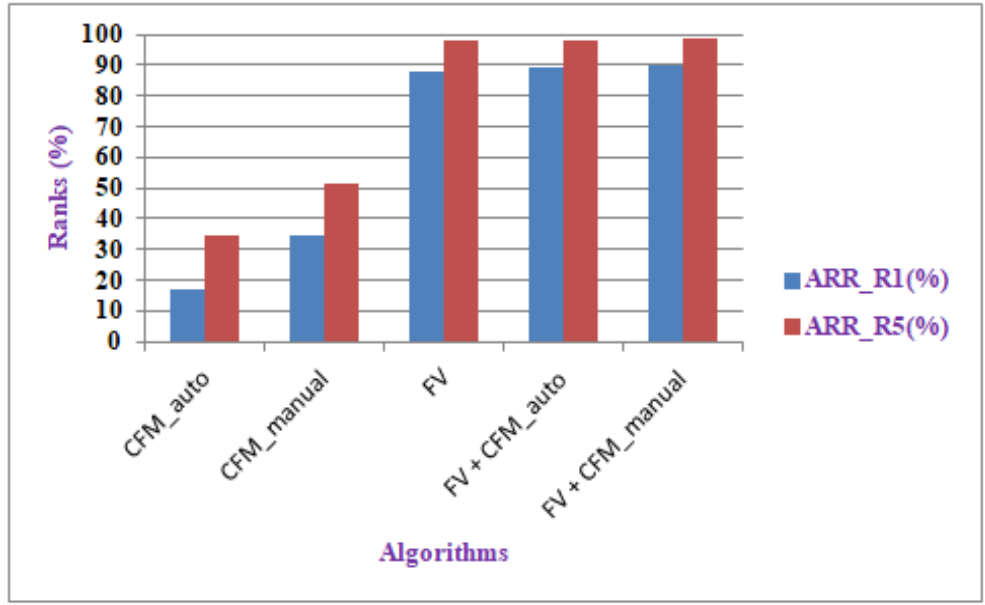

**Figure 5.15: The proposed variation of the recognition experiment's combination with the FV technique**

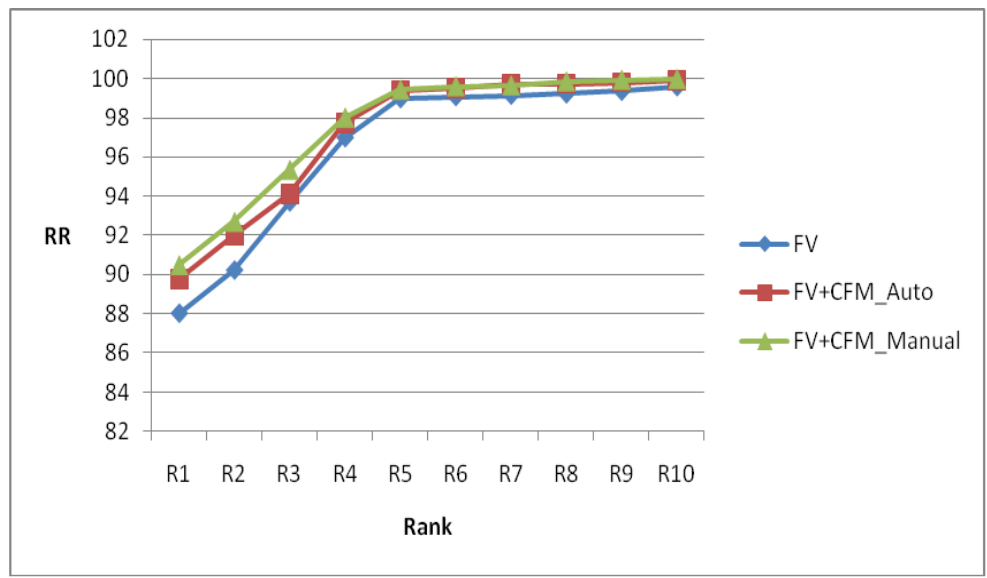

**Figure 5.16: The graph is based on the face identification ARR curve (%)**

Figure 5.16 presents the cumulative average recognition rate curve in percentage. FV technique alone, a combination of FV with automatic detection, and a combination of FV with manual annotation are shown in the graph. It improved the results when combined with manual annotation and the FV technique. As compared to the results of R1, R10 gives a more proficient outcome.

## 5.4 CONCLUSIONS

In this study presented an algorithm to detect the facial marks from the makeup or cosmetic used to face for secure biometric passport authentication. As a method for recognizing traveler's biometrics authentication ought to be utilized around the world. This research work anticipated a verification system where depends on facial marks detection to validate individuals by using the algorithm ASM into AAM utilizing PCA, Canny edge detector, and GF-HOG. This study is introduced to recognize and distinguish indistinguishable twins and similar faces. Here demonstrated makeup or cosmetic based face recognition to improve passport authentication. Nevertheless, the utilization of the proposed method confronted a portion of the difficulties because of cosmetic or makeup application. Thus, in this proposed chapter, the mixed of the algorithm for facial marks detection matching classical techniques could achieve fewer errors. The technique is adequate to distinguishing precisely the facial marks with fewer falls positive from the cosmetic or make-up face.

# CHAPTER 6

# CONCLUSIONS AND FUTURE WORKS

In this research work, proposed the face recognition technique by overcoming the failure on account of inadequate matching of photos of unknown faces. To secure both the personal and biometric information by encrypting the user's biometric data and encoded into QR code and HCC2D code. The conclusions and future works are given below.

## 6.1  CONCLUSIONS

For a secure biometric passport, we have proposed a novel system for personal information and biometric features encryption and encoded into the QR code. The method of electronic passport authentication is focused on the detection of facial mark size, encrypted data, detection of hand geometry, and encrypted biometrics into the QR code for border control applications and national security. The biometric passport data is encrypted with the AES and SHA-256 algorithm to protect the data from the decipherable threats of data leakage. Also, face recognition approaches focused on the identification of facial blemishes are also proposed by applying AAM using PCA to detect the facial landmarks, Canny edge detector to detect the edges of the facial blemishes, and the SURF algorithm is employed to recognize the facial blemishes features. Here, facial blemishes features are considered as a signature to authenticate a person. These facial blemishes features are encrypted by applying a cryptographic SF algorithm to secure biometric information. To keep the information safe, the encrypted biometric data is encoded into the HCC2D code and inserted in an electronic passport. This method ensures that the confidentiality and protection of the e-passport since encrypted biometric information encoded into the HCC2D code, which can not employ as an active feature.  Certainly, HCC2D codes are economical, passive read-only, and information cannot be altered.

We have also proposed and executed an algorithm for the detection of facial marks from make-up or cosmetic faces for a secure biometric passport in the field of personal recognition for national security. The proposed method relies on facial markings as a signature to authenticate biometric passports based on face recognition. In this research, conceived that it is important to use a means of identifying travelers with biometrics must be employed globally. It has also been introduced to identify and recognize similar twins and identical faces. The proposed technique would improve national security and biometric authentication for biometric passports. This research work concentrates on protection and privacy for future improvement by using 2D barcodes such as QR code, HCC2D code, etc. to enlarge the data size to store the information.

## 6.2. RESEARCH CONTRIBUTIONS

The followings are the major research contributions:

- Biometric encryption into the QR and HCC2D code.
- Authentication of multi-biometrics that detects the size of the facial mark and the detection of hand geometry to enhance protection.
- The use of AES and SHA - 256 algorithms encrypts personal data and biometric information.
- To be passive and protect the data by encrypting it into the QR code, it is important to secure the data that is not revealed without the users' knowledge or permission.
- While performing the verification, the encrypted biometric information in the QR code will update state-of-the-art biometric passport security features as it is not an active part.
- The biometric information is encrypted and encoded into the HCC2D code by applying the SF algorithm.
- The stored encrypted biometric information in the HCC2D code could not reveal data without the permission or consent of its users.
- As it is not an active function, while conducting the authentication, the encrypted biometric data in the HCC2D code will improve the security of the biometric passport.
- A face recognition based on cosmetic applied is proposed to enhance the biometric passport authentication.

## 6.2 FUTURE WORKS

Passports are increasingly developing into fully digitized, electronic documents from hand-written documents with glued-in images. Using digital certificates has the ability to improve national security considerably. Moreover, via automated border control gates, machine-readable electronic IDs significantly accelerate passport control. Biometric passports, however, raise a range of problems as well. Interoperability and confidence are important to allow authorities to recognize authentic passports from other countries instantly.

To upgrade state-of-the-art for the national security to ameliorate the biometric authentication process for the biometric-passport, it requires further study on future enhancement.

- Enhance the study on face recognition performance for the biometric passport.
- To enlarge the storage size of data into the QR code and HCC2D code.

## LIST OF PUBLICATIONS

**The following articles are published in the reputed journals/proceedings:**

# TECHNICAL BIOGRAPHY

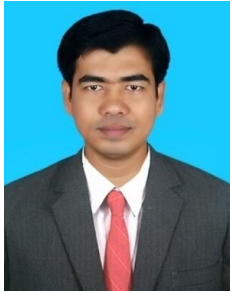

**Mr. Ziaul Haque Choudhury (RRN. 140873107002)** was born in Silchar, Cachar district, Assam, India. He received his Bachelor's degree in Computer Science from the Bharathidasan University, Trichy, Tamilnadu, India in the year 2007. He did his M.Sc.(Engineering) in Information Technology, Department of Computer Science and Engineering from the Annamalai University, Chidambaram, Tamilnadu, India in the year 2009. Also, he received his Master in Research, M.Tech.(by Research) in Information Technology from B.S. Abdur Rahman Univesity, Chennai, Tamilnadu, India in the year 2013. He is a Senior Research Fellow (SRF) and he is currently pursuing his Ph.D. degree in multibiometric security for an electronic passport in the Department of Information Technology, B.S. Abdur Rahman Crescent Institute of Science and Technology (Deemed to be University), Chennai, India. He received an award of Maulana Azad Nation Fellowship from University Grand Commission, Govt. of India, Delhi, for doing his Ph.D. degree. His area of interests includes Biometrics, Image processing, Cryptography, Network Security, Artificial Intelligence, Deep learning.

The e-mail ID is: ziaulms@gmail.com and the contact number is: +91-8072257855.